\documentclass{article}

    \PassOptionsToPackage{numbers, compress}{natbib}

\usepackage[final]{neurips_2022}

\usepackage[utf8]{inputenc} 
\usepackage[T1]{fontenc}    
\usepackage{hyperref}       
\usepackage{url}            
\usepackage{booktabs}       
\usepackage{amsfonts}       
\usepackage{nicefrac}       
\usepackage{microtype}      
\usepackage{xcolor}         

\usepackage{subcaption}
\usepackage{multirow}
\usepackage{makecell}
\usepackage{graphicx}
\usepackage{algorithm}
\usepackage{algorithmic}
\usepackage{mathabx}
\usepackage{amssymb}
\usepackage{pifont}

\newif\ifcomments
\commentsfalse
\ifcomments
  \newcommand{\colornote}[3]{{\color{#1}\bf{#2: #3}\normalfont}}
\else
  \newcommand{\colornote}[3]{}
\fi
\newcommand {\JB}[1]{\colornote{red}{JB}{#1}}

\newcommand{\argmin}{\mathop{\mathrm{argmin}}\limits}

\newcommand{\targetloss}{\mathcal{L}}

\newcommand{\metagrad}{\nabla_{\supportdataset}{F(\supportdataset)}}
\newcommand{\metagradsingle}{\nabla_{\supportdataset}{\targetloss \left( \optalg \left(\nnparam, \supportdataset \right), \targetdataset\right)}}
\newcommand{\grad}[1]{\nabla_{#1}}
\newcommand{\targetx}{X_{t}}
\newcommand{\targety}{Y_{t}}
\newcommand{\targetdata}{(\targetx, \targety)}
\newcommand{\targetdataset}{\mathcal{T}}
\newcommand{\supportx}{X_{s}}
\newcommand{\supporty}{Y_{s}}
\newcommand{\supportdata}{(\supportx, \supporty)}
\newcommand{\supportdataset}{\mathcal{S}}
\newcommand{\numclass}{C}
\newcommand{\numpool}{m}

\newcommand{\modelpool}{\mathcal{M}}

\newcommand{\modeldist}{P_{\theta}}
\newcommand{\nnparam}{\theta}
\newcommand{\maxonlineupdate}{K}

\newcommand{\ddlr}{\alpha}
\newcommand{\optalg}{\mathcal{A}lg}

\newcommand{\gram}[2]{K^{\theta}_{{#1}{#2}}}

\newcommand{\algname}{FRePo}

\title{Dataset Distillation using Neural Feature Regression}
\author{
  Yongchao Zhou \\
  Department of Computer Science\\
  University of Toronto\\
  \texttt{yongchao.zhou@mail.utoronto.ca} \\
  \And
  Ehsan Nezhadarya \\
  Toronto AI Lab \\ LG Electronics Canada \\
  \texttt{ehsan.nezhadarya@lge.com} \\
  \And
  Jimmy Ba \\
  Department of Computer Science\\
  University of Toronto\\
  \texttt{jba@cs.toronto.edu} \\
}

\begin{document}

\maketitle

\begin{abstract}
Dataset distillation aims to learn a small synthetic dataset that preserves most of the information from the original dataset. Dataset distillation can be formulated as a bi-level meta-learning problem where the outer loop optimizes the meta-dataset and the inner loop trains a model on the distilled data. Meta-gradient computation is one of the key challenges in this formulation, as differentiating through the inner loop learning procedure introduces significant computation and memory costs. In this paper, we address these challenges using neural Feature Regression with Pooling (\algname), achieving the state-of-the-art performance with an order of magnitude less memory requirement and two orders of magnitude faster training than previous methods. The proposed algorithm is analogous to truncated backpropagation through time with a pool of models to alleviate various types of overfitting in dataset distillation. \algname\ significantly outperforms the previous methods on CIFAR100, Tiny ImageNet, and ImageNet-1K. Furthermore, we show that high-quality distilled data can greatly improve various downstream applications, such as continual learning and membership inference defense. Please check out our webpage at \url{https://sites.google.com/view/frepo}.

\end{abstract}

\section{Introduction}\label{sec:intro}
Knowledge distillation \citep{DBLP:journals/corr/HintonVD15} is a technique in deep learning to compress knowledge for easy deployment. Most previous works focus on model distillation \citep{DBLP:conf/interspeech/FukudaSKTCR17,DBLP:conf/iclr/PolinoPA18} where the knowledge acquired by a large teacher model is transferred to a small student model. In contrast, dataset distillation \citep{DBLP:journals/corr/WangDD18, DBLP:conf/iclr/DC} aims to learn a small set of synthetic examples preserving most of the information from a large dataset such that a model trained on it can achieve similar test performance as one trained on the original dataset. Distilled data can accelerate model training and reduce the cost of storing and sharing a dataset. Moreover, its highly condensed and synthetic nature can also benefit various applications, such as continual learning \citep{ DBLP:conf/iclr/DC, DBLP:journals/corr/abs-2103-15851,DBLP:conf/icml/DSA, DBLP:journals/corr/DM}, neural architecture search \citep{DBLP:conf/iclr/DC, DBLP:conf/icml/DSA}, and privacy-preserving tasks \citep{DBLP:journals/corr/abs-2104-02857,DBLP:journals/corr/abs-2008-04489}.

Dataset distillation was first studied by \citet{DBLP:conf/icml/MaclaurinDA15} in the context of gradient-based hyperparameter optimization and subsequently \citet{DBLP:journals/corr/WangDD18} formally proposed dataset distillation as a new task. \JB{What is "it"? dataset distillation or hyperparemter optimization?} Dataset distillation can be naturally formulated as a bi-level meta-learning problem. The inner loop optimizes the model parameters on the distilled data (meta-parameters), while the outer loop refines the distilled data with meta-gradient updates.

One key challenge in dataset distillation is computing the meta-gradient. Several methods \citep{DBLP:journals/corr/WangDD18, DBLP:conf/icml/MaclaurinDA15, DBLP:journals/corr/BohdalFDD20,DBLP:journals/corr/SucholutskyIDD19} compute it by back-propagating through the unrolled computation graph, but they often suffer from huge compute and memory requirement \citep{DBLP:conf/icml/VicolMS21}, training instability \citep{DBLP:conf/icml/PascanuMB13, DBLP:conf/icml/MetzMNFS19}, and truncation bias \citep{DBLP:conf/iclr/WuRLG18}. To avoid unrolled optimization, surrogate objectives are used to derive the meta-gradient, such as gradient matching \citep{DBLP:conf/iclr/DC, DBLP:conf/icml/DSA, DBLP:journals/corr/DCC}, feature alignment \citep{DBLP:journals/corr/DM,DBLP:journals/corr/CAFE}, and training trajectory matching \citep{DBLP:journals/corr/MTT}. Nevertheless, a surrogate objective may introduce its own bias \citep{DBLP:journals/corr/CAFE}, and thus, may not accurately reflect the true objective. An alternative is using kernel methods, such as Neural Tangent Kernel (NTK) \citep{DBLP:conf/nips/LeeXSBNSP19}, to approximate the inner optimization \citep{DBLP:conf/iclr/KIP1, DBLP:conf/nips/KIP2}. However, computing analytical NTK for modern neural network can be extremely expensive \citep{DBLP:conf/iclr/KIP1, DBLP:conf/nips/KIP2}.

Even with an accurate meta-gradient, dataset distillation still suffers from various types of overfitting. For instance, the distilled data can easily overfit to a particular learning algorithm \citep{DBLP:journals/corr/WangDD18, DBLP:journals/corr/SucholutskyIDD19, DBLP:journals/corr/MTT}, a certain stage of optimization \citep{DBLP:journals/corr/SucholutskyIDD19, DBLP:journals/corr/CAFE}, or a certain network architecture \citep{DBLP:conf/iclr/DC, DBLP:conf/icml/DSA, DBLP:journals/corr/MTT, DBLP:conf/iclr/KIP1, DBLP:conf/nips/KIP2}. Meanwhile, the model can also overfit the distilled data during training, which is the most common cause of overfitting when we train on a small dataset. All these kinds of overfitting impose difficulties on the training and general-purpose use of the distilled data. 

We propose an efficient meta-gradient computation method and a ``model pool'' to address the overfitting problems. The bottleneck in meta-gradient computation arises due to the complexity of inner optimization, as we need to know how the inner parameters vary with the outer parameters \citep{DBLP:conf/aistats/LorraineVD20}. However, the inner optimization can be pretty simple if we only train the last layer of a neural network to convergence while keeping the feature extractor fixed. In this case, computing the prediction on the real data using the model trained on the distilled data can be expressed as a kernel ridge regression (KRR) with respect to the conjugate kernel \citep{Neal1995BayesianLF}. Hence, computing the meta-gradient is simply back-propagating through the kernel and a fixed feature extractor. To alleviate overfitting, we propose to maintain a diverse pool of models instead of periodically training and resetting a single model as in prior work \citep{DBLP:conf/icml/DSA, DBLP:journals/corr/SucholutskyIDD19, DBLP:journals/corr/DCC}. Intuitively, our algorithm targets the following question: what is the best data to train the linear classifier given the current feature extractor? Due to the diverse feature extractors we use, the distilled data generalize well to a wide range of model distributions. 

\begin{figure}
  \centering
\begin{subfigure}[b]{0.325\textwidth}
  \includegraphics[width=1.0\linewidth]{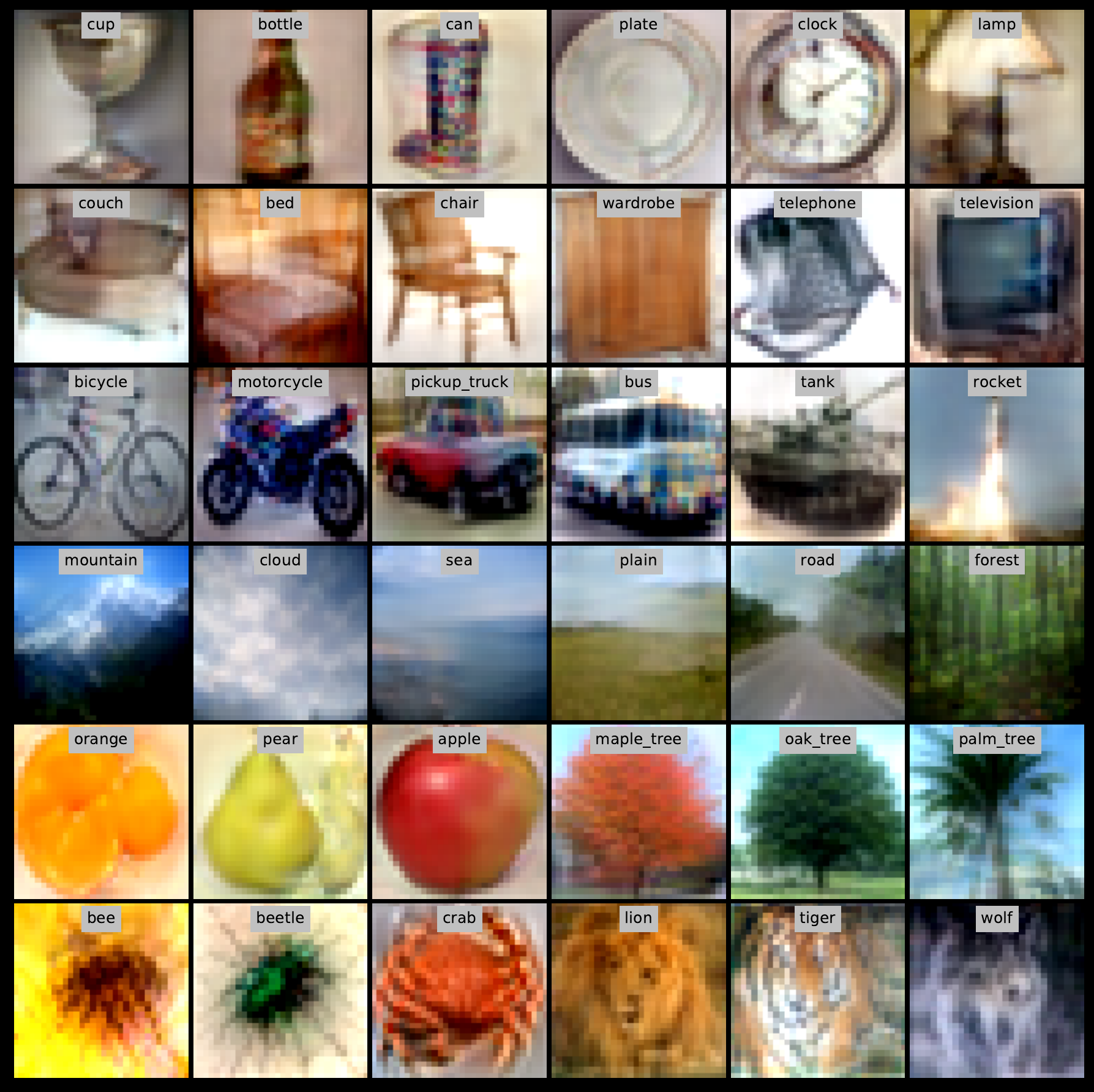}  
  \caption{CIFAR100}
\end{subfigure}
\begin{subfigure}[b]{0.325\textwidth}
  \includegraphics[width=1.0\linewidth]{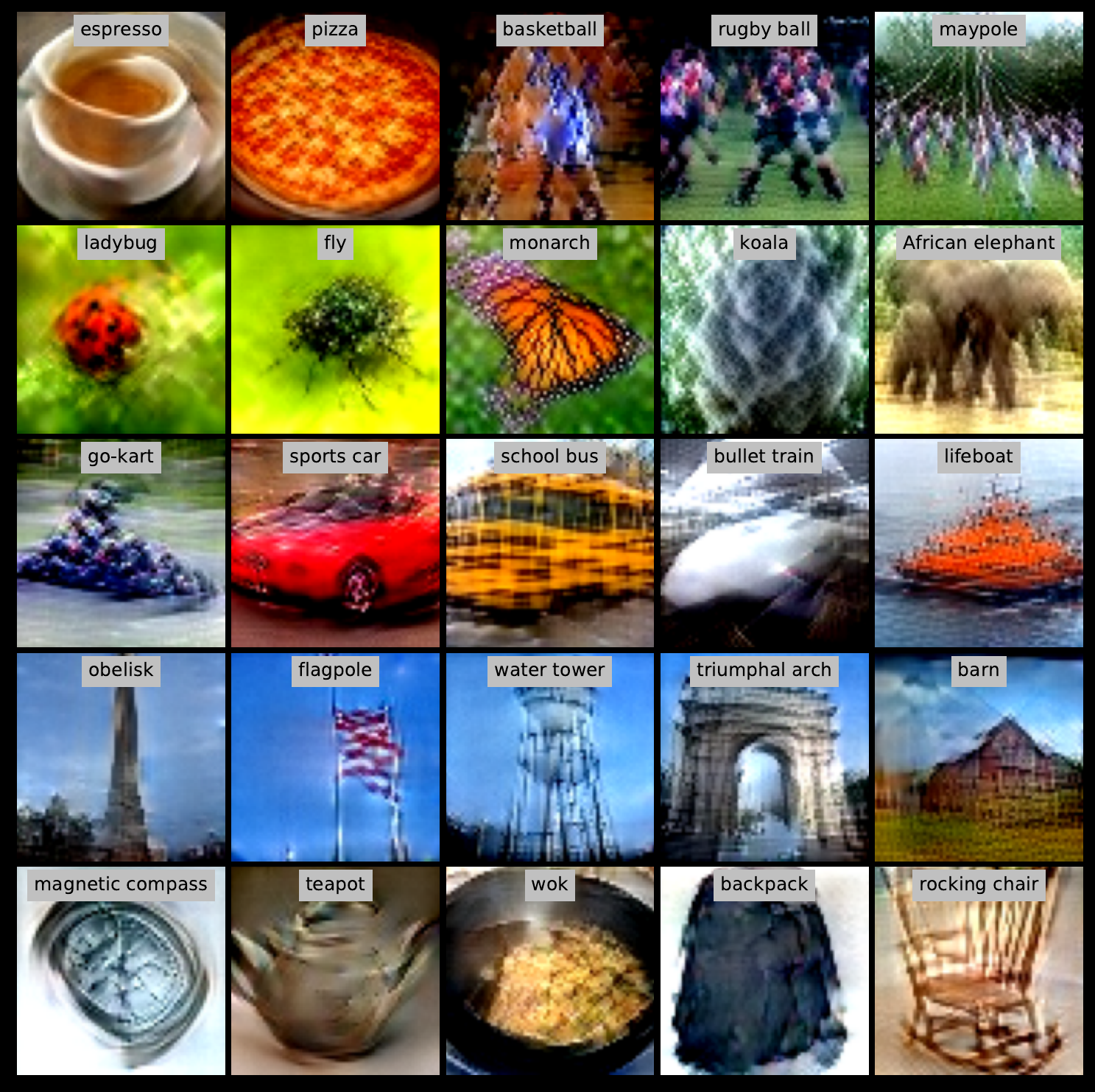}
  \caption{Tiny ImageNet}
 \end{subfigure}
 \begin{subfigure}[b]{0.325\textwidth}
  \includegraphics[width=1.0\linewidth]{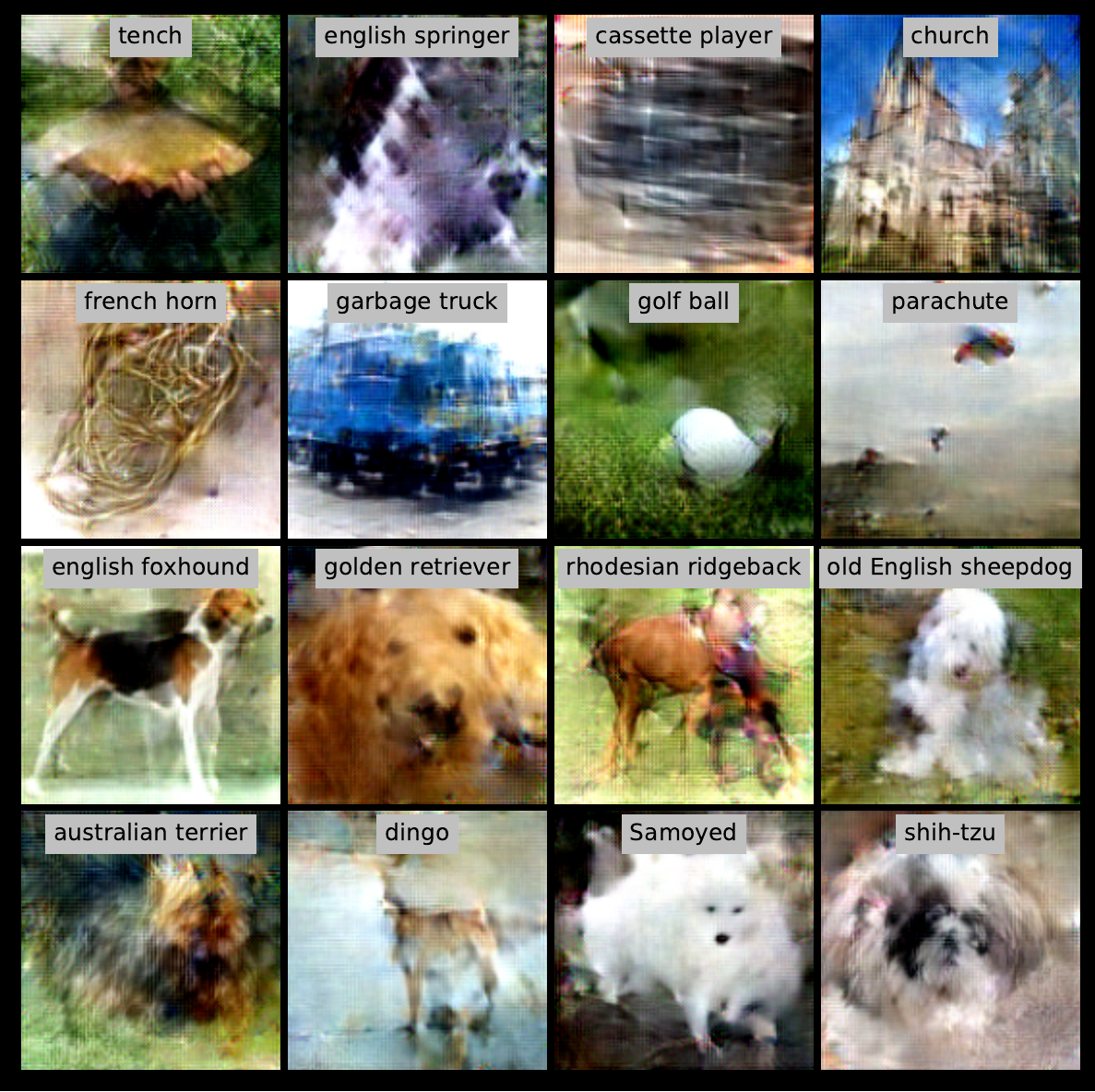}
  \caption{ImageNette and ImageWoof}
  \end{subfigure}
  \caption{Example distilled images from 32x32 CIFAR100, 64x64 Tiny ImageNet, and 128x128 ImageNet Subset. The images look real and transfer well to different architectures. They can be used for various downstream applications, such as continual learning and membership inference defense.}\label{fig:intro}
\end{figure}

\textbf{Summary of Contributions: } 
\begin{itemize}
    \item We propose an effective method for dataset distillation. Our method, named neural Feature Regression with Pooling (FRePo), achieves state-of-the-art results on various benchmark datasets with a 100x reduction in training time and a 10x reduction in GPU memory requirement. Our distilled data looks real (Figure \ref{fig:intro}) and transfers well to different architectures.
    \item We show that FRePo scales well to datasets with high-resolution images or complex label space. We achieve 7.5\% top1 accuracy on ImageNet-1K \cite{ILSVRC15} using only one image per class. The same classifier obtains only 1.1\% accuracy from a random subset of real images. The previous methods struggle in this task due to large memory and compute requirements.
    \JB{Why is 7.5\% and simple convnet impressive? a natural question the reader will have is why not using larger model and more compute and train longer? we should justify it by saying something like "where the previous methods would require a large amount of memory and compute that is not possible with the current hardware".}
    \item We demonstrate that high-quality distilled data can significantly improve various downstream applications, such as continual learning and membership inference defense.
\end{itemize}

\section{Method}
\subsection{Dataset Distillation as Bi-level Optimization}
Suppose we have a large labeled dataset $\targetdataset=\left\{\left(\mathbf{x}_{1}, \mathbf{y}_{1}\right), \dots, \left(\mathbf{x}_{|\targetdataset|}, \mathbf{y}_{|\targetdataset|}\right) \right\}$ with $|\targetdataset|$ image and label pairs. Dataset distillation aims to learn a small synthetic dataset $\supportdataset = \left\{\left(\mathbf{x}_{1}, \mathbf{y}_{1}\right), \dots, \left(\mathbf{x}_{|\supportdataset|}, \mathbf{y}_{|\supportdataset|}\right) \right\}$ that preserves most of the information in $\targetdataset$. We train several neural networks parameterized by $\theta$ on the dataset $\supportdataset$ and then compute the validation loss $\mathcal{L}(\optalg \left(\nnparam, \supportdataset \right), \targetdataset)$ on the real dataset $\targetdataset$, where $\optalg \left(\nnparam, \supportdataset \right)$ is the neural network parameters optimized by a learning algorithm $\optalg$ with the model initialization $\nnparam$ and distilled dataset $\supportdataset$ as its inputs. The validation loss $\mathcal{L}(\optalg \left(\nnparam, \supportdataset \right), \targetdataset)$ is a noisy objective with the stochasticity coming from random model initialization and inner learning algorithm. Thus, we are interested in minimizing the expected value of this loss, which we denote it as $F(\supportdataset)$. We formulate the dataset distillation as the following bi-level optimization problem.
\begin{equation}
    \overbrace{\supportdataset^{*}:=\argmin_{\supportdataset} {F(\supportdataset)} }^{\textit {outer-level}},\ \textrm { where } F(\supportdataset) = \mathbb{E}_{\nnparam \sim \modeldist} \bigg[ {\mathcal{L}\Big(\overbrace{\optalg \left(\nnparam, \supportdataset \right)}^{\textit {inner-level }},\ \targetdataset \Big)} \bigg].
\end{equation}
In this bi-level setup, the outer loop optimizes the distilled data to minimize $F(\supportdataset)$, while the inner loop trains a neural network using the learning algorithm, $\optalg$, to minimize the training loss on the distilled data $\supportdataset$. From the meta-learning perspective, the task is defined by the model initialization $\nnparam$, and we want to learn a meta-parameter $\supportdataset$ that generalizes well to different models sampled from the model distributions $\modeldist$. During learning, we optimize the meta-parameter $\supportdataset$ by minimizing the meta-training loss $F(\supportdataset)$. In contrast, at meta-test time, we train a new model from scratch on $\supportdataset$ and evaluate the trained model on a held-out real dataset. This meta-test performance reflects the quality of the distilled data.

\begin{figure}
  \centering
  \hspace*{-0.035\linewidth}
  \includegraphics[width=1.07\linewidth]{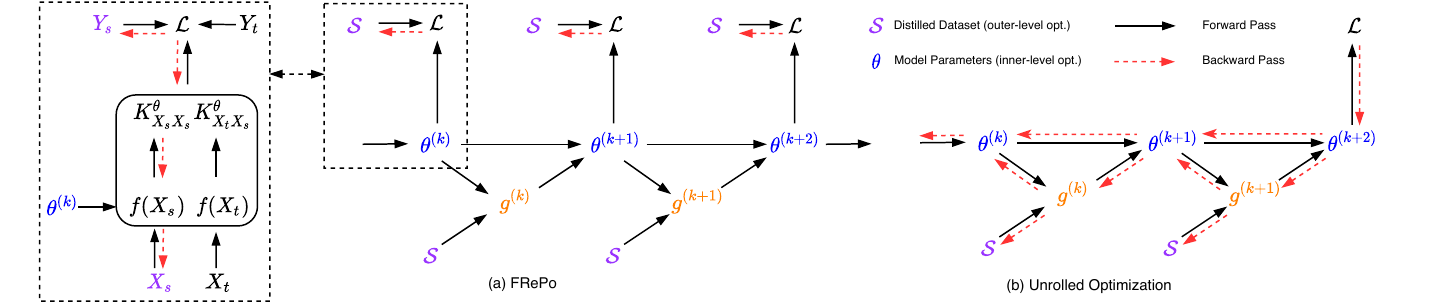}
  \caption{Comparison of \algname\ and Unrolled Optimization. $S$, $X_s$, $Y_s$ are the distilled dataset, images and labels. $\mathcal{L}$ is the meta-training loss and $\theta^{(k)}$, $g^{(k)}$ are the model parameter and gradient at step $k$. $f(X)$ is the feature for input $X$ and $\gram \targetx \supportx$ is the Gram matrix of $\targetx$ and $\supportx$. \algname\ is analogous to 1-step TBPTT as it computes the meta-gradient at each step while performing the online model update. However, instead of backpropagating through the inner optimization, \algname\ computes the meta-gradient through a kernel and feature extractor.}\label{fig:alg}
\end{figure}

\subsection{Dataset Distillation using Neural Feature Regression with Pooling (\algname)}
The outer-level problem can be solved using gradient-based methods of the form $\supportdataset \leftarrow \supportdataset - \ddlr\ \metagrad,$ where $\ddlr$ is the learning rate for the distilled data and $\metagrad$ is the meta-gradient \citep{DBLP:conf/nips/RajeswaranFKL19}. For a particular model $\nnparam$, the meta-gradient can be expressed as $\metagradsingle$. Computing this meta-gradient requires differentiating through inner optimization. If $\optalg$ is an iterative algorithm like gradient descent, then backpropagating through the unrolled computation graph \citep{DBLP:conf/icml/VicolMS21} can be a solution. However, this type of unrolled optimization introduces significant computation and memory overhead, as the whole training trajectory needs to be stored in memory (Figure~\ref{fig:alg}(b)).

Traditionally, these issues are alleviated with truncated backpropagation through time (TBPTT) \citep{Werbos1990BPTT, sutskever2013training, DBLP:journals/corr/TallecO17a}. Instead of backpropagating through an entire unrolled sequence, TBPTT performs backpropagation for each subsequence separately. It is efficient because its time and memory complexity scale linearly with respect to the truncation steps. However, truncation may yield highly biased gradients that severely impact training. To mitigate this truncation bias \citep{DBLP:conf/icml/VicolMS21}, we consider \textit{training only the top layer of a network to convergence}. The key insight is that the data helpful for training the output layer can also help train the whole network. Thus, we decompose the neural network into a feature extractor and a linear classifier. We fix the feature extractor at each meta-gradient computation and train the linear classifier to convergence before updating $\supportdataset$. After that, we adjust the feature extractor by training the whole network on the updated distilled data. We note that similar two-phase procedure has been studied in the context of representation learning \citep{ba2022high}. 

\textbf{Meta-Gradient Computation:} If we consider the mean square error loss, then the optimal weights for the linear classifier have a closed-form solution. Moreover, since the feature dimension is typically larger than the number of distilled data, we can use kernel ridge regression (KRR) with a conjugate kernel \citep{Neal1995BayesianLF} rather than solving the weights explicitly \citep{DBLP:journals/corr/BohdalFDD20}. The resulting meta-training loss (Eq. \ref{eqn:loss}) is similar to that used in KIP \citep{DBLP:conf/iclr/KIP1, DBLP:conf/nips/KIP2}, but we use a more flexible kernel rather than NTK.
\begin{equation}\label{eqn:loss}
    \targetloss \left( \optalg \left(\nnparam, \supportdataset \right), \targetdataset \right) = \frac{1}{2} ||\targety - \gram \targetx \supportx (\gram \supportx \supportx + \lambda I)^{-1} \supporty||_2^2,
\end{equation}
where $\targetdata$ and $\supportdata$ are the inputs and labels of the real data $\targetdataset$ and distilled data $\supportdataset$ respectively. The Gram matrix between real inputs and distilled inputs is denoted as $\gram \targetx \supportx \in \mathbb{R}^{|\targetdataset| \times |\supportdataset|}$, while the Gram matrix between distilled inputs is denoted as $\gram \supportx \supportx \in \mathbb{R}^{|\supportdataset| \times |\supportdataset|}$. $\lambda$ controls the regularization strength for KRR. Let us denote the neural network feature for a given input $X$ and model parameter $\theta$ as $f(X,\theta) \in \mathbb{R}^{N \times d}$, where $N$ is the number of input and $d$ is the feature dimension \footnote{In practice, we use all the synthetic data and sample a minibatch from the real dataset to compute the meta-gradient (Algorithm \ref{alg:frepo}).}. The conjugate kernel is defined by the inner product of the neural network features. Thus, the two Gram matrices are computed as follows:
\begin{equation}\label{eqn:gram}
    \gram \targetx \supportx = f(\targetx, \nnparam)f(\supportx, \nnparam)^{\top},\quad \gram \supportx \supportx = f(\supportx, \nnparam)f(\supportx, \nnparam)^{\top},
\end{equation}
Now, computing the meta-gradient $\metagradsingle$ is just back-propagating through the conjugate kernel and a fixed feature extractor, which is very efficient and takes even fewer operations than computing the gradient for the network's weights. Moreover, we decouple the meta-gradient computation from the model online update. Hence, we can train the online model using any optimizer, and the distilled data will be agnostic to the specific learning algorithm choice. Our proposed method is similar to 1-step TBPTT in that we compute the meta-gradient at each step while performing the online model update. Unlike the conventional 1-step TBPTT, we compute the meta-gradient using a KRR output layer to mitigate truncation bias, illustrated in Figure~\ref{fig:alg}(a).

\begin{algorithm}[tb]
    \caption{Dataset Distillation using Neural Feature Regression with Pooling (\algname)}
    \label{alg:frepo}
    \begin{algorithmic}
        \STATE {\bfseries Require:} $\targetdataset$: a labeled dataset; $\ddlr$: the learning rate for the distilled data
        \STATE {\bfseries Initialization:} Initialize a labeled distilled dataset $\supportdataset=\supportdata$.
        \STATE {\bfseries Initialization:} Initialize a model pool $\modelpool$ with $\numpool$ models ${\{\nnparam_{i}\}}_{i=1}^{\numpool}$ randomly initialized from $\modeldist$. \JB{We overloaded the letter P here. Can we just call model pool $Pool={\{\nnparam_{i}\}}_{i=1}^{\numpool}$? Answer: Change to letter M.}
        \JB{mention K?}
    \end{algorithmic}
    \begin{algorithmic}[1]
        \WHILE{not converged}
        \STATE $\smalltriangleright$ Sample a model uniformly from the model pool: $\theta_{i} \sim \modelpool$. \JB{How do we sample? Uniform? Answer: Uniformly sample}
        \STATE $\smalltriangleright$ Sample a target batch uniformly from the labeled dataset: $\targetdata \sim \targetdataset$.
        \STATE $\smalltriangleright$ Compute the meta-training loss $\targetloss$ using Eq. \ref{eqn:loss}
        \STATE $\smalltriangleright$ Update the distilled data $\supportdataset$: $\supportx \leftarrow \supportx - \ddlr \grad \supportx \targetloss$, and $\supporty \leftarrow \supporty - \ddlr \grad \supporty \targetloss$
        \STATE $\smalltriangleright$ Train the model $\theta_i$ on the current distilled data $\supportdataset$ for one step. 
        \JB{Can we put this step right after sampling the model as the new step 2? This step is equivalent of the inner-level step should appear first. and how many steps are we training? Answer: This is my implementation 1 month ago. I guess we may want to use the old parameter to compute the meta gradient and train on the new distilled data. The idea is that we find a good data to train the last layer, so we can also use it to train the whole network. However, empirically, we find no differences between those two implementations}
        \STATE $\smalltriangleright$ Reinitialize the model $\theta_i \sim \modeldist$ if $\theta_i$ has been updated more than $\maxonlineupdate$ steps.
        \ENDWHILE
    \end{algorithmic}
    \begin{algorithmic}
        \STATE {\bfseries Output:} Learned distilled dataset $\supportdataset=\supportdata$
    \end{algorithmic}
\end{algorithm}

\textbf{Model Pool:} As discussed in Section \ref{sec:intro}, there are various types of overfitting in dataset distillation. \JB{It is ok to call our own method a trick but probably not the previous methods no matter how simple they are. I changed all the ``trick'' to something else. Answer: I see} Several techniques have been proposed to alleviate such problem, such as random initialization \citep{DBLP:journals/corr/WangDD18}, periodic reset \citep{DBLP:journals/corr/SucholutskyIDD19, DBLP:conf/iclr/DC, DBLP:conf/icml/DSA}, and dynamic bi-level optimization \citep{DBLP:journals/corr/CAFE}. These techniques share the same underlying principle: the model diversity matters. Thus, we propose to \textit{maintain a ``model pool'' filled with diverse set of parameters} obtained from different number of training steps and different random initializations. \JB{Maybe put italic style over the previous sentence as it is one of the two core ideas in the paper to address the previous challenges. Answer: Sure} Unlike the previous methods that periodically training and resetting a single model, \algname\, randomly sample a model from the pool at each meta-gradient computation and update it using the current distilled data. However, if a model has been updated more than $\maxonlineupdate$ steps, we reinitialize it with a new random seed. From the meta-learning perspective, we maintain a diverse set of meta-tasks to sample from and avoid sampling very similar tasks at each consecutive gradient computation to avoid overfitting to a particular setup.

\textbf{Pool Diversity:} We can increase the regularization strength by increasing the diversity of the model pool by setting a larger $\maxonlineupdate$, using data augmentation when training the model on the distilled data, or using models with different architectures. To keep our method simple, we use the same architecture for all models in the pool and do not use any data augmentation when training the model on the distilled data. Thus, our model pool only contains models with different initialization, at different optimization stages, and trained at different time-step of the distilled data. 
\section{Related Work}
\textbf{Unrolling in Bi-Level Optimization:}
One way to compute the meta-gradient is to differentiate through the unrolled inner optimization \citep{DBLP:journals/corr/WangDD18, DBLP:conf/icml/MaclaurinDA15, DBLP:journals/corr/BohdalFDD20, DBLP:journals/corr/SucholutskyIDD19}. However, this approach inherits several difficulties of the unrolled optimization, such as: 1) large computation and memory cost \citep{DBLP:conf/icml/VicolMS21}; 2) truncation bias with short unrolls \citep{DBLP:conf/iclr/WuRLG18}; 3) exploding or vanishing gradients with long unrolls \citep{DBLP:conf/icml/PascanuMB13}; 4) chaotic and poorly conditioned loss landscapes with long unrolls \citep{DBLP:conf/icml/MetzMNFS19}. In contrast, our method considers approximating the inner optimization with kernel ridge regression instead of unrolled optimization. \JB{The readers still do not know about our method at this point. Maybe move the entire related works section after the method}

\textbf{Surrogate Objective:}
To avoid unrolled optimization, several works turn to surrogate objectives. DC \citep{DBLP:conf/iclr/DC}, DSA \citep{DBLP:conf/icml/DSA}, and DCC \citep{DBLP:journals/corr/DCC} formulate the dataset distillation as a gradient matching problem between the gradients of neural network weights computed on the real and distilled data. In contrast, DM \citep{DBLP:journals/corr/DM} and CAFE \citep{DBLP:journals/corr/CAFE} consider the feature distribution alignment between the real and distilled data. Moreover, MTT \citep{DBLP:journals/corr/MTT} shows that knowledge from many expert training trajectories can be distilled to a dataset by using a training trajectory matching objective. Nevertheless, surrogate objectives may introduce new biases and thus, may not accurately reflect the true objective. For example, gradient matching approaches \citep{DBLP:conf/iclr/DC, DBLP:conf/icml/DSA, DBLP:journals/corr/DCC} only focus on short-range behavior and may easily overfit to a biased set of samples that produce dominant gradients \citep{DBLP:journals/corr/CAFE, DBLP:journals/corr/MTT}.

\textbf{Closed-form Approximation:}
An alternative way to circumvent unrolled optimization is to find a closed-form approximation to the inner optimization. Based on the correspondence between infinitely-wide neural networks and kernel methods, KIP \citep{DBLP:conf/iclr/KIP1,DBLP:conf/nips/KIP2} approximates the inner optimization with NTK \citep{DBLP:conf/nips/LeeXSBNSP19}. In this case, the meta-gradient can be computed by back-propagating through the NTK. However, computing NTK for modern neural networks is extremely expensive. Thus, using NTK for dataset distillation requires thousands of GPU hours and sophisticated implementation of the distributed kernel computation framework \citep{DBLP:conf/nips/KIP2}. 
Similar to ours, \citet{DBLP:journals/corr/BohdalFDD20} also decomposes the neural network as a feature extractor and a linear classifier. However, they only learn the label and explicitly solve for the optimal classifier weights rather than perform KRR.
\section{Dataset Distillation}
\subsection{Implementation Details}
We compare our method to four state-of-the-art dataset distillation methods \citep{DBLP:conf/icml/DSA,DBLP:journals/corr/DM, DBLP:journals/corr/MTT, DBLP:conf/nips/KIP2} on various benchmark datasets \citep{ILSVRC15, DBLP:journals/pieee/LeCunBBH98, DBLP:journals/corr/abs-1708-07747, Krizhevsky09learningmultiple, Le2015TinyIV, imagenette, cub200}. We train the distilled data using Algorithm \ref{alg:frepo} with the same set of hyperparameters for all experiments except stated otherwise.
Unlike prior work \citep{DBLP:conf/icml/DSA, DBLP:journals/corr/DM, DBLP:journals/corr/MTT}, we do not apply data augmentation during training. However, we apply the same data augmentation \citep{DBLP:conf/icml/DSA, DBLP:journals/corr/MTT} during evaluation for a fair comparison.
We preprocess the data in a similar way as in previous works \citep{DBLP:journals/corr/MTT, DBLP:conf/nips/KIP2} but use a wider architecture than previous works \citep{DBLP:conf/icml/DSA,DBLP:journals/corr/DM, DBLP:journals/corr/MTT} because the KRR component does not behave well when the feature dimension is low, resulting in a significant performance drop for our method. Results on the original architecture are included in Appendix \ref{as:model}. We evaluate each distilled data using five random neural networks and report the mean and standard deviation. For the baseline method, we report the best of the reported value in the original paper and our reproducing results. 

For the sake of brevity, we provide implementation details about data preprocessing, distilled data initialization, and hyperparameters in Appendix \ref{app:imple} and various ablation studies regarding the model pool, batch size, distilled data initialization, label learning, and model architectures in Appendix \ref{as:title}. More distilled image visualizations can be found in Appendix \ref{app:addvis}. Our code is available at \url{https://github.com/yongchao97/FRePo}.

\subsection{Standard Benchmarks} \label{sec:benchmark}

\begin{figure}[t]
  \centering
\begin{subfigure}[b]{0.235\textwidth}
  \includegraphics[width=1.0\linewidth]{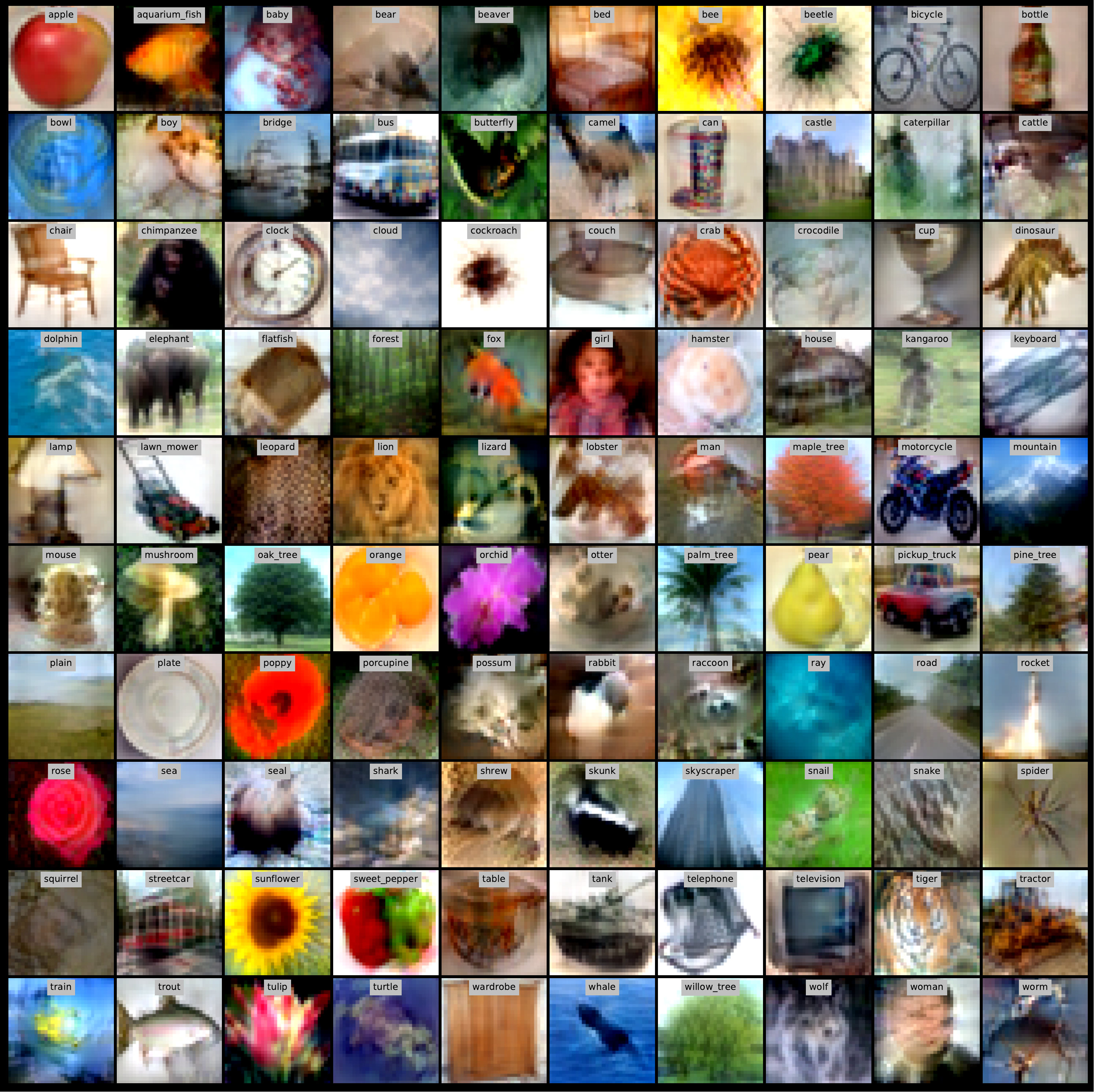}  
  \caption{\algname} \label{fig:vis_frepo}
\end{subfigure}
\begin{subfigure}[b]{0.235\textwidth}
  \includegraphics[width=1.0\linewidth]{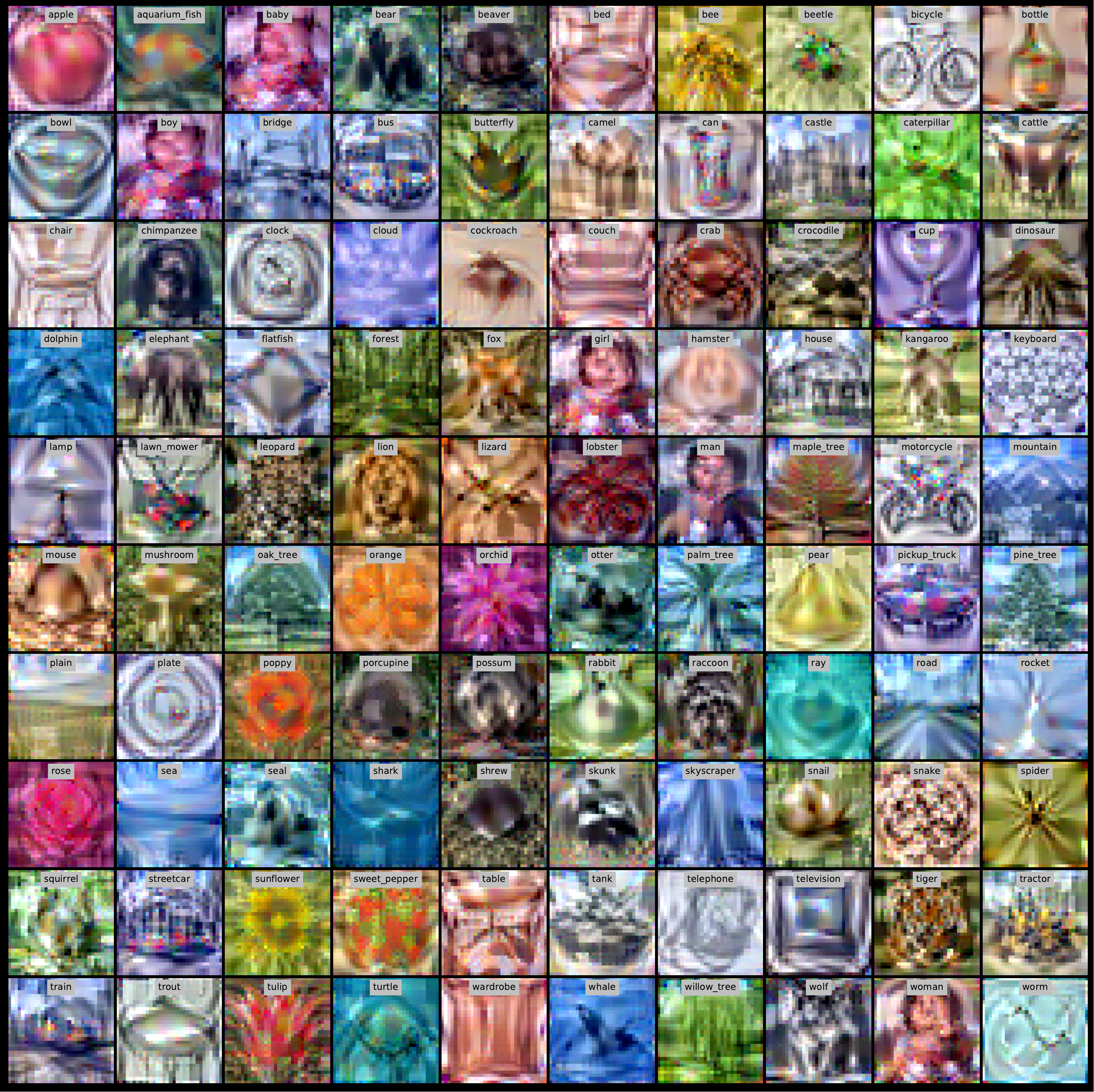}
  \caption{MTT \citep{DBLP:journals/corr/MTT}} \label{fig:vis_mtt}
 \end{subfigure}
 \begin{subfigure}[b]{0.235\textwidth}
  \includegraphics[width=1.0\linewidth]{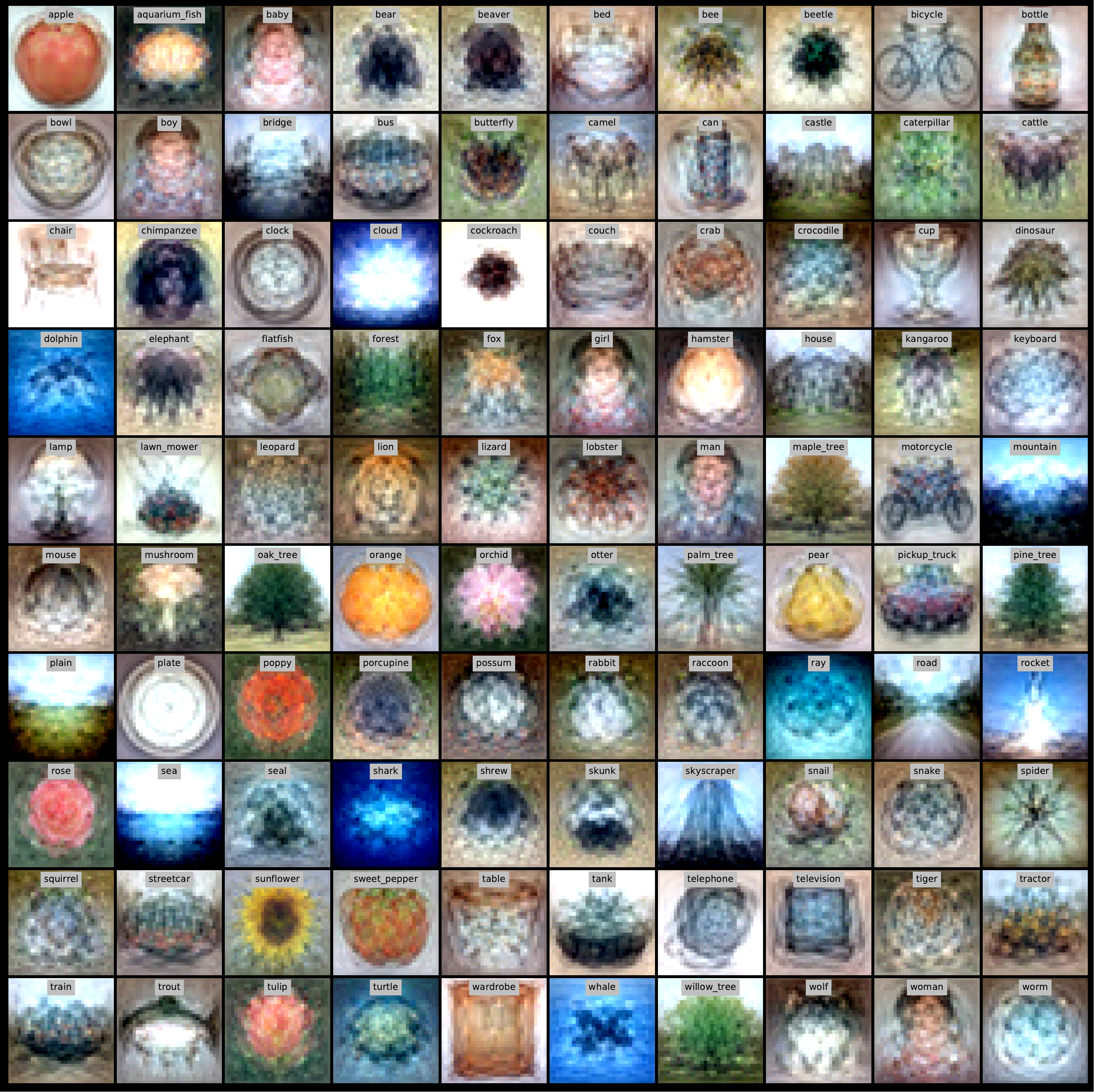}
  \caption{DSA \citep{DBLP:conf/icml/DSA}} \label{fig:vis_dsa}
  \end{subfigure}
  \begin{subfigure}[b]{0.272\textwidth}
  \includegraphics[width=1.0\linewidth]{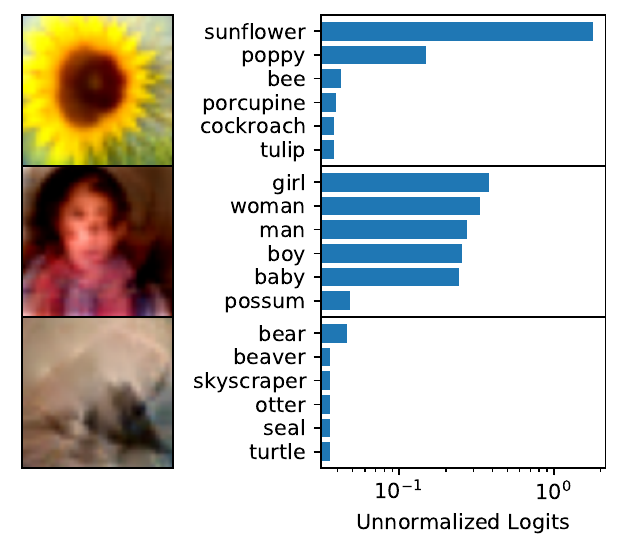}
  \caption{Label Visualization}\label{fig:lb-analysis}
  \end{subfigure}
  \vspace{-0.1in}
  \caption{(a,b,c) Distilled 1 img/cls from CIFAR100 using \algname, MTT, and DSA. High quality images also produce high test accuracy. (d) Three categories of learned labels. (Top) High confidence, large margin;  (Middle) High confidence, small margin; (Bottom) Low confidence, small margin. }\label{fig:vis-compare}
\end{figure}

\textbf{Distillation Performance:} We first evaluate our method on six standard benchmark datasets. We learn 1, 10, and 50 images per class for datasets with only ten classes, while we learn 1 and 10 images per class for CIFAR100 \citep{Krizhevsky09learningmultiple} with 100 classes, Tiny ImageNet \citep{Le2015TinyIV} with 200 classes, and CUB-200 \citep{cub200} with 200 fine-grained classes. As shown in Table in \ref{tab:dd_sota}, we achieve the state-of-the-art performance in most settings despite the hyperparameter may be suboptimal. Our method performs exceptionally well on datasets with a complex label space when learning few images per class. For example, we improve the CIFAR100, Tiny ImageNet, and CUB-200 in one image per class setting from 24.3\%, 8.8\%, and 2.2\% to 28.7\%, 15.4\%, and 12.4\%, respectively. Figure \ref{fig:vis-compare} shows that our distilled images look real and natural though we do not directly optimize for this objective. We observe a strong correlation between the test accuracy and image quality: the better the image quality, the higher the test accuracy. Our results suggest that a highly condensed dataset does not need to be very different from the real dataset as it may just reflect the most common pattern in a dataset. We also report the KRR predictor's test accuracy using the feature extractor trained on the distilled data. When the dataset is as simple as MNIST \citep{DBLP:journals/pieee/LeCunBBH98}, the KRR predictor achieves similar performance as the neural network predictor. In contrast, for more complex datasets, the KRR predictor consistently outperforms the neural network predictor, with the most significant gap being 3.7\% for Tiny ImageNet in the one image per class setting.

\begin{table}[t]
  \caption{Test accuracies of models trained on the distilled data from scratch. 
  $^\dag$ denotes performance better than the original reported performance. KRR preformance is shown in bracket. \algname~performs extremely well for one image per class setting on CIFAR100, Tiny ImageNet and CUB-200.}
  \label{tab:dd_sota}
  \small
  \centering
    \begin{tabular}{cc|cccc|c}
        \toprule
        & Img/Cls & DSA \cite{DBLP:conf/icml/DSA} & DM \cite{DBLP:journals/corr/DM}& KIP \cite{DBLP:conf/nips/KIP2} & MTT \cite{DBLP:journals/corr/MTT} & \algname \\
        \midrule
        \multirow{3}{*}{MNIST} & 1 & $88.7 \pm 0.6$ & $89.9 \pm 0.8$\rlap{$^\dag$} & $90.1 \pm 0.1 $ & $91.4 \pm 0.9$\rlap{$^\dag$} & $\textbf{93.0} \pm \textbf{0.4}$ $(92.6 \pm 0.4)$\\
         & 10 & $97.9 \pm 0.1$\rlap{$^\dag$}  & $97.6 \pm 0.1$\rlap{$^\dag$} & $97.5\pm 0.0$ & $97.3 \pm 0.1$\rlap{$^\dag$} & $ \textbf{98.6}\pm \textbf{0.1} $ $(98.6 \pm 0.1)$\\
         & 50 & $\textbf{99.2} \pm \textbf{0.1}$ & $98.6 \pm 0.1$ & $98.3\pm 0.1$ & $98.5\pm 0.1$\rlap{$^\dag$} & $ \textbf{99.2} \pm \textbf{0.0}$ $(99.2 \pm 0.1)$\\
        \midrule
        \multirow{3}{*}{F-MNIST} & 1 & $ 70.6 \pm 0.6$ & $ 71.5 \pm 0.5$\rlap{$^\dag$} & $ 73.5 \pm 0.5$ & $75.1 \pm 0.9$\rlap{$^\dag$} & $\textbf{75.6} \pm \textbf{0.3}$ $(77.1 \pm 0.2)$\\
         & 10 & $ 84.8\pm 0.3$\rlap{$^\dag$} & $ 83.6 \pm 0.2$\rlap{$^\dag$} & $ 86.8 \pm 0.1$ & $ \textbf{87.2} \pm \textbf{0.3}$\rlap{$^\dag$}  & $86.2 \pm 0.2 $ $(86.8 \pm 0.1)$\\
         & 50 & $ 88.8\pm 0.2$\rlap{$^\dag$} & $ 88.2 \pm 0.1$\rlap{$^\dag$} & $ 88.0 \pm 0.1$ & $88.3\pm 0.1$\rlap{$^\dag$} & $ \textbf{89.6}\pm \textbf{0.1}$ $(89.9 \pm 0.1)$\\
        \midrule
        \multirow{3}{*}{CIFAR10} & 1 & $36.7 \pm 0.8$\rlap{$^\dag$} & $31.0 \pm 0.6$\rlap{$^\dag$} & $ \textbf{49.9}\pm \textbf{0.2} $ & $ 46.3\pm 0.8 $ & $ 46.8 \pm 0.7 $ $(47.9 \pm 0.6)$\\
         & 10 & $53.2 \pm 0.8$\rlap{$^\dag$} & $49.2 \pm 0.8$\rlap{$^\dag$} & $ 62.7\pm 0.3 $ & $ 65.3 \pm 0.7 $ & $ \textbf{65.5}\pm \textbf{0.4}$ $(68.0 \pm 0.2) $ \\
         & 50 & $66.8 \pm 0.4$\rlap{$^\dag$} & $63.7 \pm 0.5$\rlap{$^\dag$} & $ 68.6 \pm 0.2 $ & $ 71.6\pm 0.2 $ & $ \textbf{71.7}\pm \textbf{0.2}$ $ (74.4 \pm 0.1)$\\
        \midrule
        \multirow{3}{*}{CIFAR100} & 1 & $16.8 \pm 0.2$\rlap{$^\dag$} & $12.2 \pm 0.4$\rlap{$^\dag$} & $ 15.7\pm 0.2 $ & $ 24.3 \pm 0.3 $ & $ \textbf{28.7}\pm \textbf{0.1}$ $(32.3 \pm 0.1)$\\
         & 10 & $ 32.3 \pm 0.3$ & $ 29.7 \pm 0.3$ & $ 28.3\pm 0.1 $ & $ 40.1 \pm 0.4 $ & $ \textbf{42.5}\pm \textbf{0.2}$ $(44.9 \pm 0.2)$\\
         & 50 & $ 42.8 \pm 0.4$ & $ 43.6 \pm 0.4$ & $-$ & $\textbf{47.7} \pm \textbf{0.2}$ & $ 44.3\pm 0.2$ $(43.0 \pm 0.3)$\\
        \midrule
        \multirow{2}{*}{T-ImageNet} & 1 & $ 6.6 \pm 0.2$\rlap{$^\dag$} & $3.9 \pm 0.2$ & $ - $ & $ 8.8\pm 0.3 $ & $ \textbf{15.4}\pm \textbf{0.3}$ $(19.1 \pm 0.3)$\\
         & 10 & $-$ & $ 12.9 \pm 0.4$ &$-$ & $ 23.2\pm 0.2 $ & $\textbf{25.4} \pm \textbf{0.2}$ $(26.5 \pm 0.1)$\\
        \midrule
        \multirow{2}{*}{CUB-200} & 1 & $ 1.3 \pm 0.1$\rlap{$^\dag$} & $1.6 \pm 0.1$\rlap{$^\dag$} & $ - $ & $2.2\pm 0.1 $\rlap{$^\dag$} & $ \textbf{12.4}\pm \textbf{0.2}$ $(13.7 \pm 0.2)$\\
         & 10 & $ 4.5 \pm 0.3$\rlap{$^\dag$} & $4.4 \pm 0.2$\rlap{$^\dag$} & $-$ & $-$ & $\textbf{16.8} \pm \textbf{0.1}$ $(16.1 \pm 0.3)$\\
        \bottomrule
    \end{tabular}
\end{table}

\begin{figure}
  \centering
\begin{subfigure}[b]{0.245\textwidth}
 \includegraphics[width=1.0\linewidth]{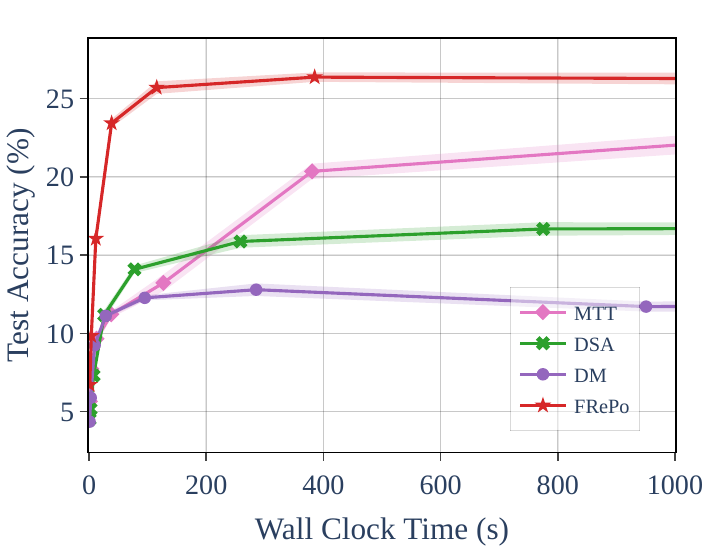}
 \caption{Test Acc }\label{fig:complexity_acc_time_300}
\end{subfigure}
\begin{subfigure}[b]{0.245\textwidth}
 \includegraphics[width=1.0\linewidth]{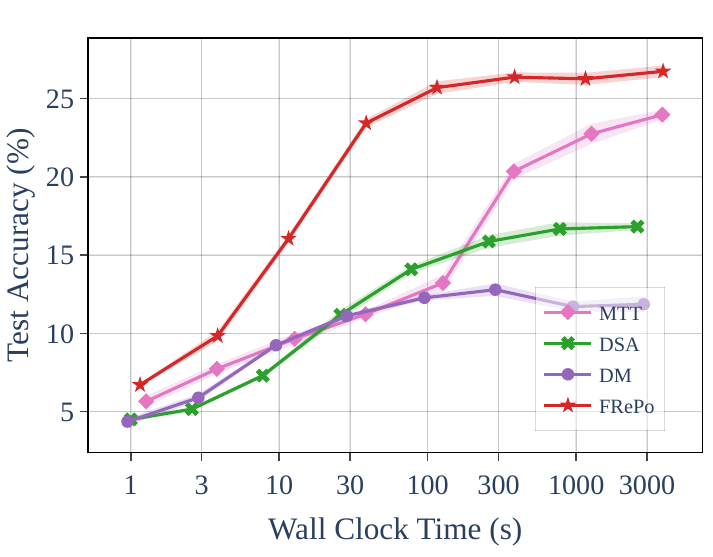}  
 \caption{Test Acc (Log Scale)}\label{fig:complexity_acc_time}
\end{subfigure}
\begin{subfigure}[b]{0.245\textwidth}
 \includegraphics[width=1.0\linewidth]{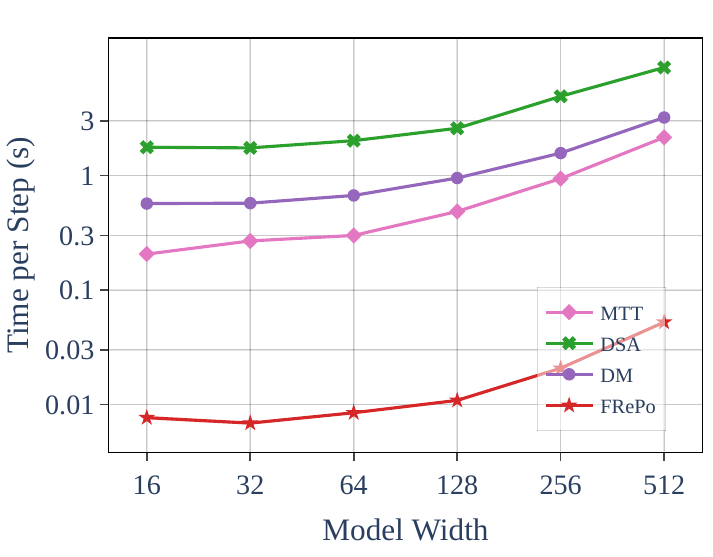}
 \caption{Time Per Step}\label{fig:complexity_timeperstep}
\end{subfigure}
\begin{subfigure}[b]{0.245\textwidth}
 \includegraphics[width=1.0\linewidth]{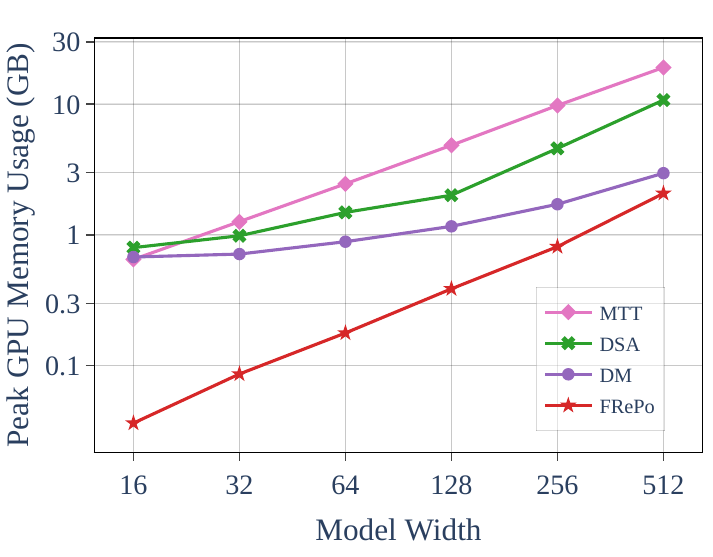}
 \caption{GPU Memory Usage}\label{fig:complexity_memory}
\end{subfigure}
\caption{(a,b) Training efficiency comparison when learning 1 Img/Cls on CIFAR100. (c,d) Time per iteration and peak memory usage as we increase the model size. 
\algname~is significantly more efficient than the previous methods, almost two orders of magnitude faster than the second-best method (i.e., MTT), with only 1/10 of the GPU memory requirement. 
}\label{fig:complexity}
\end{figure}

\begin{table}[t]
  \caption{Cross-architecture transfer performance on CIFAR10 with 10 Img/Cls. Despite being trained for a specific architecture, our distilled data transfer well to various architectures unseen during training. Conv is the default evaluation model used for each method. NN, DN, IN, and BN stand for no normalization, default normalization, Instance Normalization, Batch Normalization respectively.}
  \label{tab:dd_ca}
  \small
  \centering
    \begin{tabular}{cccccccc}
        \toprule
        \multirow{2}{*}[-4pt]{} & \multirow{2}{*}[-4pt]{Train Arch} & \multicolumn{6}{c}{Evaluation Architecture}\\
        \cmidrule(l){3-8}
        && Conv & Conv-NN & ResNet-DN & ResNet-BN & VGG-BN & AlexNet \\
        \midrule
        DSA \citep{DBLP:conf/icml/DSA} &  Conv-IN & $53.2 \pm 0.8$ & $36.4 \pm 1.5$ & $42.1 \pm 0.7 $ & $ 34.1 \pm 1.4$ & $ 46.3 \pm 1.3$ & $ 34.0 \pm 2.3$ \\
        DM \citep{DBLP:journals/corr/DM} &  Conv-IN & $49.2 \pm 0.8$ & $35.2 \pm 0.5$ & $36.8 \pm 1.2 $ & $ 35.5 \pm 1.3$ & $ 41.2 \pm 1.8$ & $ 34.9 \pm 1.1$ \\
        MTT \citep{DBLP:journals/corr/MTT} &  Conv-IN & $64.4 \pm 0.9$ & $41.6 \pm 1.3$ & $49.2 \pm 1.1 $ & $ 42.9 \pm 1.5$ & $ 46.6 \pm 2.0$ & $ 34.2 \pm 2.6$ \\
        KIP \citep{DBLP:conf/nips/KIP2} &  Conv-NTK & $62.7 \pm 0.3$ & $58.2 \pm 0.4$ & $49.0 \pm 1.2 $ & $ 45.8 \pm 1.4$ & $ 30.1 \pm 1.5$ & $ 57.2 \pm 0.4$ \\
        \algname & Conv-BN & $ \textbf{65.5} \pm \textbf{0.4}$ & $ \textbf{65.5} \pm \textbf{0.4}$ & $ \textbf{58.1} \pm \textbf{0.6}$ & $ \textbf{57.7} \pm \textbf{0.7}$ & $\textbf{59.4} \pm \textbf{0.7}$ & $\textbf{61.9} \pm \textbf{0.7} $\\
        \bottomrule
    \end{tabular}
\end{table}

\textbf{Label Learning:} A similar trend can also be observed for label learning. When the dataset is simple and has only a few classes, label learning may not be necessary. However, it becomes crucial for complex datasets with many labels, such as CIFAR100 and Tiny-ImageNet (See more details in Appendix \ref{as:labellearning}). Similar to the teacher label in the knowledge distillation \citep{DBLP:journals/corr/HintonVD15}, we observe that the distilled label also encodes the class similarity. We identify three typical cases in Figure \ref{fig:lb-analysis}. The first group consists of highly confident labels with a much higher value for one class than other classes (large margin), such as sunflower, bicycle, and chair. In contrast, the distilled labels in the second group are confident but may get confused with some closely-related classes (small margin). For instance, the learned label for "girl" has almost equally high values for the girl, woman, man, boy, and baby, suggesting that these classes are very similar and may be difficult for the model to distinguish them apart. The last group contains distilled labels with low values for all classes, such as bear, beaver, and squirrel. It is often hard for humans to recognize the distilled images in such a group, suggesting that they may be the challenging classes in a dataset.

\textbf{Training Cost Analysis:} Figure \ref{fig:complexity_acc_time_300}, \ref{fig:complexity_acc_time} shows that our method is significantly more time-efficient than the previous methods. When learning one image per class on CIFAR100, \algname\ reaches a similar test accuracy (23.4\%) to the second-best method (24.0\%) in 38 seconds, compared to 3805 seconds for MTT, which is roughly two orders of magnitude faster. Moreover, \algname\ achieves 92\% of its final test accuracy (26.4\% out of 28.7\%) in only 385 seconds. As shown in Figure \ref{fig:complexity_timeperstep}, our algorithm takes much less time to perform one gradient step on the distilled data. Thus, we can perform more gradient steps in a fixed time. Furthermore, Figure \ref{fig:complexity_memory} suggests that our algorithm has much less GPU memory requirement. Therefore, we can potentially use a much larger and more complex model to take advantage of the advancement in neural network architecture. 

\textbf{Cross-Architecture Generalization:}
One desired property of our distilled data is that it generalizes well to architecture it has not seen during the training. 
Similar to previous works \cite{DBLP:conf/iclr/DC, DBLP:journals/corr/MTT}, we evaluate the distilled data from CIFAR10 on a wide range of architectures which it has not seen during training, including AlexNet \cite{DBLP:conf/nips/AlexNet}, VGG \cite{DBLP:journals/corr/VGG}, and ResNet \cite{DBLP:conf/cvpr/ResNet}. Table \ref{tab:dd_ca} shows that our method outperforms previous methods on all unseen architectures. Instance Normalization (IN) \cite{DBLP:journals/corr/UlyanovVL16}, as the vital ingredient in several methods (DSA, DM, MTT), seems to hurt the cross-architecture transfer. The performance degrades a lot when no normalization (NN) is applied (Conv-NN, AlexNet) or using a different normalization, like Batch Normalization (BN) \cite{DBLP:conf/icml/IoffeS15}. 
It suggests that the distilled data generated by those methods encode the inductive bias of a particular training architecture. In contrast, our distilled data generalize well to various architectures, including those without normalization (Conv-NN, AlexNet). Note that Figure \ref{fig:intro}, \ref{fig:vis-compare} also indicate that our distilled data encode less architectural bias as the distilled images look natural and authentic. A simple idea to further alleviate the overfitting of a particular architecture is to include more architectures in the model pool. However, the training may not be stable as the meta-gradient computed by different architectures can be very different. 

\subsection{ImageNet}
\textbf{High Resolution ImageNet Subset} To understand how well our method performs on high-resolution images, we evaluate it on ImageNette and ImageWoof datasets \citep{imagenette} with a resolution of 128x128. We learn 1 and 10 images per class on both datasets and report the performance in Table \ref{tab:imagenette} and visualize some distilled images in Figure \ref{fig:intro}. As shown in Table \ref{tab:imagenette}, we outperform MTT on all settings and achieve much better performance when we distill ten images per class on a more difficult dataset ImageWoof. It suggests that our distilled data is better at capturing the discriminative features for each class. Figure \ref{fig:intro} shows that our distilled images look real and capture the distinguishable feature of different classes. For the easy dataset (i.e., ImageNette), all images have clear different structures, while for ImageWoof, the texture of each dog seems to be crucial.

\textbf{Resized ImageNet-1K:} We also evaluate our method on a resized version of ILSVRC2012 \cite{ILSVRC15} with a resolution of 64x64 to see how it performs on a complex label space. Surprisingly, we can achieve 7.5\% and 9.7\% Top1 accuracy using only 1k and 2k training examples, compared to 1.1\% and 1.4\% using an equally-sized real subset. 

\begin{table}
  \caption{Distillation performance on higher resolution (128x128) dataset (i.e. ImageNette, ImageWoof) and medium resolution (64x64) dataset with a complex label space (i.e. ImageNet-1K). \algname\ scales to high-resolution images and learns the discriminate feature of complex datasets.}
  \label{tab:imagenette}
  \small
  \centering
    \begin{tabular}{ccccccc}
        \toprule
        \multirow{2}{*}[-4pt]{} &
        \multicolumn{2}{c}{ImageNette (128x128)} & \multicolumn{2}{c}{ImageWoof (128x128)} & \multicolumn{2}{c}{ImageNet (64x64)}\\
        \cmidrule(l){2-3} \cmidrule(l){4-5} \cmidrule(l){6-7}
        Img/Cls & 1 & 10 & 1 & 10 & 1 & 2\\
        \midrule
        Random Subset & $23.5 \pm 4.8$ & $47.7 \pm 2.4$ & $14.2 \pm 0.9$ & $27.0 \pm 1.9$ & $1.1 \pm 0.1$ & $1.4 \pm 0.1$ \\
        MTT~\citep{DBLP:journals/corr/MTT} & $47.7 \pm 0.9$ & $63.0 \pm 1.3$ & $ 28.6 \pm 0.8 $ & $35.8 \pm 1.8$ & $ - $ & $ - $ \\
        \midrule
        \algname & $ \textbf{48.1} \pm \textbf{0.7}$ & $ \textbf{66.5} \pm \textbf{0.8}$ & $ \textbf{29.7} \pm \textbf{0.6} $ & $ \textbf{42.2} \pm \textbf{0.9}$ & $ \textbf{7.5} \pm \textbf{0.3}$ & $\textbf{9.7} \pm \textbf{0.2}$ \\
        \bottomrule
        \vspace{-0.3in}
    \end{tabular}
\end{table}
\section{Application}
\subsection{Continual Learning}\label{sec:cl}
Continual learning (CL) \citep{DBLP:journals/corr/KirkpatrickPRVD16} aims to address the catastrophic forgetting problem \cite{DBLP:journals/corr/KirkpatrickPRVD16, french99cf, DBLP:journals/connection/Robins95} when a model learns sequentially from a stream of tasks. A commonly used strategy to recall past knowledge is based on a replay buffer, which stores representative samples from previous tasks \citep{DBLP:conf/icpr/BuzzegaBPC20,  DBLP:conf/cvpr/LiuSS21, DBLP:journals/corr/RebuffiKL16, DBLP:conf/eccv/PrabhuTD20}. Since sample selection is an important component of constructing an effective buffer \citep{DBLP:journals/corr/RebuffiKL16, DBLP:conf/eccv/PrabhuTD20, DBLP:conf/nips/AljundiLGB19, DBLP:conf/nips/AljundiBTCCLP19}, we believe distilled data can be a key ingredient for a continual learning algorithm due to its highly condensed nature. Several works \citep{DBLP:journals/corr/abs-2103-15851, DBLP:conf/icml/DSA, DBLP:journals/corr/DM,  DBLP:conf/cvpr/LiuSLSS20} have successfully applied the dataset distillation to the continual learning scenario. Our work shows that we can achieve much better results by using a better dataset distillation technique. 

We follow \citet{DBLP:journals/corr/DM} that sets up the baseline based on GDumb \citep{DBLP:conf/eccv/PrabhuTD20} which greedily stores class-balanced training examples in memory and train model from scratch on the latest memory only. In that case, the continual learning performance only depends on the quality of the replay buffer. We perform 5 and 10 step class-incremental learning \citep{DBLP:journals/corr/abs-1904-07734} on CIFAR100 with an increasing buffer size of 20 images per class. Specifically, we distill 400 and 200 images at each step and put them into the replay buffer. We follow the same class split as \citet{DBLP:journals/corr/DM} and compare our method to random \citep{DBLP:conf/eccv/PrabhuTD20}, herding \citep{DBLP:conf/eccv/CastroMGSA18, DBLP:conf/uai/ChenWS10}, DSA \citep{DBLP:conf/icml/DSA}, and DM \citep{DBLP:journals/corr/DM}. We use the default data preprocessing and default model for each method in this experiment as we find it gives the best performance for each method. We use the test accuracy on all observed classes as the performance measure \cite{DBLP:journals/corr/DM, DBLP:journals/corr/RebuffiKL16}.

\begin{figure}[t!]
\centering
\begin{subfigure}[b]{0.24\textwidth}
  \includegraphics[width=1.0\linewidth]{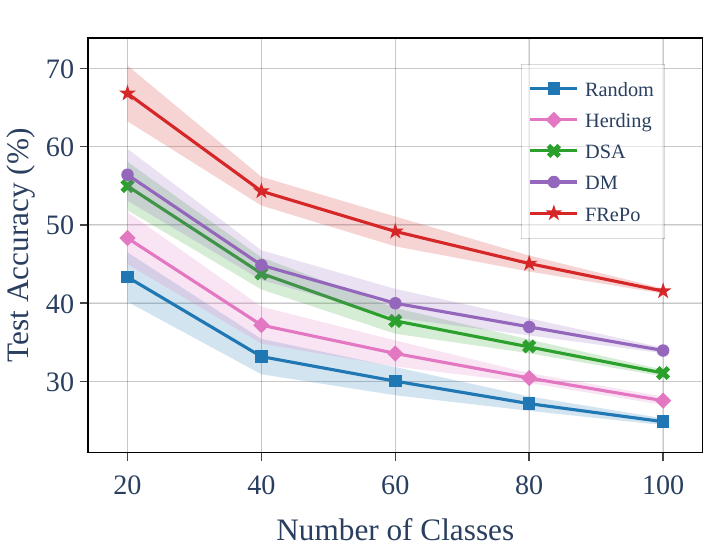}  
  \caption{5-step CL} \label{fig:cl5}
\end{subfigure}
\begin{subfigure}[b]{0.24\textwidth}
  \includegraphics[width=1.0\linewidth]{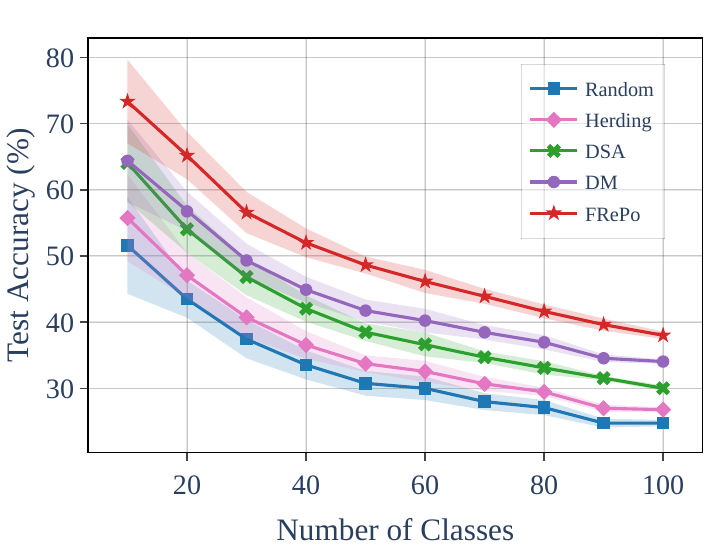}
  \caption{10-step CL} \label{fig:cl10}
\end{subfigure}
\begin{subfigure}[b]{0.24\textwidth}
  \includegraphics[width=1.0\linewidth]{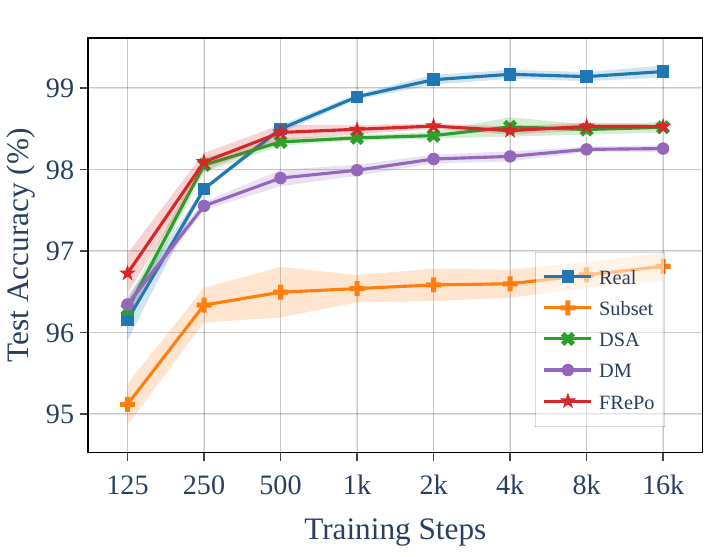}  
  \caption{MIA: ACC} \label{fig:mia-acc}
\end{subfigure}
\begin{subfigure}[b]{0.24\textwidth}
  \includegraphics[width=1.0\linewidth]{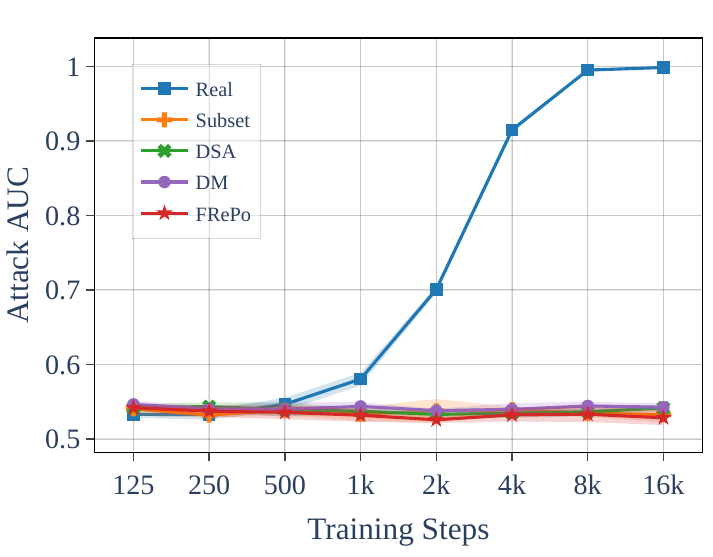}
  \caption{MIA: AUC} \label{fig:mia-auc}
\end{subfigure}
\caption{(a,b) Multi-class accuracies across all classes observed up to a certain time point. We perform significantly better than other methods in both 5 and 10 step class-incremental continual learning. (c,d) Test accuracy and attack AUC as we increase the number of training steps. AUC keeps increasing when training a model on the real data for more steps. In contrast, AUC keeps low when training on distilled data.}\label{fig:cl}
\end{figure}

Figure \ref{fig:cl} shows that our method performs significantly better than all previous methods. The final test accuracy for all classes for our method (\algname) and the second-best method (DM) are 41.6\%, 33.9\% in 5-step learning, and 38.0\%, 34.0\% in 10-step learning. However, we notice that for \algname, distilling 2000 images in a continual learning setup achieves a similar test accuracy (41.6\%) as distilling only 1000 images from the whole dataset (41.3\% from Table \ref{tab:dd_sota}). In addition, performance drops as we perform more steps. It suggests that \algname\ considers all available classes to derive the most condensed dataset. Splitting the data into multiple groups and performing independent distillation may generate redundant information or fail to capture the distinguishable features.

\begin{table}
  \caption{AUC of five attackers on models trained on the real and distilled MNIST data. The model trained on the real data is vulnerable to MIAs, while the model trained on the distilled data is robust to MIAs. Training on distilled data allows privacy preservation while retaining model performance.}
  \label{tab:mia}
  \small
  \centering
    \begin{tabular}{ccccccc}
        \toprule
        & \multirow{2}{*}[-4pt]{Test Acc (\%)} & \multicolumn{5}{c}{Attack AUC} \\
        \cmidrule(l){3-7} 
        & & Threshold & LR & MLP & RF & KNN \\
        \midrule 
        Real & $\textbf{99.2} \pm \textbf{0.1}$ & $0.99 \pm 0.01$ & $0.99 \pm 0.00$ & $1.00 \pm 0.00$ & $1.00 \pm 0.00$ & $0.97 \pm 0.00$ \\
        Subset & $96.8 \pm 0.2$ & $0.52 \pm 0.00$ & $ \textbf{0.50} \pm \textbf{0.01}$ & $\textbf{0.53} \pm \textbf{0.01}$ & $0.55 \pm 0.00$ & $0.54 \pm 0.00$ \\
        DSA & $98.5 \pm 0.1$ & $\textbf{0.50} \pm \textbf{0.00}$ & $0.51 \pm 0.00$ & $0.54 \pm 0.00$ & $0.54 \pm 0.01$ & $0.54 \pm 0.01$ \\
        DM & $98.3 \pm 0.0$ & $\textbf{0.50} \pm \textbf{0.00}$ & $0.51 \pm 0.01$ & $0.54 \pm 0.01$ & $0.54 \pm 0.01$ & $0.53 \pm 0.01$ \\
        \algname & $98.5 \pm 0.1$ & $0.52 \pm 0.00$ & $0.51 \pm 0.00$ & $\textbf{0.53} \pm \textbf{0.01}$ & $\textbf{0.52} \pm \textbf{0.01}$ & $\textbf{0.51} \pm \textbf{0.01}$ \\
        \bottomrule
    \end{tabular}
\end{table}

\subsection{Membership Inference Defense} \label{sec:mia}
\vspace{-0.05in}
Membership inference attacks (MIA) aim to infer whether a given data point has been used to train the model or not \citep{DBLP:conf/sp/ShokriSSS17, DBLP:journals/corr/abs-1802-04889, DBLP:conf/ndss/Salem0HBF019}. Ideally, we want a model to learn from the data but not memorize it to preserve privacy. However, deep neural networks are well-known for their ability to memorize all the training examples, even on large and randomly labeled datasets \citep{DBLP:conf/iclr/ZhangBHRV17}. Several methods have been proposed to defend against such attacks by either modifying the training procedure \citep{DBLP:conf/ccs/NasrSH18} or changing the inference workflow \citep{DBLP:conf/ccs/JiaSBZG19}. This section shows that the distilled data contain little information regarding sample presence in the original dataset. Thus, instead of training on the original datasets, training on distilled data allows privacy preservation while retaining model performance. 

We consider three distilled data generated by DSA \citep{DBLP:conf/icml/DSA}, DM \citep{DBLP:journals/corr/DM} and \algname. We perform five popular "black box" MIA provided by Tensorflow Privacy \citep{tfprivacy} on models trained on the real data or the data distilled from it. The attack methods include a threshold attack and four model-based attacks using logistic regression (LR), multi-layer perceptron (MLP), random forest (RF) and K-nearest neighbor (KNN). The inputs to those attack methods are ground-truth labels, model predictions, and losses. To measure the privacy vulnerability of the trained model, we compute the area under the ROC curve (AUC) of an attack classifier. Following prior work, \cite{DBLP:conf/sp/ShokriSSS17,DBLP:conf/csfw/YeomGFJ18}, we keep a balanced set of training examples (member) and test examples (non-member) with 10K each to maximize the uncertainty of MIA. Thus, the random guessing strategy results in a 50\% MIA accuracy. We conduct experiments on MNIST and FashionMNIST with a distillation size of 500. For space reasons, we provide more implementation details and results in appendix.

As shown in Table \ref{tab:mia}, all models trained on the distilled data preserve privacy as their attack AUCs are closed to random guessing. However, we observe a small drop in test accuracy compared to the model trained on the full dataset, which is expected as we only distill 500 examples instead of 10,000 examples. Compared to the model trained on an equally sized subset of the original data, the model trained on distilled data results in much better test performance. Figure \ref{fig:mia-acc}, \ref{fig:mia-auc} demonstrate the trade-off between test accuracy and attack effectiveness as measured by ROC AUC. It shows that early stopping can be an effective technique to preserve privacy. However, we will still be under high MIA risk if we perform early stopping by monitoring the validation loss. In contrast, training a model on the distilled data does not have this problem as the attack AUCs keep at a very low level regardless of training steps.
\section{Conclusion}
We propose neural Feature Regression with Pooling (\algname) to overcome two challenges in dataset distillation: meta-gradient computation and various types of overfitting in dataset distillation. We obtain state-of-the-art performance on various datasets with a 100x reduction in training time and a 10x reduction in GPU memory requirement. The distilled data generated by \algname\ looks real and natural and generalizes well to a wide range of architectures. Furthermore, we demonstrate two applications that take advantage of the high-quality distilled data, namely, continual learning and membership inference defense.

{{\bf Broader Impact}}\ ``Synthetic data'', in the broader sense of artificial data created by generative models, can help researchers understand how an otherwise opaque learning machine ``sees'' the world. There have been concerns regarding the risk of fake data. This paper explores a new research direction in generating synthetic data only for downstream classification tasks. We believe this work can provide additional interpretability and potentially address the common concerns in machine learning regarding training data privacy.

\begin{ack}
We would like to thank Harris Chan, Andrew Jung, Michael Zhang, Philip Fradkin, Denny Wu, Chong Shao, Leo Lee, Alice Gao, Keiran Paster, and Lazar Atanackovic for their valuable feedback. Jimmy Ba was supported by NSERC Grant [2020-06904], CIFAR AI Chairs program, Google Research Scholar Program and Amazon Research Award. This project was supported by LG Electronics Canada. Resources used in preparing this research were provided, in part, by the Province of Ontario, the Government of Canada through CIFAR, and companies sponsoring the Vector Institute for Artificial Intelligence.
\end{ack}

\bibliographystyle{unsrtnat}
\bibliography{ref}

\begin{thebibliography}{70}
\providecommand{\natexlab}[1]{#1}
\providecommand{\url}[1]{\texttt{#1}}
\expandafter\ifx\csname urlstyle\endcsname\relax
  \providecommand{\doi}[1]{doi: #1}\else
  \providecommand{\doi}{doi: \begingroup \urlstyle{rm}\Url}\fi

\bibitem[Hinton et~al.(2015)Hinton, Vinyals, and
  Dean]{DBLP:journals/corr/HintonVD15}
Geoffrey~E. Hinton, Oriol Vinyals, and Jeffrey Dean.
\newblock Distilling the knowledge in a neural network.
\newblock \emph{CoRR}, abs/1503.02531, 2015.
\newblock URL \url{http://arxiv.org/abs/1503.02531}.

\bibitem[Fukuda et~al.(2017)Fukuda, Suzuki, Kurata, Thomas, Cui, and
  Ramabhadran]{DBLP:conf/interspeech/FukudaSKTCR17}
Takashi Fukuda, Masayuki Suzuki, Gakuto Kurata, Samuel Thomas, Jia Cui, and
  Bhuvana Ramabhadran.
\newblock Efficient knowledge distillation from an ensemble of teachers.
\newblock In Francisco Lacerda, editor, \emph{Interspeech 2017, 18th Annual
  Conference of the International Speech Communication Association, Stockholm,
  Sweden, August 20-24, 2017}, pages 3697--3701. {ISCA}, 2017.
\newblock URL
  \url{http://www.isca-speech.org/archive/Interspeech\_2017/abstracts/0614.html}.

\bibitem[Polino et~al.(2018)Polino, Pascanu, and
  Alistarh]{DBLP:conf/iclr/PolinoPA18}
Antonio Polino, Razvan Pascanu, and Dan Alistarh.
\newblock Model compression via distillation and quantization.
\newblock In \emph{6th International Conference on Learning Representations,
  {ICLR} 2018, Vancouver, BC, Canada, April 30 - May 3, 2018, Conference Track
  Proceedings}. OpenReview.net, 2018.
\newblock URL \url{https://openreview.net/forum?id=S1XolQbRW}.

\bibitem[Wang et~al.(2018)Wang, Zhu, Torralba, and
  Efros]{DBLP:journals/corr/WangDD18}
Tongzhou Wang, Jun{-}Yan Zhu, Antonio Torralba, and Alexei~A. Efros.
\newblock Dataset distillation.
\newblock \emph{CoRR}, abs/1811.10959, 2018.
\newblock URL \url{http://arxiv.org/abs/1811.10959}.

\bibitem[Zhao et~al.(2021)Zhao, Mopuri, and Bilen]{DBLP:conf/iclr/DC}
Bo~Zhao, Konda~Reddy Mopuri, and Hakan Bilen.
\newblock Dataset condensation with gradient matching.
\newblock In \emph{9th International Conference on Learning Representations,
  {ICLR} 2021, Virtual Event, Austria, May 3-7, 2021}. OpenReview.net, 2021.
\newblock URL \url{https://openreview.net/forum?id=mSAKhLYLSsl}.

\bibitem[Rosasco et~al.(2021)Rosasco, Carta, Cossu, Lomonaco, and
  Bacciu]{DBLP:journals/corr/abs-2103-15851}
Andrea Rosasco, Antonio Carta, Andrea Cossu, Vincenzo Lomonaco, and Davide
  Bacciu.
\newblock Distilled replay: Overcoming forgetting through synthetic samples.
\newblock \emph{CoRR}, abs/2103.15851, 2021.
\newblock URL \url{https://arxiv.org/abs/2103.15851}.

\bibitem[Zhao and Bilen(2021{\natexlab{a}})]{DBLP:conf/icml/DSA}
Bo~Zhao and Hakan Bilen.
\newblock Dataset condensation with differentiable siamese augmentation.
\newblock In Marina Meila and Tong Zhang, editors, \emph{Proceedings of the
  38th International Conference on Machine Learning, {ICML} 2021, 18-24 July
  2021, Virtual Event}, volume 139 of \emph{Proceedings of Machine Learning
  Research}, pages 12674--12685. {PMLR}, 2021{\natexlab{a}}.
\newblock URL \url{http://proceedings.mlr.press/v139/zhao21a.html}.

\bibitem[Zhao and Bilen(2021{\natexlab{b}})]{DBLP:journals/corr/DM}
Bo~Zhao and Hakan Bilen.
\newblock Dataset condensation with distribution matching.
\newblock \emph{CoRR}, abs/2110.04181, 2021{\natexlab{b}}.
\newblock URL \url{https://arxiv.org/abs/2110.04181}.

\bibitem[Li et~al.(2021)Li, Togo, Ogawa, and
  Haseyama]{DBLP:journals/corr/abs-2104-02857}
Guang Li, Ren Togo, Takahiro Ogawa, and Miki Haseyama.
\newblock Soft-label anonymous gastric x-ray image distillation.
\newblock \emph{CoRR}, abs/2104.02857, 2021.
\newblock URL \url{https://arxiv.org/abs/2104.02857}.

\bibitem[Goetz and Tewari(2020)]{DBLP:journals/corr/abs-2008-04489}
Jack Goetz and Ambuj Tewari.
\newblock Federated learning via synthetic data.
\newblock \emph{CoRR}, abs/2008.04489, 2020.
\newblock URL \url{https://arxiv.org/abs/2008.04489}.

\bibitem[Maclaurin et~al.(2015)Maclaurin, Duvenaud, and
  Adams]{DBLP:conf/icml/MaclaurinDA15}
Dougal Maclaurin, David Duvenaud, and Ryan~P. Adams.
\newblock Gradient-based hyperparameter optimization through reversible
  learning.
\newblock In Francis~R. Bach and David~M. Blei, editors, \emph{Proceedings of
  the 32nd International Conference on Machine Learning, {ICML} 2015, Lille,
  France, 6-11 July 2015}, volume~37 of \emph{{JMLR} Workshop and Conference
  Proceedings}, pages 2113--2122. JMLR.org, 2015.
\newblock URL \url{http://proceedings.mlr.press/v37/maclaurin15.html}.

\bibitem[Bohdal et~al.(2020)Bohdal, Yang, and
  Hospedales]{DBLP:journals/corr/BohdalFDD20}
Ondrej Bohdal, Yongxin Yang, and Timothy~M. Hospedales.
\newblock Flexible dataset distillation: Learn labels instead of images.
\newblock \emph{CoRR}, abs/2006.08572, 2020.
\newblock URL \url{https://arxiv.org/abs/2006.08572}.

\bibitem[Sucholutsky and Schonlau(2019)]{DBLP:journals/corr/SucholutskyIDD19}
Ilia Sucholutsky and Matthias Schonlau.
\newblock Improving dataset distillation.
\newblock \emph{CoRR}, abs/1910.02551, 2019.
\newblock URL \url{http://arxiv.org/abs/1910.02551}.

\bibitem[Vicol et~al.(2021)Vicol, Metz, and
  Sohl{-}Dickstein]{DBLP:conf/icml/VicolMS21}
Paul Vicol, Luke Metz, and Jascha Sohl{-}Dickstein.
\newblock Unbiased gradient estimation in unrolled computation graphs with
  persistent evolution strategies.
\newblock In Marina Meila and Tong Zhang, editors, \emph{Proceedings of the
  38th International Conference on Machine Learning, {ICML} 2021, 18-24 July
  2021, Virtual Event}, volume 139 of \emph{Proceedings of Machine Learning
  Research}, pages 10553--10563. {PMLR}, 2021.
\newblock URL \url{http://proceedings.mlr.press/v139/vicol21a.html}.

\bibitem[Pascanu et~al.(2013)Pascanu, Mikolov, and
  Bengio]{DBLP:conf/icml/PascanuMB13}
Razvan Pascanu, Tom{\'{a}}s Mikolov, and Yoshua Bengio.
\newblock On the difficulty of training recurrent neural networks.
\newblock In \emph{Proceedings of the 30th International Conference on Machine
  Learning, {ICML} 2013, Atlanta, GA, USA, 16-21 June 2013}, volume~28 of
  \emph{{JMLR} Workshop and Conference Proceedings}, pages 1310--1318.
  JMLR.org, 2013.
\newblock URL \url{http://proceedings.mlr.press/v28/pascanu13.html}.

\bibitem[Metz et~al.(2019)Metz, Maheswaranathan, Nixon, Freeman, and
  Sohl{-}Dickstein]{DBLP:conf/icml/MetzMNFS19}
Luke Metz, Niru Maheswaranathan, Jeremy Nixon, C.~Daniel Freeman, and Jascha
  Sohl{-}Dickstein.
\newblock Understanding and correcting pathologies in the training of learned
  optimizers.
\newblock In Kamalika Chaudhuri and Ruslan Salakhutdinov, editors,
  \emph{Proceedings of the 36th International Conference on Machine Learning,
  {ICML} 2019, 9-15 June 2019, Long Beach, California, {USA}}, volume~97 of
  \emph{Proceedings of Machine Learning Research}, pages 4556--4565. {PMLR},
  2019.
\newblock URL \url{http://proceedings.mlr.press/v97/metz19a.html}.

\bibitem[Wu et~al.(2018)Wu, Ren, Liao, and Grosse]{DBLP:conf/iclr/WuRLG18}
Yuhuai Wu, Mengye Ren, Renjie Liao, and Roger~B. Grosse.
\newblock Understanding short-horizon bias in stochastic meta-optimization.
\newblock In \emph{6th International Conference on Learning Representations,
  {ICLR} 2018, Vancouver, BC, Canada, April 30 - May 3, 2018, Conference Track
  Proceedings}. OpenReview.net, 2018.
\newblock URL \url{https://openreview.net/forum?id=H1MczcgR-}.

\bibitem[Lee et~al.(2022)Lee, Chun, Jung, Yun, and
  Yoon]{DBLP:journals/corr/DCC}
Saehyung Lee, Sanghyuk Chun, Sangwon Jung, Sangdoo Yun, and Sungroh Yoon.
\newblock Dataset condensation with contrastive signals.
\newblock \emph{CoRR}, abs/2202.02916, 2022.
\newblock URL \url{https://arxiv.org/abs/2202.02916}.

\bibitem[Wang et~al.(2022)Wang, Zhao, Peng, Zhu, Yang, Wang, Huang, Bilen,
  Wang, and You]{DBLP:journals/corr/CAFE}
Kai Wang, Bo~Zhao, Xiangyu Peng, Zheng Zhu, Shuo Yang, Shuo Wang, Guan Huang,
  Hakan Bilen, Xinchao Wang, and Yang You.
\newblock {CAFE:} learning to condense dataset by aligning features.
\newblock \emph{CoRR}, abs/2203.01531, 2022.
\newblock \doi{10.48550/arXiv.2203.01531}.
\newblock URL \url{https://doi.org/10.48550/arXiv.2203.01531}.

\bibitem[Cazenavette et~al.(2022)Cazenavette, Wang, Torralba, Efros, and
  Zhu]{DBLP:journals/corr/MTT}
George Cazenavette, Tongzhou Wang, Antonio Torralba, Alexei~A. Efros, and
  Jun{-}Yan Zhu.
\newblock Dataset distillation by matching training trajectories.
\newblock \emph{CoRR}, abs/2203.11932, 2022.
\newblock \doi{10.48550/arXiv.2203.11932}.
\newblock URL \url{https://doi.org/10.48550/arXiv.2203.11932}.

\bibitem[Lee et~al.(2019)Lee, Xiao, Schoenholz, Bahri, Novak, Sohl{-}Dickstein,
  and Pennington]{DBLP:conf/nips/LeeXSBNSP19}
Jaehoon Lee, Lechao Xiao, Samuel~S. Schoenholz, Yasaman Bahri, Roman Novak,
  Jascha Sohl{-}Dickstein, and Jeffrey Pennington.
\newblock Wide neural networks of any depth evolve as linear models under
  gradient descent.
\newblock In Hanna~M. Wallach, Hugo Larochelle, Alina Beygelzimer, Florence
  d'Alch{\'{e}}{-}Buc, Emily~B. Fox, and Roman Garnett, editors, \emph{Advances
  in Neural Information Processing Systems 32: Annual Conference on Neural
  Information Processing Systems 2019, NeurIPS 2019, December 8-14, 2019,
  Vancouver, BC, Canada}, pages 8570--8581, 2019.
\newblock URL
  \url{https://proceedings.neurips.cc/paper/2019/hash/0d1a9651497a38d8b1c3871c84528bd4-Abstract.html}.

\bibitem[Nguyen et~al.(2021{\natexlab{a}})Nguyen, Chen, and
  Lee]{DBLP:conf/iclr/KIP1}
Timothy Nguyen, Zhourong Chen, and Jaehoon Lee.
\newblock Dataset meta-learning from kernel ridge-regression.
\newblock In \emph{9th International Conference on Learning Representations,
  {ICLR} 2021, Virtual Event, Austria, May 3-7, 2021}. OpenReview.net,
  2021{\natexlab{a}}.
\newblock URL \url{https://openreview.net/forum?id=l-PrrQrK0QR}.

\bibitem[Nguyen et~al.(2021{\natexlab{b}})Nguyen, Novak, Xiao, and
  Lee]{DBLP:conf/nips/KIP2}
Timothy Nguyen, Roman Novak, Lechao Xiao, and Jaehoon Lee.
\newblock Dataset distillation with infinitely wide convolutional networks.
\newblock In Marc'Aurelio Ranzato, Alina Beygelzimer, Yann~N. Dauphin, Percy
  Liang, and Jennifer~Wortman Vaughan, editors, \emph{Advances in Neural
  Information Processing Systems 34: Annual Conference on Neural Information
  Processing Systems 2021, NeurIPS 2021, December 6-14, 2021, virtual}, pages
  5186--5198, 2021{\natexlab{b}}.
\newblock URL
  \url{https://proceedings.neurips.cc/paper/2021/hash/299a23a2291e2126b91d54f3601ec162-Abstract.html}.

\bibitem[Lorraine et~al.(2020)Lorraine, Vicol, and
  Duvenaud]{DBLP:conf/aistats/LorraineVD20}
Jonathan Lorraine, Paul Vicol, and David Duvenaud.
\newblock Optimizing millions of hyperparameters by implicit differentiation.
\newblock In Silvia Chiappa and Roberto Calandra, editors, \emph{The 23rd
  International Conference on Artificial Intelligence and Statistics, {AISTATS}
  2020, 26-28 August 2020, Online [Palermo, Sicily, Italy]}, volume 108 of
  \emph{Proceedings of Machine Learning Research}, pages 1540--1552. {PMLR},
  2020.
\newblock URL \url{http://proceedings.mlr.press/v108/lorraine20a.html}.

\bibitem[Neal(1995)]{Neal1995BayesianLF}
Radford~M. Neal.
\newblock Bayesian learning for neural networks, 1995.

\bibitem[Russakovsky et~al.(2015)Russakovsky, Deng, Su, Krause, Satheesh, Ma,
  Huang, Karpathy, Khosla, Bernstein, Berg, and Fei-Fei]{ILSVRC15}
Olga Russakovsky, Jia Deng, Hao Su, Jonathan Krause, Sanjeev Satheesh, Sean Ma,
  Zhiheng Huang, Andrej Karpathy, Aditya Khosla, Michael Bernstein,
  Alexander~C. Berg, and Li~Fei-Fei.
\newblock {ImageNet Large Scale Visual Recognition Challenge}.
\newblock \emph{International Journal of Computer Vision (IJCV)}, 115\penalty0
  (3):\penalty0 211--252, 2015.
\newblock \doi{10.1007/s11263-015-0816-y}.

\bibitem[Rajeswaran et~al.(2019)Rajeswaran, Finn, Kakade, and
  Levine]{DBLP:conf/nips/RajeswaranFKL19}
Aravind Rajeswaran, Chelsea Finn, Sham~M. Kakade, and Sergey Levine.
\newblock Meta-learning with implicit gradients.
\newblock In Hanna~M. Wallach, Hugo Larochelle, Alina Beygelzimer, Florence
  d'Alch{\'{e}}{-}Buc, Emily~B. Fox, and Roman Garnett, editors, \emph{Advances
  in Neural Information Processing Systems 32: Annual Conference on Neural
  Information Processing Systems 2019, NeurIPS 2019, December 8-14, 2019,
  Vancouver, BC, Canada}, pages 113--124, 2019.
\newblock URL
  \url{https://proceedings.neurips.cc/paper/2019/hash/072b030ba126b2f4b2374f342be9ed44-Abstract.html}.

\bibitem[Werbos(1990)]{Werbos1990BPTT}
P.J. Werbos.
\newblock Backpropagation through time: what it does and how to do it.
\newblock \emph{Proceedings of the IEEE}, 78\penalty0 (10):\penalty0
  1550--1560, 1990.
\newblock \doi{10.1109/5.58337}.

\bibitem[Sutskever(2013)]{sutskever2013training}
Ilya Sutskever.
\newblock \emph{Training recurrent neural networks}.
\newblock University of Toronto Toronto, ON, Canada, 2013.

\bibitem[Tallec and Ollivier(2017)]{DBLP:journals/corr/TallecO17a}
Corentin Tallec and Yann Ollivier.
\newblock Unbiasing truncated backpropagation through time.
\newblock \emph{CoRR}, abs/1705.08209, 2017.
\newblock URL \url{http://arxiv.org/abs/1705.08209}.

\bibitem[Ba et~al.(2022)Ba, Erdogdu, Suzuki, Wang, Wu, and Yang]{ba2022high}
Jimmy Ba, Murat~A Erdogdu, Taiji Suzuki, Zhichao Wang, Denny Wu, and Greg Yang.
\newblock High-dimensional asymptotics of feature learning: How one gradient
  step improves the representation.
\newblock \emph{arXiv preprint arXiv:2205.01445}, 2022.

\bibitem[LeCun et~al.(1998)LeCun, Bottou, Bengio, and
  Haffner]{DBLP:journals/pieee/LeCunBBH98}
Yann LeCun, L{\'{e}}on Bottou, Yoshua Bengio, and Patrick Haffner.
\newblock Gradient-based learning applied to document recognition.
\newblock \emph{Proc. {IEEE}}, 86\penalty0 (11):\penalty0 2278--2324, 1998.
\newblock \doi{10.1109/5.726791}.
\newblock URL \url{https://doi.org/10.1109/5.726791}.

\bibitem[Xiao et~al.(2017)Xiao, Rasul, and
  Vollgraf]{DBLP:journals/corr/abs-1708-07747}
Han Xiao, Kashif Rasul, and Roland Vollgraf.
\newblock Fashion-mnist: a novel image dataset for benchmarking machine
  learning algorithms.
\newblock \emph{CoRR}, abs/1708.07747, 2017.
\newblock URL \url{http://arxiv.org/abs/1708.07747}.

\bibitem[Krizhevsky(2009)]{Krizhevsky09learningmultiple}
Alex Krizhevsky.
\newblock Learning multiple layers of features from tiny images.
\newblock Technical report, University of Toronto, 2009.

\bibitem[Le and Yang(2015)]{Le2015TinyIV}
Ya~Le and Xuan~S. Yang.
\newblock Tiny imagenet visual recognition challenge.
\newblock Technical report, Standford University, 2015.

\bibitem[Howard(2020)]{imagenette}
Jeremy Howard.
\newblock A smaller subset of 10 easily classified classes from imagenet, and a
  little more french, 2020.
\newblock URL \url{https://github.com/fastai/imagenette/}.

\bibitem[Wah et~al.(2011)Wah, Branson, Welinder, Perona, and Belongie]{cub200}
C.~Wah, S.~Branson, P.~Welinder, P.~Perona, and S.~Belongie.
\newblock The caltech-ucsd birds-200-2011 dataset.
\newblock Technical Report CNS-TR-2011-001, California Institute of Technology,
  2011.

\bibitem[Krizhevsky et~al.(2012)Krizhevsky, Sutskever, and
  Hinton]{DBLP:conf/nips/AlexNet}
Alex Krizhevsky, Ilya Sutskever, and Geoffrey~E. Hinton.
\newblock Imagenet classification with deep convolutional neural networks.
\newblock In Peter~L. Bartlett, Fernando C.~N. Pereira, Christopher J.~C.
  Burges, L{\'{e}}on Bottou, and Kilian~Q. Weinberger, editors, \emph{Advances
  in Neural Information Processing Systems 25: 26th Annual Conference on Neural
  Information Processing Systems 2012. Proceedings of a meeting held December
  3-6, 2012, Lake Tahoe, Nevada, United States}, pages 1106--1114, 2012.
\newblock URL
  \url{https://proceedings.neurips.cc/paper/2012/hash/c399862d3b9d6b76c8436e924a68c45b-Abstract.html}.

\bibitem[Simonyan and Zisserman(2015)]{DBLP:journals/corr/VGG}
Karen Simonyan and Andrew Zisserman.
\newblock Very deep convolutional networks for large-scale image recognition.
\newblock In Yoshua Bengio and Yann LeCun, editors, \emph{3rd International
  Conference on Learning Representations, {ICLR} 2015, San Diego, CA, USA, May
  7-9, 2015, Conference Track Proceedings}, 2015.
\newblock URL \url{http://arxiv.org/abs/1409.1556}.

\bibitem[He et~al.(2016)He, Zhang, Ren, and Sun]{DBLP:conf/cvpr/ResNet}
Kaiming He, Xiangyu Zhang, Shaoqing Ren, and Jian Sun.
\newblock Deep residual learning for image recognition.
\newblock In \emph{2016 {IEEE} Conference on Computer Vision and Pattern
  Recognition, {CVPR} 2016, Las Vegas, NV, USA, June 27-30, 2016}, pages
  770--778. {IEEE} Computer Society, 2016.
\newblock \doi{10.1109/CVPR.2016.90}.
\newblock URL \url{https://doi.org/10.1109/CVPR.2016.90}.

\bibitem[Ulyanov et~al.(2016)Ulyanov, Vedaldi, and
  Lempitsky]{DBLP:journals/corr/UlyanovVL16}
Dmitry Ulyanov, Andrea Vedaldi, and Victor~S. Lempitsky.
\newblock Instance normalization: The missing ingredient for fast stylization.
\newblock \emph{CoRR}, abs/1607.08022, 2016.
\newblock URL \url{http://arxiv.org/abs/1607.08022}.

\bibitem[Ioffe and Szegedy(2015)]{DBLP:conf/icml/IoffeS15}
Sergey Ioffe and Christian Szegedy.
\newblock Batch normalization: Accelerating deep network training by reducing
  internal covariate shift.
\newblock In Francis~R. Bach and David~M. Blei, editors, \emph{Proceedings of
  the 32nd International Conference on Machine Learning, {ICML} 2015, Lille,
  France, 6-11 July 2015}, volume~37 of \emph{{JMLR} Workshop and Conference
  Proceedings}, pages 448--456. JMLR.org, 2015.
\newblock URL \url{http://proceedings.mlr.press/v37/ioffe15.html}.

\bibitem[Kirkpatrick et~al.(2016)Kirkpatrick, Pascanu, Rabinowitz, Veness,
  Desjardins, Rusu, Milan, Quan, Ramalho, Grabska{-}Barwinska, Hassabis,
  Clopath, Kumaran, and Hadsell]{DBLP:journals/corr/KirkpatrickPRVD16}
James Kirkpatrick, Razvan Pascanu, Neil~C. Rabinowitz, Joel Veness, Guillaume
  Desjardins, Andrei~A. Rusu, Kieran Milan, John Quan, Tiago Ramalho, Agnieszka
  Grabska{-}Barwinska, Demis Hassabis, Claudia Clopath, Dharshan Kumaran, and
  Raia Hadsell.
\newblock Overcoming catastrophic forgetting in neural networks.
\newblock \emph{CoRR}, abs/1612.00796, 2016.
\newblock URL \url{http://arxiv.org/abs/1612.00796}.

\bibitem[French(1999)]{french99cf}
Robert French.
\newblock Catastrophic forgetting in connectionist networks.
\newblock \emph{Trends in cognitive sciences}, 3:\penalty0 128--135, 05 1999.
\newblock \doi{10.1016/S1364-6613(99)01294-2}.

\bibitem[Robins(1995)]{DBLP:journals/connection/Robins95}
Anthony~V. Robins.
\newblock Catastrophic forgetting, rehearsal and pseudorehearsal.
\newblock \emph{Connect. Sci.}, 7\penalty0 (2):\penalty0 123--146, 1995.
\newblock \doi{10.1080/09540099550039318}.
\newblock URL \url{https://doi.org/10.1080/09540099550039318}.

\bibitem[Buzzega et~al.(2020)Buzzega, Boschini, Porrello, and
  Calderara]{DBLP:conf/icpr/BuzzegaBPC20}
Pietro Buzzega, Matteo Boschini, Angelo Porrello, and Simone Calderara.
\newblock Rethinking experience replay: a bag of tricks for continual learning.
\newblock In \emph{25th International Conference on Pattern Recognition, {ICPR}
  2020, Virtual Event / Milan, Italy, January 10-15, 2021}, pages 2180--2187.
  {IEEE}, 2020.
\newblock \doi{10.1109/ICPR48806.2021.9412614}.
\newblock URL \url{https://doi.org/10.1109/ICPR48806.2021.9412614}.

\bibitem[Liu et~al.(2021)Liu, Schiele, and Sun]{DBLP:conf/cvpr/LiuSS21}
Yaoyao Liu, Bernt Schiele, and Qianru Sun.
\newblock Adaptive aggregation networks for class-incremental learning.
\newblock In \emph{{IEEE} Conference on Computer Vision and Pattern
  Recognition, {CVPR} 2021, virtual, June 19-25, 2021}, pages 2544--2553.
  Computer Vision Foundation / {IEEE}, 2021.
\newblock URL
  \url{https://openaccess.thecvf.com/content/CVPR2021/html/Liu\_Adaptive\_Aggregation\_Networks\_for\_Class-Incremental\_Learning\_CVPR\_2021\_paper.html}.

\bibitem[Rebuffi et~al.(2016)Rebuffi, Kolesnikov, and
  Lampert]{DBLP:journals/corr/RebuffiKL16}
Sylvestre{-}Alvise Rebuffi, Alexander Kolesnikov, and Christoph~H. Lampert.
\newblock icarl: Incremental classifier and representation learning.
\newblock \emph{CoRR}, abs/1611.07725, 2016.
\newblock URL \url{http://arxiv.org/abs/1611.07725}.

\bibitem[Prabhu et~al.(2020)Prabhu, Torr, and
  Dokania]{DBLP:conf/eccv/PrabhuTD20}
Ameya Prabhu, Philip H.~S. Torr, and Puneet~K. Dokania.
\newblock Gdumb: {A} simple approach that questions our progress in continual
  learning.
\newblock In Andrea Vedaldi, Horst Bischof, Thomas Brox, and Jan{-}Michael
  Frahm, editors, \emph{Computer Vision - {ECCV} 2020 - 16th European
  Conference, Glasgow, UK, August 23-28, 2020, Proceedings, Part {II}}, volume
  12347 of \emph{Lecture Notes in Computer Science}, pages 524--540. Springer,
  2020.
\newblock \doi{10.1007/978-3-030-58536-5\_31}.
\newblock URL \url{https://doi.org/10.1007/978-3-030-58536-5\_31}.

\bibitem[Aljundi et~al.(2019{\natexlab{a}})Aljundi, Lin, Goujaud, and
  Bengio]{DBLP:conf/nips/AljundiLGB19}
Rahaf Aljundi, Min Lin, Baptiste Goujaud, and Yoshua Bengio.
\newblock Gradient based sample selection for online continual learning.
\newblock In Hanna~M. Wallach, Hugo Larochelle, Alina Beygelzimer, Florence
  d'Alch{\'{e}}{-}Buc, Emily~B. Fox, and Roman Garnett, editors, \emph{Advances
  in Neural Information Processing Systems 32: Annual Conference on Neural
  Information Processing Systems 2019, NeurIPS 2019, December 8-14, 2019,
  Vancouver, BC, Canada}, pages 11816--11825, 2019{\natexlab{a}}.
\newblock URL
  \url{https://proceedings.neurips.cc/paper/2019/hash/e562cd9c0768d5464b64cf61da7fc6bb-Abstract.html}.

\bibitem[Aljundi et~al.(2019{\natexlab{b}})Aljundi, Belilovsky, Tuytelaars,
  Charlin, Caccia, Lin, and Page{-}Caccia]{DBLP:conf/nips/AljundiBTCCLP19}
Rahaf Aljundi, Eugene Belilovsky, Tinne Tuytelaars, Laurent Charlin, Massimo
  Caccia, Min Lin, and Lucas Page{-}Caccia.
\newblock Online continual learning with maximal interfered retrieval.
\newblock In Hanna~M. Wallach, Hugo Larochelle, Alina Beygelzimer, Florence
  d'Alch{\'{e}}{-}Buc, Emily~B. Fox, and Roman Garnett, editors, \emph{Advances
  in Neural Information Processing Systems 32: Annual Conference on Neural
  Information Processing Systems 2019, NeurIPS 2019, December 8-14, 2019,
  Vancouver, BC, Canada}, pages 11849--11860, 2019{\natexlab{b}}.
\newblock URL
  \url{https://proceedings.neurips.cc/paper/2019/hash/15825aee15eb335cc13f9b559f166ee8-Abstract.html}.

\bibitem[Liu et~al.(2020)Liu, Su, Liu, Schiele, and
  Sun]{DBLP:conf/cvpr/LiuSLSS20}
Yaoyao Liu, Yuting Su, An{-}An Liu, Bernt Schiele, and Qianru Sun.
\newblock Mnemonics training: Multi-class incremental learning without
  forgetting.
\newblock In \emph{2020 {IEEE/CVF} Conference on Computer Vision and Pattern
  Recognition, {CVPR} 2020, Seattle, WA, USA, June 13-19, 2020}, pages
  12242--12251. Computer Vision Foundation / {IEEE}, 2020.
\newblock \doi{10.1109/CVPR42600.2020.01226}.
\newblock URL
  \url{https://openaccess.thecvf.com/content\_CVPR\_2020/html/Liu\_Mnemonics\_Training\_Multi-Class\_Incremental\_Learning\_Without\_Forgetting\_CVPR\_2020\_paper.html}.

\bibitem[van~de Ven and Tolias(2019)]{DBLP:journals/corr/abs-1904-07734}
Gido~M. van~de Ven and Andreas~S. Tolias.
\newblock Three scenarios for continual learning.
\newblock \emph{CoRR}, abs/1904.07734, 2019.
\newblock URL \url{http://arxiv.org/abs/1904.07734}.

\bibitem[Castro et~al.(2018)Castro, Mar{\'{\i}}n{-}Jim{\'{e}}nez, Guil, Schmid,
  and Alahari]{DBLP:conf/eccv/CastroMGSA18}
Francisco~M. Castro, Manuel~J. Mar{\'{\i}}n{-}Jim{\'{e}}nez, Nicol{\'{a}}s
  Guil, Cordelia Schmid, and Karteek Alahari.
\newblock End-to-end incremental learning.
\newblock In Vittorio Ferrari, Martial Hebert, Cristian Sminchisescu, and Yair
  Weiss, editors, \emph{Computer Vision - {ECCV} 2018 - 15th European
  Conference, Munich, Germany, September 8-14, 2018, Proceedings, Part {XII}},
  volume 11216 of \emph{Lecture Notes in Computer Science}, pages 241--257.
  Springer, 2018.
\newblock \doi{10.1007/978-3-030-01258-8\_15}.
\newblock URL \url{https://doi.org/10.1007/978-3-030-01258-8\_15}.

\bibitem[Chen et~al.(2010)Chen, Welling, and Smola]{DBLP:conf/uai/ChenWS10}
Yutian Chen, Max Welling, and Alexander~J. Smola.
\newblock Super-samples from kernel herding.
\newblock In Peter Gr{\"{u}}nwald and Peter Spirtes, editors, \emph{{UAI} 2010,
  Proceedings of the Twenty-Sixth Conference on Uncertainty in Artificial
  Intelligence, Catalina Island, CA, USA, July 8-11, 2010}, pages 109--116.
  {AUAI} Press, 2010.
\newblock URL
  \url{https://dslpitt.org/uai/displayArticleDetails.jsp?mmnu=1\&smnu=2\&article\_id=2148\&proceeding\_id=26}.

\bibitem[Shokri et~al.(2017)Shokri, Stronati, Song, and
  Shmatikov]{DBLP:conf/sp/ShokriSSS17}
Reza Shokri, Marco Stronati, Congzheng Song, and Vitaly Shmatikov.
\newblock Membership inference attacks against machine learning models.
\newblock In \emph{2017 {IEEE} Symposium on Security and Privacy, {SP} 2017,
  San Jose, CA, USA, May 22-26, 2017}, pages 3--18. {IEEE} Computer Society,
  2017.
\newblock \doi{10.1109/SP.2017.41}.
\newblock URL \url{https://doi.org/10.1109/SP.2017.41}.

\bibitem[Long et~al.(2018)Long, Bindschaedler, Wang, Bu, Wang, Tang, Gunter,
  and Chen]{DBLP:journals/corr/abs-1802-04889}
Yunhui Long, Vincent Bindschaedler, Lei Wang, Diyue Bu, Xiaofeng Wang, Haixu
  Tang, Carl~A. Gunter, and Kai Chen.
\newblock Understanding membership inferences on well-generalized learning
  models.
\newblock \emph{CoRR}, abs/1802.04889, 2018.
\newblock URL \url{http://arxiv.org/abs/1802.04889}.

\bibitem[Salem et~al.(2019)Salem, Zhang, Humbert, Berrang, Fritz, and
  Backes]{DBLP:conf/ndss/Salem0HBF019}
Ahmed Salem, Yang Zhang, Mathias Humbert, Pascal Berrang, Mario Fritz, and
  Michael Backes.
\newblock Ml-leaks: Model and data independent membership inference attacks and
  defenses on machine learning models.
\newblock In \emph{26th Annual Network and Distributed System Security
  Symposium, {NDSS} 2019, San Diego, California, USA, February 24-27, 2019}.
  The Internet Society, 2019.

\bibitem[Zhang et~al.(2017)Zhang, Bengio, Hardt, Recht, and
  Vinyals]{DBLP:conf/iclr/ZhangBHRV17}
Chiyuan Zhang, Samy Bengio, Moritz Hardt, Benjamin Recht, and Oriol Vinyals.
\newblock Understanding deep learning requires rethinking generalization.
\newblock In \emph{5th International Conference on Learning Representations,
  {ICLR} 2017, Toulon, France, April 24-26, 2017, Conference Track
  Proceedings}. OpenReview.net, 2017.
\newblock URL \url{https://openreview.net/forum?id=Sy8gdB9xx}.

\bibitem[Nasr et~al.(2018)Nasr, Shokri, and Houmansadr]{DBLP:conf/ccs/NasrSH18}
Milad Nasr, Reza Shokri, and Amir Houmansadr.
\newblock Machine learning with membership privacy using adversarial
  regularization.
\newblock In David Lie, Mohammad Mannan, Michael Backes, and XiaoFeng Wang,
  editors, \emph{Proceedings of the 2018 {ACM} {SIGSAC} Conference on Computer
  and Communications Security, {CCS} 2018, Toronto, ON, Canada, October 15-19,
  2018}, pages 634--646. {ACM}, 2018.
\newblock \doi{10.1145/3243734.3243855}.
\newblock URL \url{https://doi.org/10.1145/3243734.3243855}.

\bibitem[Jia et~al.(2019)Jia, Salem, Backes, Zhang, and
  Gong]{DBLP:conf/ccs/JiaSBZG19}
Jinyuan Jia, Ahmed Salem, Michael Backes, Yang Zhang, and Neil~Zhenqiang Gong.
\newblock Memguard: Defending against black-box membership inference attacks
  via adversarial examples.
\newblock In Lorenzo Cavallaro, Johannes Kinder, XiaoFeng Wang, and Jonathan
  Katz, editors, \emph{Proceedings of the 2019 {ACM} {SIGSAC} Conference on
  Computer and Communications Security, {CCS} 2019, London, UK, November 11-15,
  2019}, pages 259--274. {ACM}, 2019.
\newblock \doi{10.1145/3319535.3363201}.
\newblock URL \url{https://doi.org/10.1145/3319535.3363201}.

\bibitem[tfp(2022)]{tfprivacy}
tensorflow/privacy: library for training machine learning models with privacy
  for training data, 2022.
\newblock URL \url{https://github.com/tensorflow/privacy}.

\bibitem[Yeom et~al.(2018)Yeom, Giacomelli, Fredrikson, and
  Jha]{DBLP:conf/csfw/YeomGFJ18}
Samuel Yeom, Irene Giacomelli, Matt Fredrikson, and Somesh Jha.
\newblock Privacy risk in machine learning: Analyzing the connection to
  overfitting.
\newblock In \emph{31st {IEEE} Computer Security Foundations Symposium, {CSF}
  2018, Oxford, United Kingdom, July 9-12, 2018}, pages 268--282. {IEEE}
  Computer Society, 2018.
\newblock \doi{10.1109/CSF.2018.00027}.
\newblock URL \url{https://doi.org/10.1109/CSF.2018.00027}.

\bibitem[Heek et~al.(2020)Heek, Levskaya, Oliver, Ritter, Rondepierre, Steiner,
  and van {Z}ee]{flax2020github}
Jonathan Heek, Anselm Levskaya, Avital Oliver, Marvin Ritter, Bertrand
  Rondepierre, Andreas Steiner, and Marc van {Z}ee.
\newblock {F}lax: A neural network library and ecosystem for {JAX}, 2020.
\newblock URL \url{http://github.com/google/flax}.

\bibitem[You et~al.(2020)You, Li, Reddi, Hseu, Kumar, Bhojanapalli, Song,
  Demmel, Keutzer, and Hsieh]{DBLP:conf/iclr/YouLRHKBSDKH20}
Yang You, Jing Li, Sashank~J. Reddi, Jonathan Hseu, Sanjiv Kumar, Srinadh
  Bhojanapalli, Xiaodan Song, James Demmel, Kurt Keutzer, and Cho{-}Jui Hsieh.
\newblock Large batch optimization for deep learning: Training {BERT} in 76
  minutes.
\newblock In \emph{8th International Conference on Learning Representations,
  {ICLR} 2020, Addis Ababa, Ethiopia, April 26-30, 2020}. OpenReview.net, 2020.
\newblock URL \url{https://openreview.net/forum?id=Syx4wnEtvH}.

\bibitem[Kingma and Ba(2015)]{DBLP:journals/corr/KingmaB14}
Diederik~P. Kingma and Jimmy Ba.
\newblock Adam: {A} method for stochastic optimization.
\newblock In Yoshua Bengio and Yann LeCun, editors, \emph{3rd International
  Conference on Learning Representations, {ICLR} 2015, San Diego, CA, USA, May
  7-9, 2015, Conference Track Proceedings}, 2015.
\newblock URL \url{http://arxiv.org/abs/1412.6980}.

\bibitem[Bradbury et~al.(2018)Bradbury, Frostig, Hawkins, Johnson, Leary,
  Maclaurin, Necula, Paszke, Vander{P}las, Wanderman-{M}ilne, and
  Zhang]{jax2018github}
James Bradbury, Roy Frostig, Peter Hawkins, Matthew~James Johnson, Chris Leary,
  Dougal Maclaurin, George Necula, Adam Paszke, Jake Vander{P}las, Skye
  Wanderman-{M}ilne, and Qiao Zhang.
\newblock {JAX}: composable transformations of {P}ython+{N}um{P}y programs,
  2018.
\newblock URL \url{http://github.com/google/jax}.

\bibitem[Ba et~al.(2016)Ba, Kiros, and Hinton]{DBLP:journals/corr/BaKH16}
Lei~Jimmy Ba, Jamie~Ryan Kiros, and Geoffrey~E. Hinton.
\newblock Layer normalization.
\newblock \emph{CoRR}, abs/1607.06450, 2016.
\newblock URL \url{http://arxiv.org/abs/1607.06450}.

\bibitem[Wu and He(2018)]{DBLP:conf/eccv/WuH18}
Yuxin Wu and Kaiming He.
\newblock Group normalization.
\newblock In Vittorio Ferrari, Martial Hebert, Cristian Sminchisescu, and Yair
  Weiss, editors, \emph{Computer Vision - {ECCV} 2018 - 15th European
  Conference, Munich, Germany, September 8-14, 2018, Proceedings, Part {XIII}},
  volume 11217 of \emph{Lecture Notes in Computer Science}, pages 3--19.
  Springer, 2018.
\newblock \doi{10.1007/978-3-030-01261-8\_1}.
\newblock URL \url{https://doi.org/10.1007/978-3-030-01261-8\_1}.

\bibitem[Yang et~al.(2022)Yang, Hu, Babuschkin, Sidor, Liu, Farhi, Ryder,
  Pachocki, Chen, and Gao]{DBLP:journals/corr/abs-2203-03466}
Greg Yang, Edward~J. Hu, Igor Babuschkin, Szymon Sidor, Xiaodong Liu, David
  Farhi, Nick Ryder, Jakub Pachocki, Weizhu Chen, and Jianfeng Gao.
\newblock Tensor programs {V:} tuning large neural networks via zero-shot
  hyperparameter transfer.
\newblock \emph{CoRR}, abs/2203.03466, 2022.
\newblock \doi{10.48550/arXiv.2203.03466}.
\newblock URL \url{https://doi.org/10.48550/arXiv.2203.03466}.

\end{thebibliography}

\clearpage
\newpage
\appendix
\section{Experimental Details}

\subsection{Implementation Details}\label{app:imple}
\textbf{Datasets:} We evaluate our methods on the following datasets:
i) \textbf{MNIST} \citep{DBLP:journals/pieee/LeCunBBH98}: A standard image dataset consists of 10 classes of 28x28 grey-scale images of handwritten digits, including 60,000 training examples and 10,000 test examples. 
ii) \textbf{FashionMNIST} \citep{DBLP:journals/corr/abs-1708-07747}: A direct drop-in replacement for MNIST \citep{DBLP:journals/pieee/LeCunBBH98} consisting of 10 classes of 28x28 grey-scale images of clothing, including 60,000 training examples and 10,000 test examples. We denote FashionMNIST as F-MNIST for simplicity.
iii) \textbf{CIFAR} \citep{Krizhevsky09learningmultiple}: A standard image dataset with two tasks: one coarse-grained over 10 classes (CIFAR10) and one fine-grained over 100 classes (CIFAR100). Both CIFAR10 and CIFAR100 have 50,000 training examples and 10,000 test examples with a resolution of 32x32. 
iv) \textbf{Tiny ImageNet} \citep{Le2015TinyIV}: A higher resolution (64x64) dataset with 200 classes, including 100,000 training examples and 10,000 test examples. We denote Tiny ImageNet as T-ImageNet for simplicity.
v) \textbf{ImageNette} and \textbf{ImageWoof} \citep{imagenette}: ImageNette (assorted objects) and ImageWoof (dog breeds) are two 10-class subsets from ILSVRC2012 \citep{ILSVRC15} designed to be easy and hard to learn, respectively. ImageNette consists of 9,469 training examples and 3,925 testing examples, while ImageWoof contains 9025 training examples and 3929 testing examples. We resize all examples to a resolution of 128x128. 
vi) \textbf{ImageNet} \citep{ILSVRC15}: A standard image benchmark dataset consists of 1000 classes from the Large Scale Visual Recognition Challenge 2012 \cite{ILSVRC15}, including 1,281,167 training examples and 50,000 testing examples. We resize all examples to a resolution of 64x64. 
vii) \textbf{CUB-200} \citep{cub200}: A fine-grained classification task, consisting of 200 subcategories belonging to birds. It has 5,994 data points for training and 5,794 data points for testing. We resize all examples to a resolution of 32x32.

\textbf{Data Preprocessing:} We use the standard preprocessing for all datasets but add regularized ZCA transformation for RGB datasets as described in KIP \cite{DBLP:conf/iclr/KIP1, DBLP:conf/nips/KIP2}. However, unlike KIP, we do not apply layer normalization to all examples. Moreover, for simplicity, we apply the same regularization strength $\lambda=0.1$ to all datasets rather than tune each dataset. This regularization strength is tuned for CIFAR in KIP \cite{DBLP:conf/nips/KIP2}. Besides, for ImageNette and ImageWoof, performing the full-size ZCA transformation is extremely expensive due to the high resolution, so we use a checkboard ZCA instead. It is the reason why our distilled images in Figure \ref{fig:intro}(c) have checkboard artifacts.
As for the previous method, DSA \citep{DBLP:conf/icml/DSA}, DM \citep{DBLP:journals/corr/DM} only use standard preprocessing, while MTT \citep{DBLP:journals/corr/MTT} and KIP \citep{DBLP:conf/nips/KIP2} consider both standard preprocessing and ZCA preprocessing, but they implement ZCA preprocessing differently. To account for the difference in data preprocessing, we reproduce each method using our data preprocessing and report the best of the reported value in the original paper and our reproducing results. As shown in Table \ref{tab:dd_sota}, it turns out that our data preprocessing can improve DSA \citep{DBLP:conf/icml/DSA} and DM \citep{DBLP:journals/corr/DM} a lot on RGB datasets but achieves a comparable performance for MTT \citep{DBLP:journals/corr/MTT}. When visualizing the distilled images, we directly apply the reverse transformation according to the corresponding data preprocessing without any other modifications.

\textbf{Models:} We use a simple convolutional neural network for all experiments based on the architecture used in previous works \citep{DBLP:conf/icml/DSA, DBLP:journals/corr/MTT, DBLP:conf/nips/KIP2}. It consists of several blocks of 3x3 convolution, normalization, RELU, and 2×2 average pooling layer with stride 2. We use 3, 4, and 5 blocks for datasets with resolutions 32x32, 64x64, and 128x128, respectively. Unlike previous works, which use the same number of filters at every layer, we double the number of filters if the feature map size is halved, following the modern neural network designs that preserve the time complexity per layer\citep{DBLP:journals/corr/VGG, DBLP:conf/cvpr/ResNet}. We observe this to be crucial for our method when we distill thousands of data because we need the feature dimension to be much larger than the distilled size to make the KRR work properly. Furthermore, we replace the Instance Normalization \citep{DBLP:journals/corr/UlyanovVL16} with Batch Normalization \citep{DBLP:conf/icml/IoffeS15} during training and do not use any normalization during evaluation. For DSA, DM, and MTT, the default normalization for both training and evaluation is the Instance Normalization \citep{DBLP:journals/corr/UlyanovVL16}. In contrast, KIP \citep{DBLP:conf/nips/KIP2} uses analytical NTK during training and uses a 1024-width network without normalization for evaluation. Moreover, KIP \citep{DBLP:conf/nips/KIP2} adds an extra convolution layer at the first layer of the neural network. We initialize our model using Lecun Initialization, which is the default in Flax library \citep{flax2020github}. In contrast, DSA, DM, and MTT use Kaiming Initialization, which is the default in PyTorch, while KIP \citep{DBLP:conf/nips/KIP2} initializes the model using random Gaussian with standard deviations $\sqrt{2}$ and 0.1 for weights and biases, respectively. We denote the default model used in each method as Conv for simplicity. We reproduce each method using our model and report the best of the reported value in the original paper and our reproducing results. We find our model achieves comparable or worse performance than each method's default methods, so we use their default model in most settings. Besides, we notice that the distilled data can encode the architecture's inductive bias, so we provide an ablation study on how the model architecture affects the distilled data in Section \ref{as:model}.

\textbf{Initialization:} Staying consistent with previous works, \citep{DBLP:conf/icml/DSA, DBLP:journals/corr/MTT, DBLP:conf/nips/KIP2}, we initialize the distilled image with randomly sampled real images. We also investigate initializing the distilled image with random Gaussian noise and observe that initializing does not affect the final performance too much but affects the convergence speed. We find the real initialization gives a decent convergence speed, so we choose the real initialization for simplicity rather than fine-tune the scale of random Gaussian initialization. As for labels, we initialize them using a scaled mean-centered one-hot vector for the corresponding image, where the scaling factor (i.e., $1/(\sqrt{\numclass/10})$) depends on the number of classes $\numclass$ . We find that this label initialization scheme speeds up the convergence but has little impact on the final performance. We provide an ablation study regarding the initialization scheme in Section \ref{as:initialization}.

\textbf{Label Learning:} Whether to learn the label is a Boolean hyperparameter in our experiments. When true, we optimize it using Algorithm \ref{alg:frepo}. When false, we stop the gradient so that they remain fixed at its initialization. We provide an ablation study on label learning in Section \ref{as:labellearning}. 

\textbf{Training and Evaluation:} The proposed algorithm has two versions depending on the order of meta-gradient computation and online model update. We present the two versions in Algorithm \ref{alg:frepo_app} and highlight the difference in red. We denote computing the meta-gradient after the online model update as v1 and denote computing the meta-gradient before the online model update as v2. The first implementation (v1) is more similar to the standard back-propagation through time as we first do the forward pass (online model update) and then the backward pass. However, instead of backpropagating the gradient through the inner optimization, we backpropagate through the kernel. In contrast, the insight behind the second implementation (v2) is that we want to find good data to train the last layer first, then we use it to train the whole network. Empirically, we do not observe any difference between these two implementations, and we choose v2 as the default for all experiments. Besides, we use the distilled data and its flipped version in meta-gradient computation to mitigate the mirroring effect for RGB datasets. For evaluation, we follow the standard protocol \citep{DBLP:conf/icml/DSA, DBLP:journals/corr/MTT}: training a randomly initialized neural network from scratch on distilled data and evaluating on a held-out test dataset. We apply the same data augmentation as in previous work \citep{DBLP:conf/icml/DSA, DBLP:journals/corr/MTT} during evaluation for a fair comparison. 

\begin{figure}[t]
\begin{algorithm}[H]
    \caption{Dataset Distillation using Neural Feature Regression with Pooling (\algname)}
    \label{alg:frepo_app}
    \begin{algorithmic}
        \STATE {\bfseries Require:} $\targetdataset$: a labeled dataset; $\ddlr$: the learning rate for the distilled data
        \STATE {\bfseries Initialization:} Initialize a labeled distilled dataset $\supportdataset=\supportdata$.
        \STATE {\bfseries Initialization:} Initialize a model pool $\modelpool$ with $\numpool$ models ${\{\nnparam_{i}\}}_{i=1}^{\numpool}$ randomly initialized from $\modeldist$. 
    \end{algorithmic}
    \begin{algorithmic}[1]
        \WHILE{not converged}
        \STATE $\smalltriangleright$ Sample a model uniformly from the model pool: $\theta_{i} \sim \modelpool$. 
        \STATE $\smalltriangleright$ \textcolor{red}{v1: Train the model $\theta_i$ on the current distilled data $\supportdataset$ for one step.}
        \STATE $\smalltriangleright$ Sample a target batch uniformly from the labeled dataset: $\targetdata \sim \targetdataset$.
        \STATE $\smalltriangleright$ Compute the meta-training loss $\targetloss$ using Eq. \ref{eqn:loss}
        \STATE $\smalltriangleright$ Update the distilled data $\supportdataset$: $\supportx \leftarrow \supportx - \ddlr \grad \supportx \targetloss$, and $\supporty \leftarrow \supporty - \ddlr \grad \supporty \targetloss$
        \STATE $\smalltriangleright$ \textcolor{red}{v2: Train the model $\theta_i$ on the current distilled data $\supportdataset$ for one step.}
        \STATE $\smalltriangleright$ Reinitialize the model $\theta_i \sim \modeldist$ if $\theta_i$ has been updated more than $\maxonlineupdate$ steps.
        \ENDWHILE
    \end{algorithmic}
    \begin{algorithmic}
        \STATE {\bfseries Output:} Learned distilled dataset $\supportdataset=\supportdata$
    \end{algorithmic}
\end{algorithm}
\vspace{-0.5in}
\end{figure}

\textbf{Hyperparameters:} We aim to keep our method as simple and efficient as possible. As a result, our method requires very few hyperparameter tuning efforts than all the previous methods. We use the same set of hyperparameters for all experiments, except stated otherwise. Specifically, we use LAMB optimizer \citep{DBLP:conf/iclr/YouLRHKBSDKH20} with a cosine learning rate schedule starting from 0.0003 for both images and labels. We use a batch size of 1024 and train the distilled data up to 2 million steps to see its long-run behavior. In practice, we observe that most of the convergence (more than 95\% of its final test accuracy) is achieved after a few thousand steps with a slow, logarithmic increase with more iterations, as shown in Figure \ref{fig:complexity_acc_time}. The model pool contains ten models trained up to $\maxonlineupdate=100$ steps. Each model is trained using Adam optimizer \citep{DBLP:journals/corr/KingmaB14} with a constant learning rate of 0.0003, and no weight decay is applied. We use the same kernel regularizer $\lambda$ as KIP \citep{DBLP:conf/nips/KIP2}. 
Instead of being a fixed constant, the regularizer is adapted to the scale of $\gram \supportx \supportx$, $\lambda = \lambda_0 Tr(\gram \supportx \supportx)$, where $\lambda_0 = 10^{-6}$. We note that our hyperparameter choice may be sub-optimal, but it is a good starting point. We conduct some ablation study regarding the hyperparameters in Section \ref{as:title}. Moreover, we provide a hyperparameter tuning guideline for practitioners in Section \ref{app:tuningguide} accompanied by a list of additional tricks that we find to improve the performance but do not include in the current algorithm. 

\textbf{Summary:} We implement our method in JAX \citep{jax2018github} and reproduce previous methods using their released code. To take into account the differences in data processing and architectures, we try our best to reproduce previous results by varying different data preprocessing and models. Experiments show that our data preprocessing can sometimes improve performance, but our model does not give better performance than previous methods. We report the best of the reported value in the original paper and our reproducing results. We evaluate each distilled data using five random neural networks and report the mean and standard deviation. 

\subsection{Experimental Setups}

\textbf{Figure \ref{fig:intro}:} Selected images from (a) 1 Img/Cls CIFAR100 (ZCA, learn label=True), (b) 1 Img/Cls Tiny ImageNet (ZCA, learn label=True), (c) 10 Img/Cls ImageNette (Top 2 rows) (Checkboard ZCA, learn label=True), 10 Img/Cls ImageWoof (Bottom 2 rows) (Checkboard ZCA, learn label=True). 

\textbf{Figure \ref{fig:complexity_acc_time_300}, \ref{fig:complexity_acc_time}:} We measure the time per step by measuring the average wall clock time ten times on ten steps. We take the first measurement after 50 steps when the statistics become stable and report the mean and standard deviation of 10 runs. To generate Figure \ref{fig:complexity_acc_time_300}, \ref{fig:complexity_acc_time}, we use the default model and default hyperparameter for each method. We first record the time per step of each method and run another program to evaluate the checkpoints at different time steps to get the test accuracy. Thus, the wall clock time is computed by multiplying the time per step and the number of steps taken at each checkpoint. All models are trained on Nvidia Quadro RTX 6000 with 22.17GB memory, and 7.5 compute capability.

\textbf{Figure \ref{fig:complexity_timeperstep}, \ref{fig:complexity_memory}, \ref{fig:complexity_timeperstep_size}, \ref{fig:complexity_memory_size}:} Similar to how we generate Figure \ref{fig:complexity_acc_time_300}, \ref{fig:complexity_acc_time}, we measure the time per step by measuring the average wall clock time ten times on ten steps. We take the first measurement after 50 steps when the statistics become stable and report the mean and standard deviation of 10 runs. We use jax.profiler and torch.profiler for the JAX program and PyTorch program to measure the peak GPU memory usage. Different from Figure \ref{fig:complexity_acc_time_300}, \ref{fig:complexity_acc_time}, we use the same model (i.e., Conv used by DSA \citep{DBLP:conf/icml/DSA}) and same batch size=256 in this section. We take an optimistic estimation for the previous methods (i.e., DSA, MTT) by choosing a smaller inner loop or outer loop number to generate more data points before encountering out of memory errors. Specifically, we use outer\_loop=1, inner\_loop=1 for DSA and use syn\_steps=15 for MTT. All models are trained on Nvidia Quadro RTX 6000 with 22.17GB memory, and 7.5 compute capability.

\textbf{Figure \ref{fig:vis_frepo}}: Distilled image visualization when distilling 1 Img/Cls from CIFAR100 using \algname.

\textbf{Figure \ref{fig:vis_mtt}}: Distilled image visualization when distilling 1 Img/Cls from CIFAR100 using MTT \citep{DBLP:journals/corr/MTT}. To give the best image quality, we use the distilled images provided by the original paper. (Url:\url{https://georgecazenavette.github.io/mtt-distillation/tensors/index.html#tensors})

\textbf{Figure \ref{fig:vis_dsa}}: Distilled image visualization when distilling 1 Img/Cls from CIFAR100 using DSA \citep{DBLP:conf/icml/DSA}. To give the best image quality, we use the distilled images provided by the original paper. (Url: \url{https://drive.google.com/drive/folders/1Dp6V6RvhJQPsB-2uZCwdlHXf1iJ9Wb_g}). Besides, we adjust the contrasts to give the best visualization.

\textbf{Figure \ref{fig:lb-analysis}}: Distilled label visualization when distilling 1 Img/Cls from CIFAR100 using \algname. The distilled images are sunflower, girl, and bear, respectively. All labels are unnormalized logits. 

\textbf{Figure \ref{fig:cl5}, \ref{fig:cl10}:} Multi-class accuracies across all classes observed up to a specific time point. For a fair comparison, we follow the same setup in DM \citep{DBLP:journals/corr/DM}, including the class split of five different runs. We use the same set of hyperparameters as other experiments and train the distilled data up to 500K steps. We report the mean and standard deviation of the five different runs. We observe that the primary source of variance comes from the class split.

\textbf{Figure \ref{fig:mia-acc}, \ref{fig:mia-auc}, \ref{fig:mia-acc-fmnist}, \ref{fig:mia-auc-fmnist}:} Test accuracies and attack AUCs as we increase the number of training steps. We use the same set of hyperparameters as other experiments except for data augmentation. We do not apply any data augmentation when we train models on the distilled data since any type of regularization can alleviate the MIA risks, making it hard to see the effects of distilled data. 

\textbf{Table \ref{tab:dd_sota}, \ref{tab:imagenette}, \ref{tab:dd_app_sota}, \ref{tab:imagenette_app}:} We try to reproduce the previous methods based on their official codebase and vary the data preprocessing and model architecture. We report the best of the reported value in the original paper and our reproducing results. As for our method, Table \ref{tab:dd_sota}, \ref{tab:imagenette} report the best value of learning label or not learning label with our default architecture and data preprocessing. We use the same set of hyperparameters for all experiments. The complete results can be found in Table \ref{tab:dd_app_sota}. Besides, we also include the results when training the model on the full dataset using mean square error loss in Table \ref{tab:imagenette_app}.

\textbf{Table \ref{tab:dd_ca}:} We generate the distilled data for each method using their default model and default hyperparameter. For KIP, since the training is too expensive, we use the checkpoint provided by the original author. We perform a sweep on the checkpoints in ``gs://kip-datasets/kip/cifar10/'' and find ``ConvNet\_ssize100\_zca\_nol\_noaug\_ckpt1000.npz'' gives the best performance that matches the reported value in the original paper. We evaluate DSA, DM, and MTT in PyTorch and KIP and \algname\ in JAX. We notice that our reproducing results for KIP are much better than the reproducing results reported in MTT \citep{DBLP:journals/corr/MTT}. It may be due to the differences in initialization or hyperparameter choices, such as learning rate. 

\textbf{Table \ref{tab:mia}, \ref{tab:mia_app}:} Table \ref{tab:mia} and Table \ref{tab:mia_app} present the test accuracies and MIA results on MNIST and FashionMNIST, respectively. We random sample 10K data points to generate the distilled data and sample another 10K non-overlapped data as non-member data. We use the same set of hyperparameters as other experiments except for data augmentation. We do not apply any data augmentation when we train models on the distilled data since any regularization can alleviate the MIA risks, making it hard to see the effects of distilled data. 
\section{Additional Results}

\textbf{Resized ImageNet-1K:} We also evaluate our method on a resized version of ILSVRC2012 \cite{ILSVRC15} with a resolution of 64x64 to see how it performs on a complex label space. As shown in Table \ref{tab:imagenette_app}, we can achieve 7.5\% and 9.7\% Top1 accuracy using only 1k and 2k training examples, compared to 1.1\% and 1.4\% using an equally-sized real subset or 19.8\% on the full dataset using 1281167 training examples. We notice that Mean Square Error (MSE) loss may not be suitable for a complex dataset like ImageNet as the same model trained with Cross-Entropy loss achieve a Top1 accuracy of 32.2\%. However, since we use MSE loss to evaluate the distilled data, we report the model trained using MSE loss for a fair comparison. 

\begin{table}
  \caption{Distillation performance on higher resolution (128x128) dataset (i.e. ImageNette, ImageWoof) and medium resolution (64x64) dataset with a complex label space (i.e. ImageNet-1K). \algname\ scales well to high-resolution images and learns the discriminate feature of complex datasets well.}
  \label{tab:imagenette_app}
  \small
  \centering
    \begin{tabular}{ccccccc}
        \toprule
        \multirow{2}{*}[-4pt]{} &
        \multicolumn{2}{c}{ImageNette (128x128)} & \multicolumn{2}{c}{ImageWoof (128x128)} & \multicolumn{2}{c}{ImageNet (64x64)}\\
        \cmidrule(l){2-3} \cmidrule(l){4-5} \cmidrule(l){6-7}
        Img/Cls & 1 & 10 & 1 & 10 & 1 & 2\\
        \midrule
        Random Subset & $23.5 \pm 4.8$ & $47.7 \pm 2.4$ & $14.2 \pm 0.9$ & $27.0 \pm 1.9$ & $1.1 \pm 0.1$ & $1.4 \pm 0.1$ \\
        MTT~\citep{DBLP:journals/corr/MTT} & $47.7 \pm 0.9$ & $63.0 \pm 1.3$ & $ 28.6 \pm 0.8 $ & $35.8 \pm 1.8$ & $ - $ & $ - $ \\
        \algname & $ \textbf{48.1} \pm \textbf{0.7}$ & $ \textbf{66.5} \pm \textbf{0.8}$ & $ \textbf{29.7} \pm \textbf{0.6} $ & $ \textbf{42.2} \pm \textbf{0.9}$ & $ \textbf{7.5} \pm \textbf{0.3}$ & $\textbf{9.7} \pm \textbf{0.2}$ \\
        \midrule
        Full Dataset & \multicolumn{2}{c}{$87.9 \pm 1.0$} & \multicolumn{2}{c}{$74.4 \pm 1.6$} & \multicolumn{2}{c}{$19.8 \pm 0.6$} \\
        \bottomrule
    \end{tabular}
\end{table}

\begin{figure}[t!]
\centering
\begin{subfigure}[b]{0.45\textwidth}
  \includegraphics[width=1.0\linewidth]{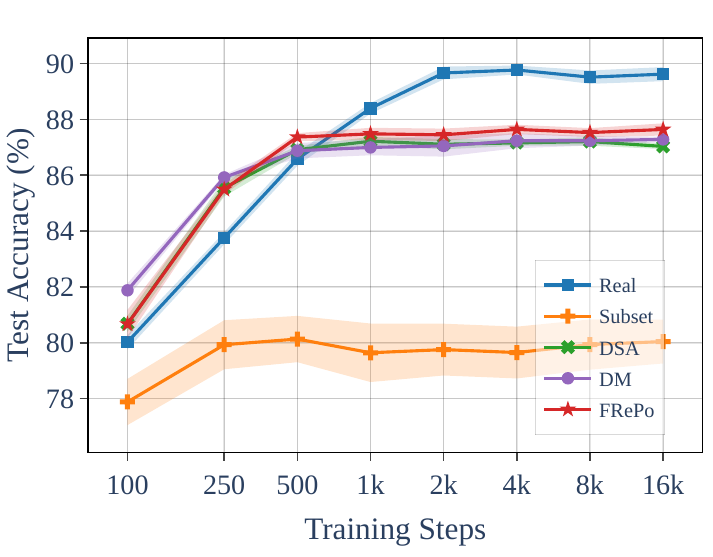}  
  \caption{MIA: ACC} \label{fig:mia-acc-fmnist}
\end{subfigure}
\begin{subfigure}[b]{0.45\textwidth}
  \includegraphics[width=1.0\linewidth]{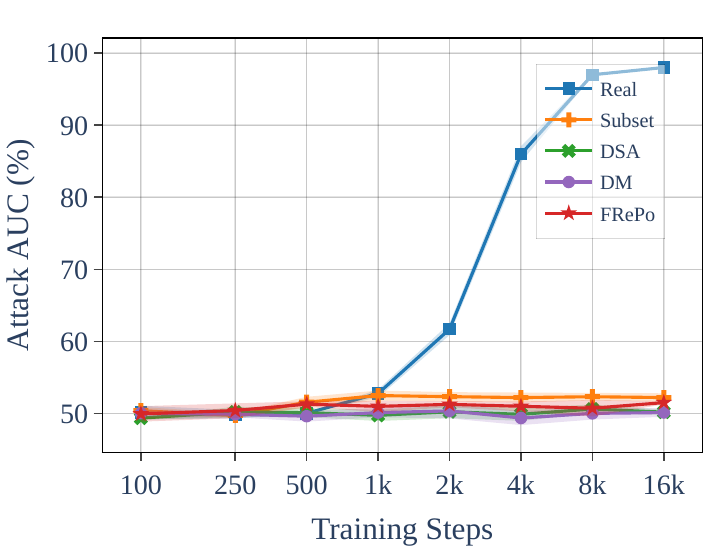}
  \caption{MIA: AUC} \label{fig:mia-auc-fmnist}
\end{subfigure}
\vspace{-0.05in}
\caption{(a,b) Test accuracy and attack AUC on Fashion MNIST as we increase the number of training steps. AUC keeps increasing when training a model on the real data for more steps. In contrast, AUC keeps low when training on distilled data.}\label{fig:mia-app}
\vspace{-0.1in}
\end{figure}

\textbf{Membership Inference Defense:} We only show MNIST results in Section \ref{sec:mia} due to space reasons. We provide the same experimental results on FashionMNIST in Figure \ref{fig:mia-app} and Table \ref{tab:mia_app}. We observe a similar trend on the two different datasets: training a model on the distilled data can preserve the privacy while achieving a good performance. We notice that there is a gap between training on the whole data set, which may be closed by distilling more data points and adding more noise when perform the distillation. 

\begin{table}
  \caption{AUC of attacker classifier trained on real data and distilled data on FashionMNSIT. We highlight the best attacker performance in bold for each model. The model trained on real data is vulnerable to membership inference attacks. In contrast, the model trained on distilled data is robust to membership inference attacks.}
  \label{tab:mia_app}
  \small
  \centering
    \begin{tabular}{ccccccc}
        \toprule
        & \multirow{2}{*}[-4pt]{Test Acc (\%)} & \multicolumn{5}{c}{Attack AUC} \\
        \cmidrule(l){3-7} 
        & & Threshold & LR & MLP & RF & KNN \\
        \midrule 
        Real & $89.7 \pm 0.2$ & $0.99 \pm 0.01$ & $0.99 \pm 0.00$ & $0.99 \pm 0.00$ & $0.99 \pm 0.00$ & $0.98 \pm 0.00$ \\
        Subset & $81.1 \pm 0.7$ & $ 0.53 \pm 0.01$ & $0.51 \pm 0.01$ & $0.52 \pm 0.01$ & $0.52 \pm 0.01 $ & $0.53 \pm 0.00$ \\
        DSA & $87.0 \pm 0.1$ & $0.51 \pm 0.00$ & $0.51 \pm 0.01$ & $ 0.51 \pm 0.01$ & $0.52 \pm 0.01$ & $ 0.51 \pm 0.01$ \\
        DM & $87.3 \pm 0.1$ & $0.52 \pm 0.00$ & $0.51 \pm 0.01$ & $ 0.50 \pm 0.01$ & $0.52 \pm 0.01$ & $0.51 \pm 0.01$ \\
        \algname & $87.6 \pm 0.2$ & $0.52 \pm 0.00$ & $0.53 \pm 0.01$ & $0.53 \pm 0.01$ & $0.53 \pm 0.01$ & $0.52 \pm 0.00$ \\
        \bottomrule
    \end{tabular}
\end{table}

\section{Ablation Study} \label{as:title}
We conduct several ablation studies to understand the key components of the proposed method, including the role of kernel approximation (Section \ref{as:tbptt}), model pool (Section \ref{as:pool}), initialization (Section \ref{as:initialization}), label learning (Section \ref{as:labellearning}), scalability (Section \ref{as:trainingcost}), and model architecture \ref{as:model}. All experiments are conducted on CIFAR100 when learning 1 Img/Cls using Algorithm \ref{alg:frepo} with the default hyperparameters except stated otherwise.

\subsection{\algname\ vs TBPTT} \label{as:tbptt}
\begin{figure}[t]
  \centering
  \hspace*{-0.035\linewidth}
  \includegraphics[width=1.07\linewidth]{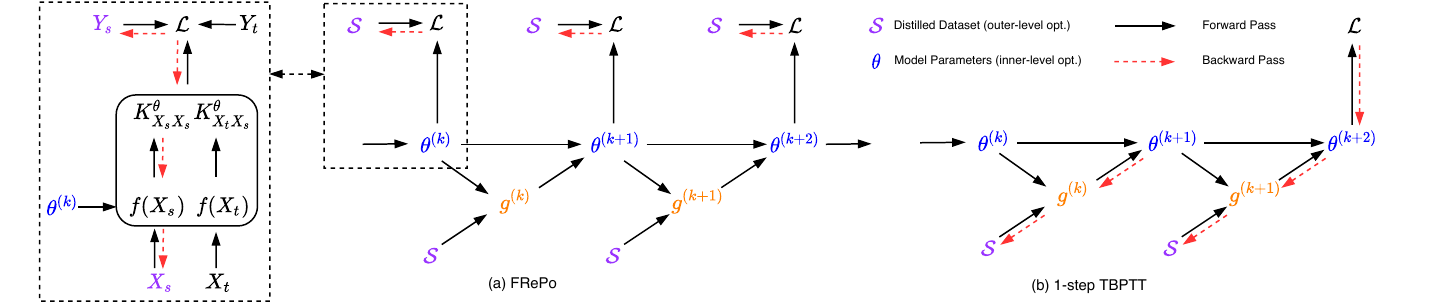}
  \caption{Comparison of \algname\ and 1-step TBPTT. $S$, $X_s$, $Y_s$ are the distilled dataset, images and labels. $\mathcal{L}$ is the meta-training loss and $\theta^{(k)}$, $g^{(k)}$ are the model parameter and gradient at step $k$. $f(X)$ is the feature for input $X$ and $\gram \targetx \supportx$ is the Gram matrix of $\targetx$ and $\supportx$. \algname\ is analogous to 1-step TBPTT as it computes the meta-gradient at each step while performing the online model update. However, instead of backpropagating through the inner optimization, \algname\ computes the meta-gradient through a kernel and feature extractor.}\label{fig:alg_app}
\end{figure}

\begin{figure}[t!]
\centering
\begin{subfigure}[b]{0.32\textwidth}
  \includegraphics[width=1.0\linewidth]{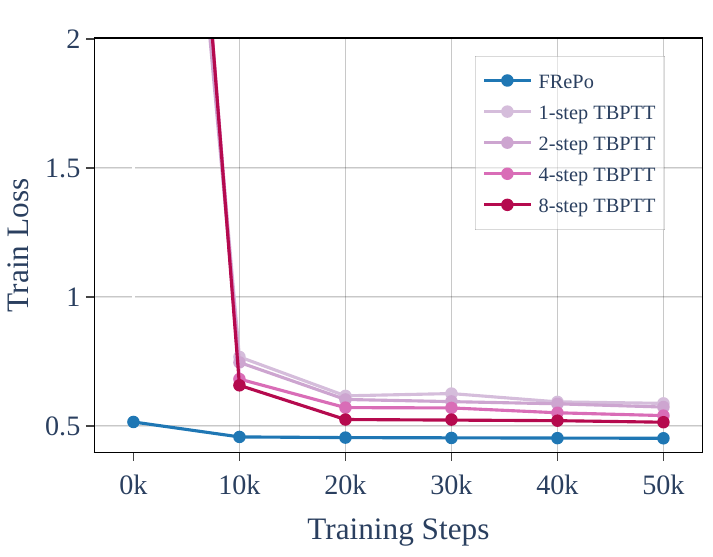}  
  \caption{Train Loss} \label{fig:as_bptt_train_loss}
\end{subfigure}
\begin{subfigure}[b]{0.32\textwidth}
  \includegraphics[width=1.0\linewidth]{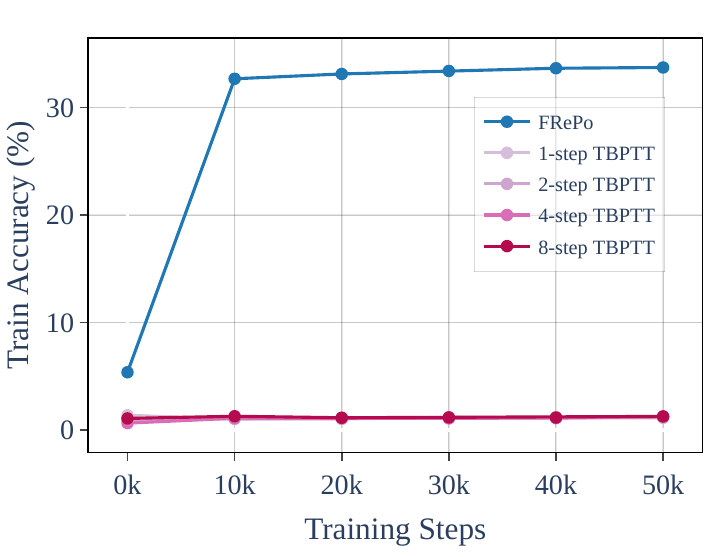}
  \caption{Train Acc} \label{fig:as_bptt_train_acc}
\end{subfigure}
\begin{subfigure}[b]{0.32\textwidth}
  \includegraphics[width=1.0\linewidth]{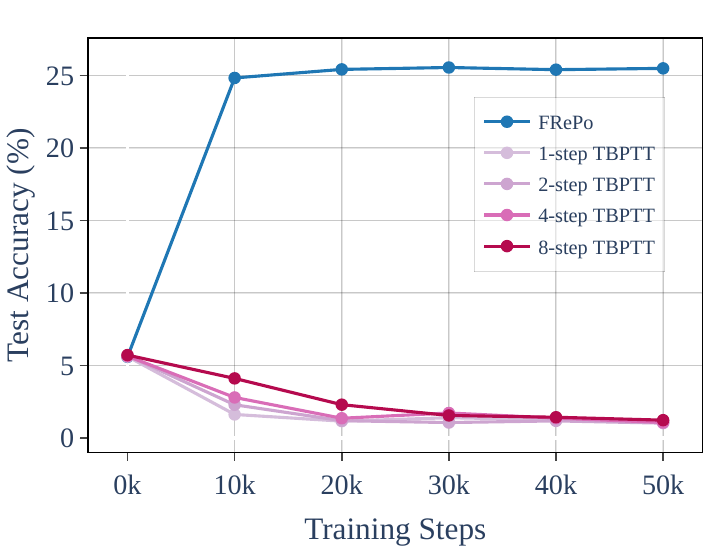} 
  \caption{Test Acc} \label{fig:as_bptt_test_acc}
\end{subfigure}
\caption{Ablation Study - Truncated Backpropagation through Time (TBPTT). Due to truncation bias, k-step TBPTT can easily overfit to the k-step training scheme. In contrast, \algname\ alleviates the truncation bias of TBPTT by training a subset of a neural network to convergence.}\label{fig:as_bptt}
\end{figure}

Figure \ref{fig:alg_app} illustrates the computation graph of \algname and 1-step TBPTT, which is different from Figure \ref{fig:alg} where we compare \algname\ with unrolled optimization. \algname\ differs from 1-step TBPTT in meta-gradient computation. As shown in Figure \ref{fig:alg_app}, 1-step TBPTT computes the meta-gradient by backpropagating through the inner optimization, while \algname\ uses backpropagating through a kernel and a feature extractor. Figure \ref{fig:as_bptt} compares the training loss, training accuracy, and test accuracy of \algname\ and TBPTT when using the same training and evaluation protocol. Note that \algname\ computes the training loss and accuracy using the KRR head during training, while TBPTT uses the neural network head. We use the same optimizer to perform the online model update and use the same optimizer to train the same neural network for the same amount of steps on the distilled data during evaluation. 

Figure \ref{fig:as_bptt_test_acc} shows that the distilled data generated by TBPTT keeps getting worse test accuracy as the training goes on. It suggests that the meta-gradient computed by TBPTT is highly biased and does not help learn a generalizable distilled dataset. As a result, the distilled data is overfitted to a k-step learning setup, where the k is the truncation step. This learning scenario is very similar to the dataset distillation setup in DD \citep{DBLP:journals/corr/WangDD18} and MTT \citep{DBLP:journals/corr/MTT}, where a specific optimizer is learned to take advantage of the distilled data. Though using more truncation steps can alleviate this problem, eliminating the truncation bias needs an infinite unrolled optimization, making it intractable. In contrast, \algname\ alleviates the truncation bias by training a subset of a neural network to convergence. Moreover, \algname\ decouples the meta-gradient computation and online model update such that the distilled data will not overfit the inner-level optimization. Conversely, TBPTT needs to fine-tune the inner optimization to get the best performance, which we do not explore further here. 

\subsection{Model pool and Batchsize} \label{as:pool}
\begin{figure}[t!]
\centering
\begin{subfigure}[b]{0.24\textwidth}
  \includegraphics[width=1.0\linewidth]{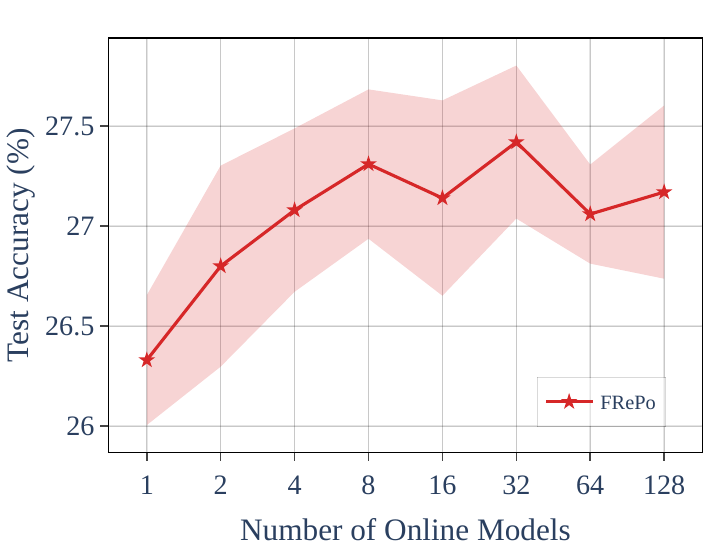}  
  \caption{Number of Models} \label{fig:as_pool_num}
\end{subfigure}
\begin{subfigure}[b]{0.24\textwidth}
  \includegraphics[width=1.0\linewidth]{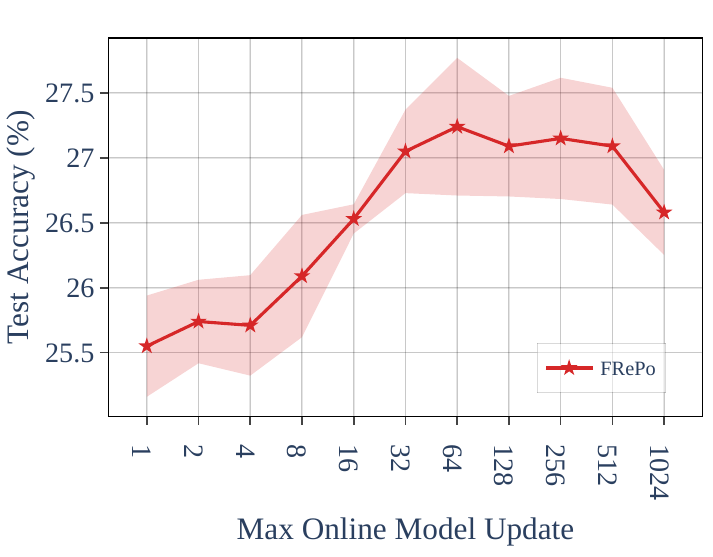}
  \caption{Max Online Updates} \label{fig:as_pool_max}
\end{subfigure}
\begin{subfigure}[b]{0.24\textwidth}
  \includegraphics[width=1.0\linewidth]{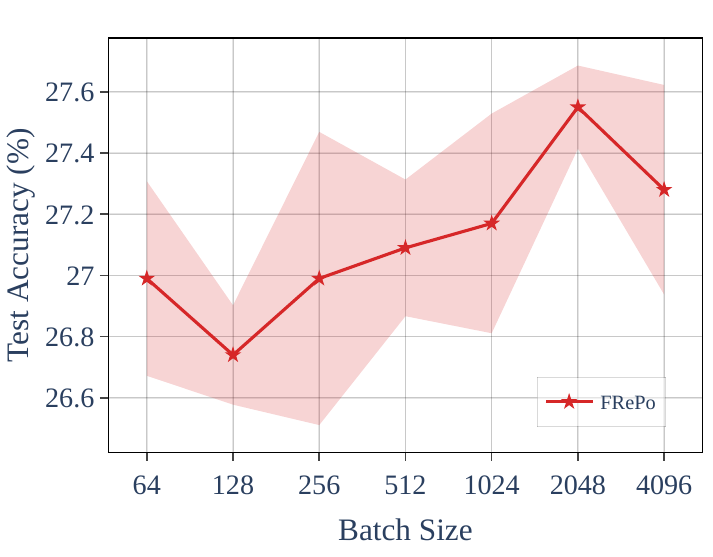} 
  \caption{Batch Size} \label{fig:as_bs}
\end{subfigure}
\begin{subfigure}[b]{0.24\textwidth}
  \includegraphics[width=1.0\linewidth]{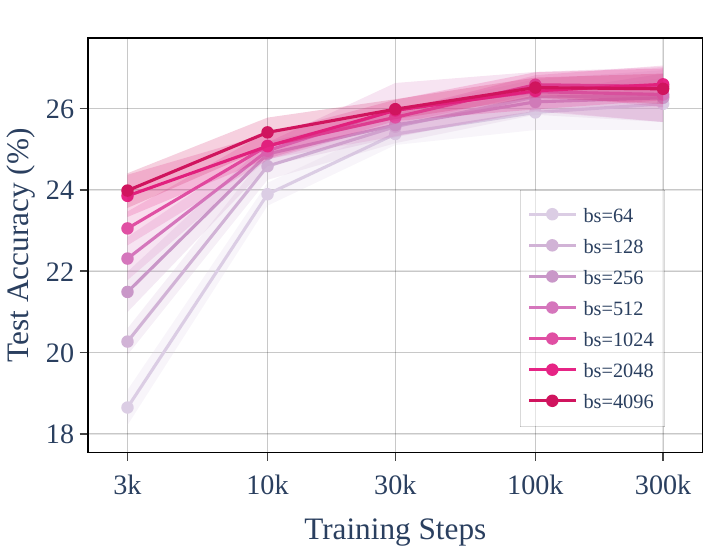}
  \caption{Batch Size} \label{fig:as_bs_step}
\end{subfigure}
\caption{Ablation Study - Model Pool and Batch Size. Our default value for number of online models, max online updates, and batch size may not yield the best performance for all datasets and settings. It provides a good starting point for further investigation.}\label{fig:as_pool}
\end{figure}

We investigate how the number of online models, the maximum online updates, and batch size affect the test accuracy. When one hyperparameter is modified, all the other hyperparameters are unchanged at their default value. As shown in Figure \ref{fig:as_pool_num}, using more than one model is better than using only one model, which is equivalent to the training and resetting strategy used in the previous methods \cite{DBLP:conf/iclr/DC, DBLP:conf/icml/DSA, DBLP:journals/corr/SucholutskyIDD19}. Our default choice of using ten models seems not to be the best but gives reasonable performance. Figure \ref{fig:as_pool_max} shows that both using too few updates and using too many updates hurt the performance, and our default choice of 100 gives a good starting point. If we want to squeeze the performance, it is worth tuning these two hyperparameters. Intuitively, a small regularization strength is needed when we distill a small number of distilled data. On the contrary, when we distill more distilled data, we may want to use a stronger regularization. Figure \ref{fig:as_bs}, \ref{fig:as_bs_step} show that using a large batch size may give slightly better results and converge faster in terms of number of training steps. However, time per step is also an increasing function of batch size, so a large batch size may not give the best test accuracy and efficiency trade-off. Our default choice of 1024 may not be optimal, but it is a good starting point.

\subsection{Initialization} \label{as:initialization}
\begin{figure}[t!]
\centering
\begin{subfigure}[b]{0.24\textwidth}
  \includegraphics[width=1.0\linewidth]{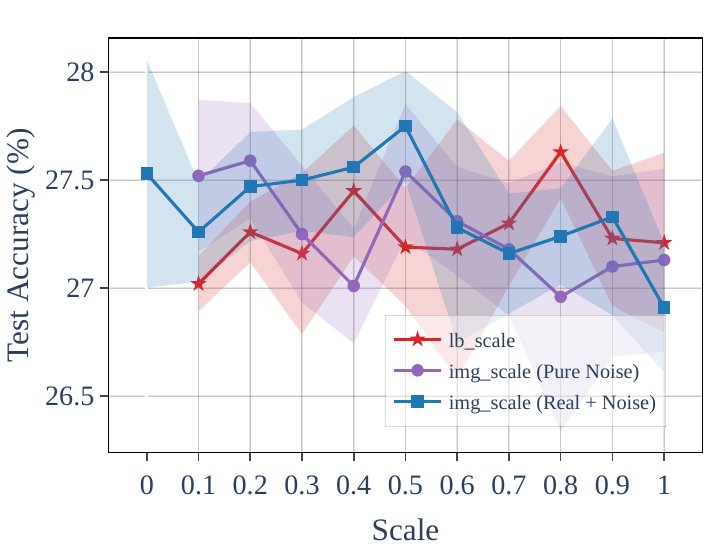}  
  \caption{Overview} \label{fig:as_init_overview}
\end{subfigure}
\begin{subfigure}[b]{0.24\textwidth}
  \includegraphics[width=1.0\linewidth]{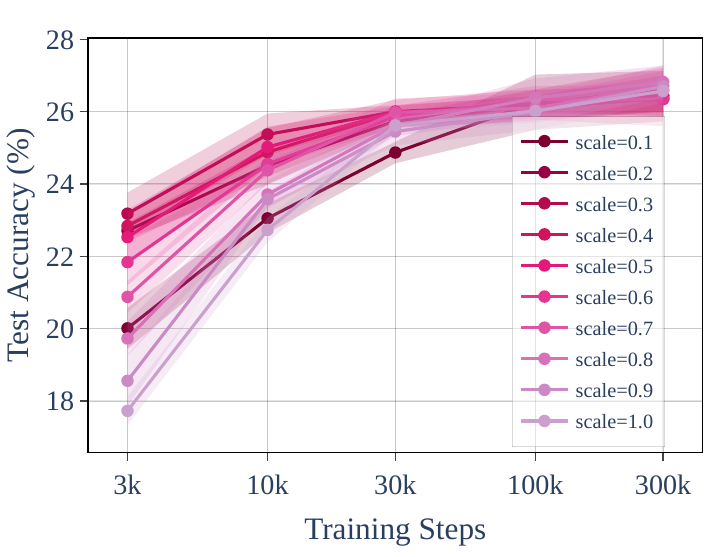}
  \caption{Label Scale} \label{fig:as_init_lb_step}
\end{subfigure}
\begin{subfigure}[b]{0.24\textwidth}
  \includegraphics[width=1.0\linewidth]{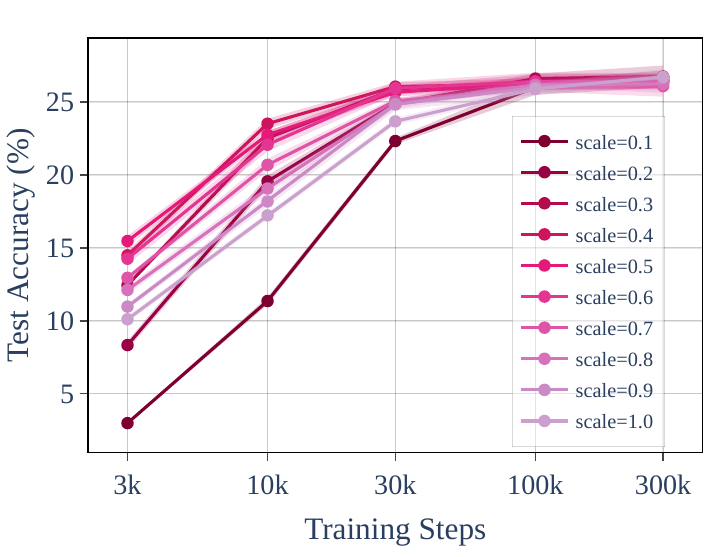}  
  \caption{Random Image} \label{fig:as_init_random_step}
\end{subfigure}
\begin{subfigure}[b]{0.24\textwidth}
  \includegraphics[width=1.0\linewidth]{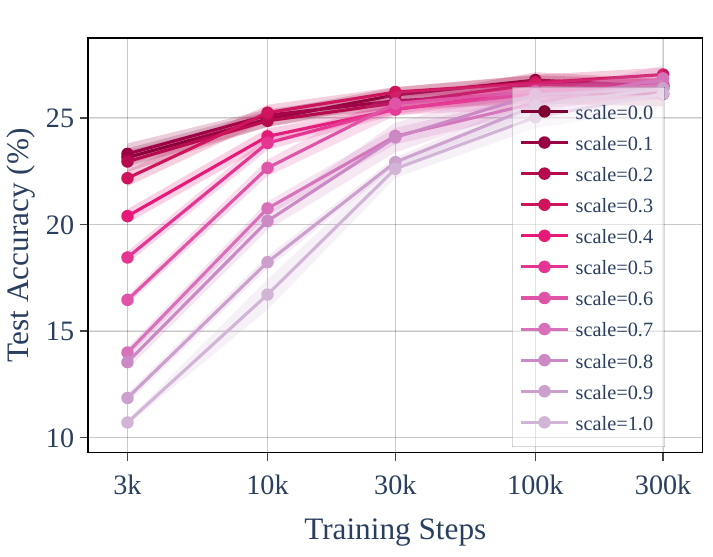}
  \caption{Real Image} \label{fig:as_init_real_step}
\end{subfigure}
\caption{Ablation Study - Initialization. Initializing the distilled image using real images does not explain the effectiveness of our algorithm. Indeed, initializing using the combination of real image and a properly chosen random Gaussian noise gives the best performance. The scale of the label and random Gaussian noise is very crucial for convergence speed.}\label{fig:as_init}
\end{figure}

\begin{figure}[t!]
\centering
\begin{subfigure}[b]{0.47\textwidth}
  \includegraphics[width=1.0\linewidth]{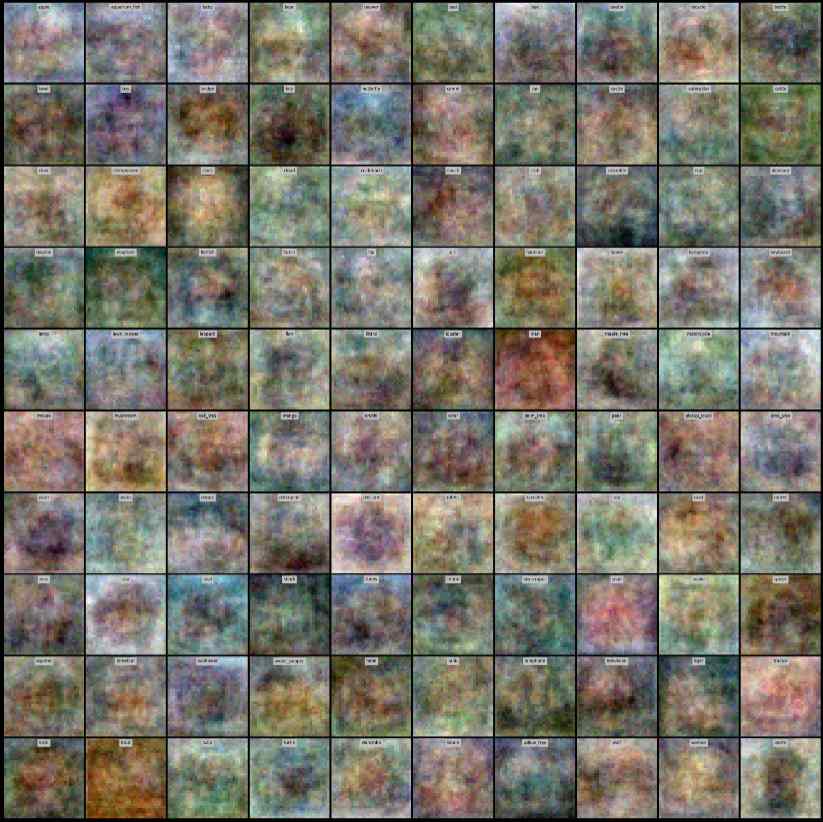}  
  \caption{Before Optimization} \label{fig:as_init_random_before}
\end{subfigure}
\hspace{0.2in}
\begin{subfigure}[b]{0.47\textwidth}
  \includegraphics[width=1.0\linewidth]{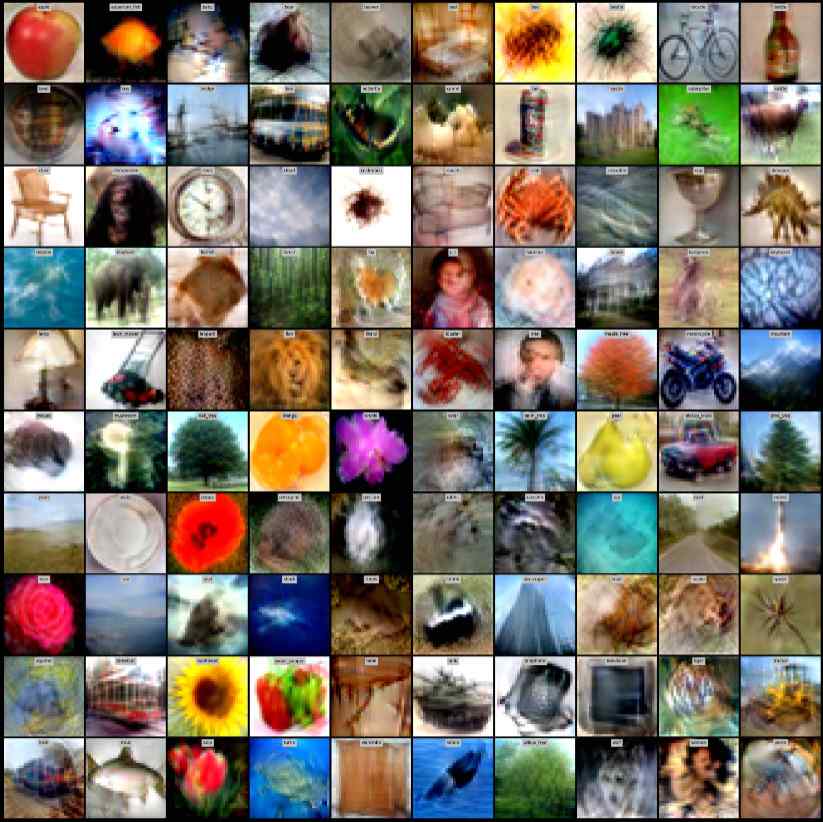}
  \caption{After Optimization} \label{fig:as_init_random_after}
\end{subfigure}
\caption{Ablation Study - Initialize the distilled image using random Gaussian noise.}\label{fig:as_init_random}
\end{figure}

\begin{figure}[t!]
\centering
\begin{subfigure}[b]{0.47\textwidth}
  \includegraphics[width=1.0\linewidth]{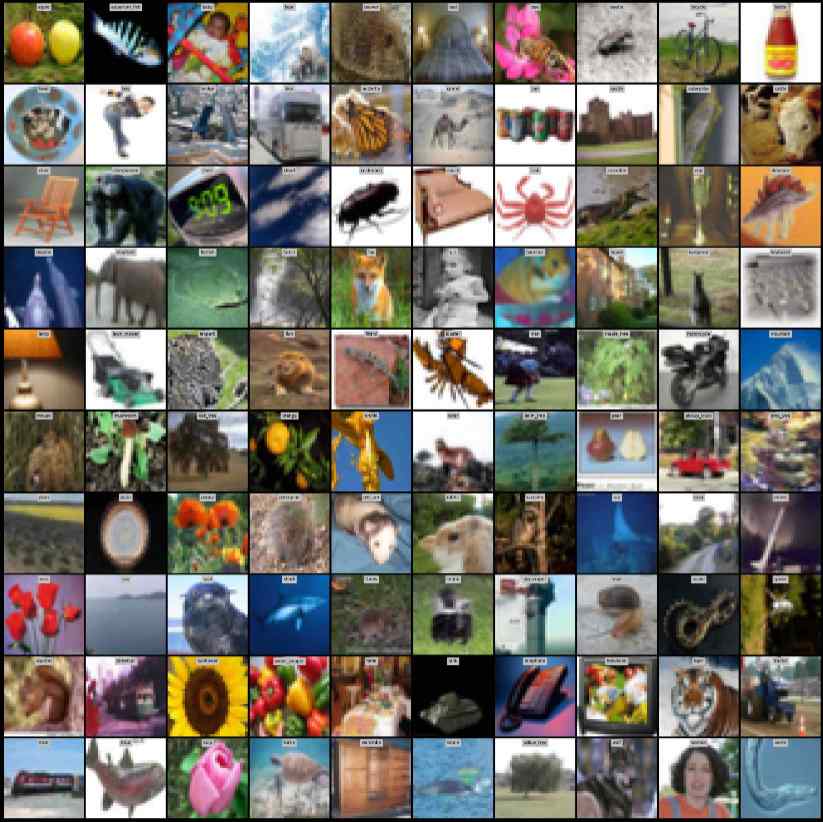}  
  \caption{Before Optimization} \label{fig:as_init_real_before}
\end{subfigure}
\hspace{0.2in}
\begin{subfigure}[b]{0.47\textwidth}
  \includegraphics[width=1.0\linewidth]{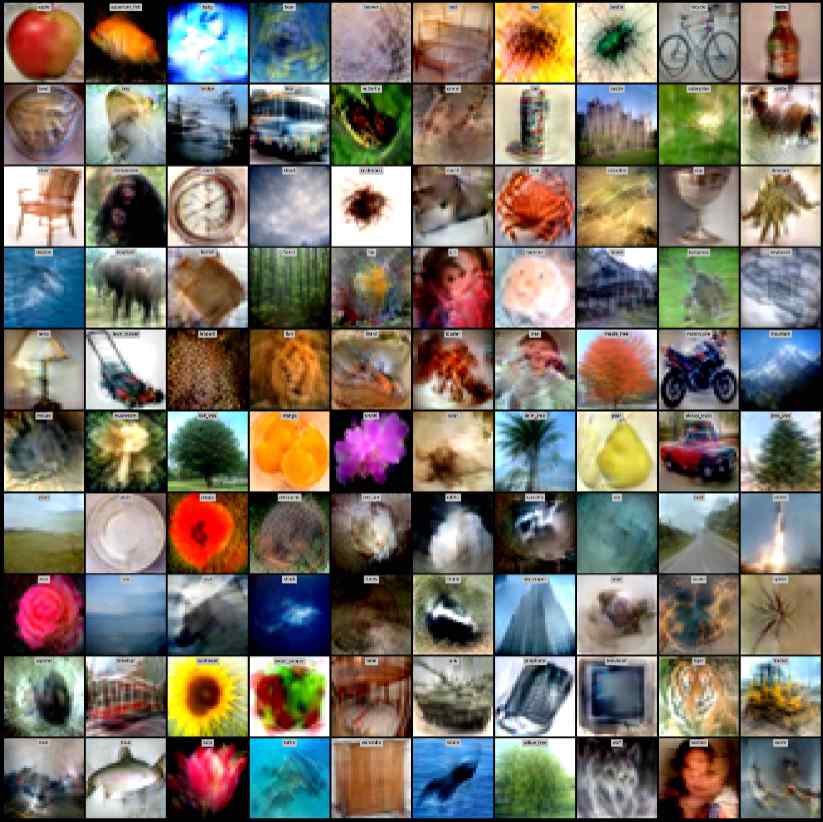}
  \caption{After Optimization} \label{fig:as_init_real_after}
\end{subfigure}
\caption{Ablation Study - Initialize the distilled image using Real images.}\label{fig:as_init_real}
\end{figure}

We initialize the distilled image using real images and initialize the distilled label using a mean-centered one-hot vector scaled by $1/(\sqrt{\numclass/10})$ as default. Is this real initialization that explains why our distilled images look real and natural? Does random initialization give a similar result? How does the label initialization affect the performance? We answer these questions by trying different initialization schemes. We investigate initializing the distilled image from random Gaussian noise or a combination of real images and random noises. We also vary the scaling factor of the label initialization and the noise scale to figure out the best setting for dataset distillation. Figure \ref{fig:as_init_overview} shows that initializing from random noise is not a problem and the best initialization scheme is the combination of real images and random noises. With a properly chosen standard deviation of 0.5, we can achieve 27.8\% test accuracy compared to 27.2\% using the default hyperparameter. Figure \ref{fig:as_init_lb_step}, \ref{fig:as_init_random_step},\ref{fig:as_init_real_step} show that a right scale of noise or label can significantly improve the convergence speed. Besides, Figure \ref{fig:as_init_lb_step} shows that scaling the mean-centered one-hot vector by 0.3 is a good choice for CIFAR100, and we also find that 1.0 works well for datasets with ten classes. Therefore, we decide to use the mean-centered one-hot vector scaled by $1/(\sqrt{\numclass/10})$ as our default label initialization. As shown in Figure \ref{fig:as_init_random}, \ref{fig:as_init_real}, though the distilled images look very different at initialization, they look quite similar after optimization. We also provide four videos to visualize the evolution of the distilled images in the supplementary.

\subsection{Label Learning} \label{as:labellearning}
\begin{figure}[t!]
\centering
\begin{subfigure}[b]{0.47\textwidth}
  \includegraphics[width=1.0\linewidth]{figure/7-vis_jpeg/image_init/real_llTrue_step500000.jpg}  
  \caption{Learn Label=True} \label{as_ll_true}
\end{subfigure}
\hspace{0.2in}
\begin{subfigure}[b]{0.47\textwidth}
  \includegraphics[width=1.0\linewidth]{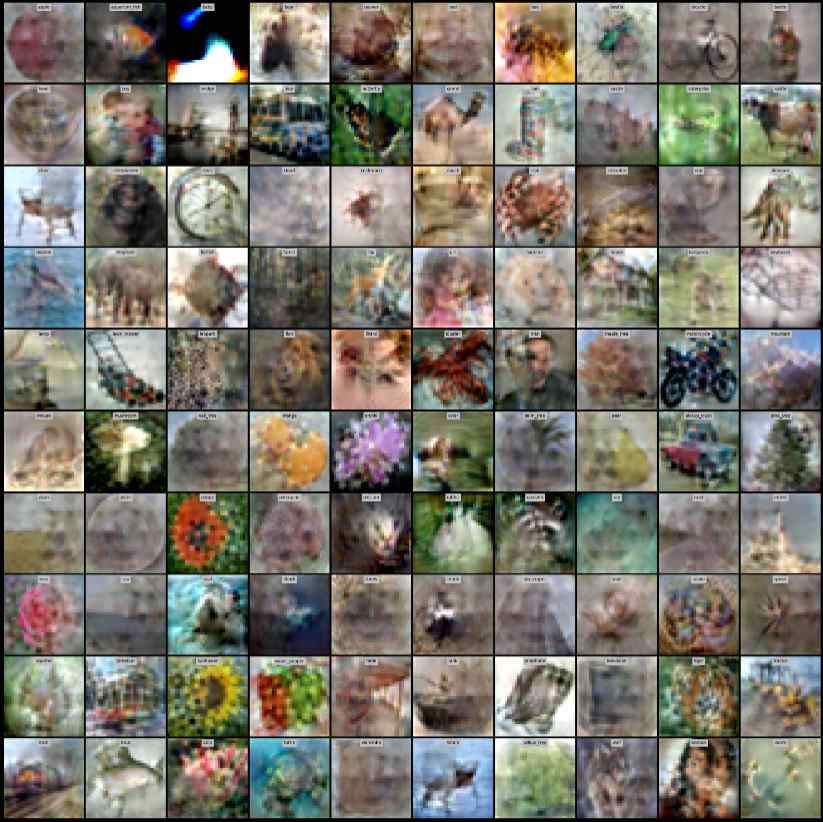}
  \caption{Learn Label=False} \label{fig:as_ll_false}
\end{subfigure}
\caption{Ablation Study - Label Learning. Learning label can generate more natural and real-looking images.}\label{fig:as_ll}
\end{figure}

As discussed in Section \ref{sec:benchmark}, label learning is an essential component of our algorithm. We provide a detailed comparison of learning and not learning labels in Table \ref{tab:dd_app_sota}. We observe that when the label space is simple, such as MNIST, F-MNIST, and CIFAR10, label learning may not be necessary (Figure \ref{fig:vis_mnist}, \ref{fig:vis_fmnist}, \ref{fig:vis_cifar10}). However, it becomes crucial for complex datasets with many classes, such as CIFAR100 (Figure \ref{fig:as_ll}), Tiny-ImageNet (Figure \ref{fig:vis_timagenet}), and ImageNet (Figure \ref{fig:vis_imagenet1}, \ref{fig:vis_imagenet2}). For ImageNet, we can achieve 7.5\% test accuracy when we learn the label, compared to 1.6\% when we fix the label. Besides the fact that the distilled label encodes rich information regarding the class similarity, we find that learning labels can make the optimization easier and distill more natural and real-looking images, as shown in Figure \ref{fig:as_ll}, \ref{fig:vis_timagenet} and supplementary videos.

\begin{table}[t]
  \caption{Test accuracies of models trained on the distilled data from scratch. We highlight the best test accuracy using neural network predictor either learn the label or not.}
  \label{tab:dd_app_sota}
  \small
  \centering
    \begin{tabular}{cccccc}
        \toprule
        &  & \multicolumn{2}{c}{Learn Label = True} & \multicolumn{2}{c}{Learn Label = False} \\
        \cmidrule(l){3-4} \cmidrule(l){5-6}
        & Img/Cls & NN Acc & KRR Acc & NN Acc & KRR Acc \\
        \midrule
        \multirow{3}{*}{MNIST} & 1 & $ 92.5 \pm 0.2$ & $92.6 \pm 0.3$ & $\textbf{93.0} \pm \textbf{0.4}$ & $92.6 \pm 0.4$ \\
         & 10 &  $ \textbf{98.6}\pm \textbf{0.1} $ & $98.6 \pm 0.1$ & $\textbf{98.6}\pm \textbf{0.1}$ & $98.6 \pm 0.1$\\
         & 50 &  $ \textbf{99.2} \pm \textbf{0.0}$ & $99.2 \pm 0.1$ & $\textbf{99.2} \pm \textbf{0.0}$ & $99.2 \pm 0.0$\\
        \midrule
        \multirow{3}{*}{F-MNIST} & 1 & $74.2 \pm 0.5$ & $76.4 \pm 0.3$ & $\textbf{75.6} \pm \textbf{0.2}$ & $77.1 \pm 0.2$ \\
         & 10 & $\textbf{86.2} \pm \textbf{0.1} $ & $86.8 \pm 0.1$ & $86.0 \pm 0.1$ & $86.6 \pm 0.1$ \\
         & 50 & $89.4 \pm 0.1$ & $89.9 \pm 0.1$& $ \textbf{89.6}\pm \textbf{0.1}$ & $89.9 \pm 0.1$\\
        \midrule
        \multirow{3}{*}{CIFAR10} & 1 & $45.5 \pm 0.9$ & $46.3 \pm 0.7$ & $ \textbf{46.8} \pm \textbf{0.7} $ & $47.9 \pm 0.6$\\
         & 10 & $ \textbf{65.5}\pm \textbf{0.6}$ & $68.0 \pm 0.2$ & $65.4 \pm 0.6$ & $66.9 \pm 0.4$\\
         & 50 & $ \textbf{71.7}\pm \textbf{0.2}$ & $74.4 \pm 0.1$ & $\textbf{71.7}\pm \textbf{0.2}$ & $73.8 \pm 0.2$\\
        \midrule
        \multirow{3}{*}{CIFAR100} & 1 & $ \textbf{28.7}\pm \textbf{0.1}$ & $32.3 \pm 0.1$ & $25.4 \pm 0.1$ & $27.3 \pm 0.1$ \\
         & 10 & $ \textbf{42.5}\pm \textbf{0.2}$ & $44.9 \pm 0.2$ & $39.6 \pm 0.3$ & $41.5 \pm 0.1$ \\
         & 50 & $ \textbf{44.3}\pm \textbf{0.2}$ & $43.0 \pm 0.3$ & $40.1 \pm 0.2$ & $37.0 \pm 0.2$ \\
        \midrule
        \multirow{2}{*}{T-ImageNet} & 1 & $ \textbf{15.4}\pm \textbf{0.1}$ & $19.1 \pm 0.3$ & $12.4 \pm 0.8$ & $15.8 \pm 0.3$ \\
         & 10 & $\textbf{25.4} \pm \textbf{0.2}$ & $26.5 \pm 0.1$ & $21.9 \pm 0.2$  & $22.5 \pm 0.2$ \\
        \midrule
        \multirow{2}{*}{CUB-200} & 1 & $ \textbf{12.4}\pm \textbf{0.2}$ & $13.7 \pm 0.2$ & $7.8 \pm 0.1$ & $9.0 \pm 0.3$ \\
         & 10 & $\textbf{16.8} \pm \textbf{0.1}$ & $16.1 \pm 0.3$ & $4.8 \pm 0.4$  & $1.0 \pm 0.2$ \\
        \midrule
        \multirow{2}{*}{ImageNette} & 1 & $ \textbf{48.1}\pm \textbf{0.7}$ & $50.6 \pm 0.6$ & $43.7 \pm 0.9$ & $46.8 \pm 0.7$ \\
         & 10 & $\textbf{66.5} \pm \textbf{0.8}$ & $67.1 \pm 0.7$ & $64.1 \pm 0.8$  & $65.1 \pm 0.8$ \\
        \midrule
        \multirow{2}{*}{ImageWoof} & 1 & $ 26.7\pm 0.6$ & $31.3 \pm 0.9$ & $\textbf{29.7} \pm \textbf{0.6}$ & $28.2 \pm 0.9$ \\
         & 10 & $\textbf{42.2} \pm \textbf{0.9}$ & $43.5 \pm 0.8$ & $41.7 \pm 0.8$  & $43.3 \pm 0.7$ \\
        \midrule
        \multirow{2}{*}{ImageNet} & 1 & $ \textbf{7.5}\pm \textbf{0.3}$ & $7.2 \pm 0.2$ & $1.6 \pm 0.3$ & $1.1 \pm 0.3$ \\
         & 2 & $\textbf{9.7} \pm \textbf{0.2}$ & $9.5 \pm 0.2$ & $2.0 \pm 0.3$  & $1.7 \pm 0.2$ \\
        \bottomrule
    \end{tabular}
    \vspace{-0.1in}
\end{table}

\subsection{Training Cost Analysis} \label{as:trainingcost}
\begin{figure}
  \centering
\begin{subfigure}[b]{0.47\textwidth}
 \includegraphics[width=1.0\linewidth]{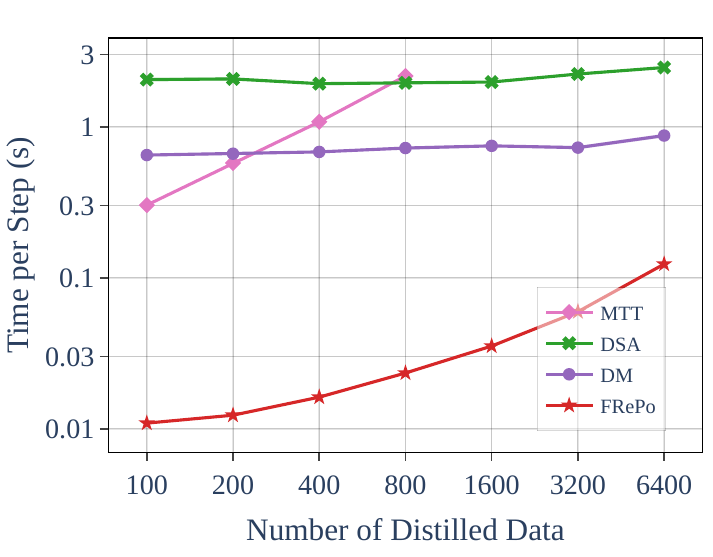}
 \caption{Time Per Step}\label{fig:complexity_timeperstep_size}
\end{subfigure}
\begin{subfigure}[b]{0.47\textwidth}
 \includegraphics[width=1.0\linewidth]{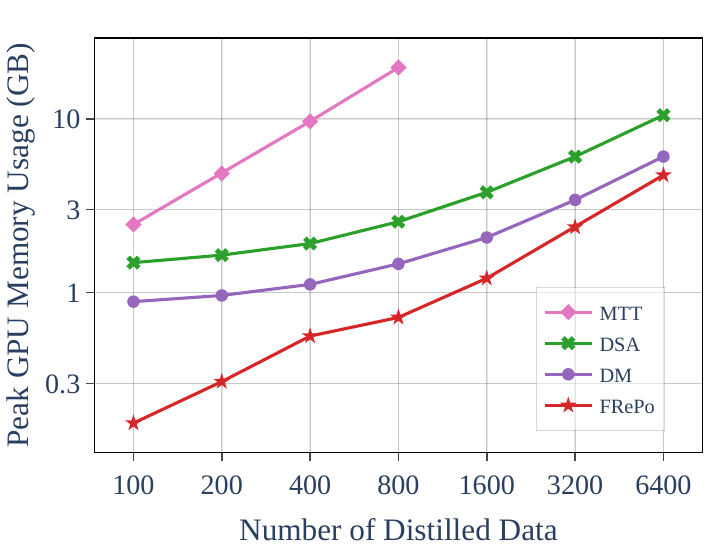}
 \caption{GPU Memory Usage}\label{fig:complexity_memory_size}
\end{subfigure}
\caption{Time per iteration and peak memory usage as we increase the number of distilled data. }\label{fig:complexity_appdix}
\end{figure}

Similar to Figure \ref{fig:complexity}, we also investigate how time per step and GPU memory usage vary when we increase the number of distilled data. As shown in Figure \ref{fig:complexity_appdix}, our method becomes more expensive as we increase the number of distilled data. It is because 1) we always use all the distilled data for gradient computation rather than sample a batch as in other methods; 2) Matrix inversion with time complexity of $O(N^3)$ in KRR becomes more and more expensive as we distill more data. Similar to other kernel methods, distilling tens of thousands of data can be difficult for our method. We can circumvent this problem by 1) sampling a batch at each gradient computation, 2) performing subset distillation, or 3) distillation by class as in Section \ref{sec:cl}. However, the performance is expected to drop because the redundant information can be generated in different groups, and the distilled data may not be able to capture all the distinguishable features when learning from a subset. We leave it for future work to address this scaling challenge. We provide the numerical values for Figure \ref{fig:complexity_timeperstep}, \ref{fig:complexity_memory}, \ref{fig:complexity_timeperstep_size}, and \ref{fig:complexity_memory_size} in Table \ref{tab:complexity_timeperstep}, \ref{tab:complexity_memory}, \ref{tab:complexity_timeperstep_size}, and \ref{tab:complexity_memory_size}.

\begin{table}[htbp]
\centering
\caption{Time per step measure in milliseconds (ms). Corresponds to Figure \ref{fig:complexity_timeperstep}.}
\label{tab:complexity_timeperstep}
\small
\begin{tabular}{ccccc}
\toprule
Width & DSA               & DM               & MTT              & FRePo          \\
\midrule
16    & 1764.3 $\pm$ 71.2 & 569.1 $\pm$ 13.2 & 206.2 $\pm$ 2.5  & 7.7 $\pm$ 0.3  \\
32    & 1742.2 $\pm$ 17.1 & 573.5 $\pm$ 11.2 & 268.6 $\pm$ 2.0  & 6.9 $\pm$ 0.2  \\
64    & 2014.6 $\pm$ 15.0 & 669.8 $\pm$ 10.1 & 299.9 $\pm$ 2.1  & 8.5 $\pm$ 0.2  \\
128   & 2583.5 $\pm$ 14.7 & 950.6 $\pm$ 13.7 & 485.6 $\pm$ 3.0  & 10.9 $\pm$ 0.1 \\
256   & 4909.8 $\pm$ 16.9 & 1569.2 $\pm$ 8.5 & 939.6 $\pm$ 10.5 & 20.8 $\pm$ 0.1 \\
512   & 8764.6 $\pm$ 13.7 & 3209.5 $\pm$ 8.6 & 2153.6 $\pm$ 9.0 & 52.5 $\pm$ 0.1 \\
\bottomrule
\end{tabular}
\end{table}

\begin{table}[htbp]
\centering
\caption{Peak GPU memory usage measured in gigabytes (GB). Corresponds to Figure \ref{fig:complexity_memory}.}
\label{tab:complexity_memory}
\small
\begin{tabular}{ccccc}
\toprule
Width & DSA    & DM    & MTT    & FRePo \\
\midrule
16    & 0.800  & 0.678 & 0.650  & 0.036 \\
32    & 0.986  & 0.714 & 1.260  & 0.086 \\
64    & 1.488  & 0.888 & 2.462  & 0.178 \\
128   & 2.014  & 1.164 & 4.858  & 0.387 \\
256   & 4.594  & 1.718 & 9.790  & 0.814 \\
512   & 10.768 & 2.970 & 19.110 & 2.080 \\
\bottomrule
\end{tabular}
\end{table}

\begin{table}[htbp]
\centering
\caption{Time per step measure in millisecond (ms). Corresponds to Figure \ref{fig:complexity_timeperstep_size}.}
\label{tab:complexity_timeperstep_size}
\small
\begin{tabular}{ccccc}
\toprule
Number of Distilled Data & DSA               & DM               & MTT               & FRePo           \\
\midrule
100                      & 2051.8 $\pm$ 52.2 & 649.4 $\pm$ 11.7 & 302.8 $\pm$ 1.8   & 10.9 $\pm$ 0.1  \\
200                      & 2073.4 $\pm$ 55.4 & 662.9 $\pm$ 15.7 & 575.0 $\pm$ 7.3   & 9.6 $\pm$ 0.1   \\
400                      & 1928.6 $\pm$ 13.6 & 681.4 $\pm$ 10.7 & 1077.5 $\pm$ 8.8  & 16.2 $\pm$ 0.2  \\
800                      & 1952.5 $\pm$ 12.3 & 722.7 $\pm$ 18.2 & 2169.1 $\pm$ 12.5 & 23.4 $\pm$ 0.2  \\
1600                     & 1977.4 $\pm$ 16.4 & 747.2 $\pm$ 13.7 & -                 & 35.3 $\pm$ 0.2  \\
3200                     & 2233.8 $\pm$ 8.8  & 727.3 $\pm$ 18.8 & -                 & 59.6 $\pm$ 0.1  \\
6400                     & 2467.8 $\pm$ 8.8  & 874.2 $\pm$ 25.7 & -                 & 123.2 $\pm$ 0.1 \\
\bottomrule
\end{tabular}
\end{table}

\begin{table}[htbp]
\centering
\caption{Peak GPU memory usage measured in gigabytes (GB). Corresponds to Figure \ref{fig:complexity_memory_size}.}
\label{tab:complexity_memory_size}
\small
\begin{tabular}{ccccc}
\toprule
Number of Distilled Data & DSA    & DM    & MTT    & FRePo \\
\midrule
100                      & 1.488  & 0.888 & 2.464  & 0.178 \\
200                      & 1.644  & 0.964 & 4.848  & 0.308 \\
400                      & 1.916  & 1.116 & 9.662  & 0.563 \\
800                      & 2.558  & 1.464 & 19.700 & 0.721 \\
1600                     & 3.776  & 2.076 & -      & 1.210 \\
3200                     & 6.056  & 3.412 & -      & 2.380 \\
6400                     & 10.482 & 6.060 & -      & 4.740 \\
\bottomrule
\end{tabular}
\end{table}

\subsection{Model Architectures} \label{as:model}
From the Cross-Architecture Generalization experiments in Section \ref{sec:benchmark}, we observe that the distilled data can encode the architecture's inductive bias. Thus, we perform a qualitative and quantitative comparison of the distilled data generated by different architectures to understand how architecture affects the distilled data. We evaluate across various architectures, including Conv (Our default model), DCConv (the default model of DSA \cite{DBLP:conf/icml/DSA}, DM \cite{DBLP:journals/corr/DM}, MTT \cite{DBLP:journals/corr/MTT}), AlexNet \citep{DBLP:conf/nips/AlexNet}, VGG \cite{DBLP:journals/corr/VGG}, and ResNet \cite{DBLP:conf/cvpr/ResNet}. We also consider a wide range range of normalization layers, such as no normalization (NN), Instance Normalization (IN) \cite{DBLP:journals/corr/UlyanovVL16}, Batch Normalization (BN) \cite{DBLP:conf/icml/IoffeS15}, Layer Normalization (LN) \cite{DBLP:journals/corr/BaKH16}, and Group Normalization (GN) \cite{DBLP:conf/eccv/WuH18}. Besides, we also vary the depth of Conv and denote two-layer, three-layer, and four-layer Conv as Conv-BN-D2, Conv-BN-D3, and Conv-BN-D4, respectively. We do not rescale any image for better visualization, so the over-saturated images indicate that the images have different statistics from the real images. 

\begin{figure}[htbp]
\centering
\begin{subfigure}[b]{0.32\textwidth}
  \includegraphics[width=1.0\linewidth]{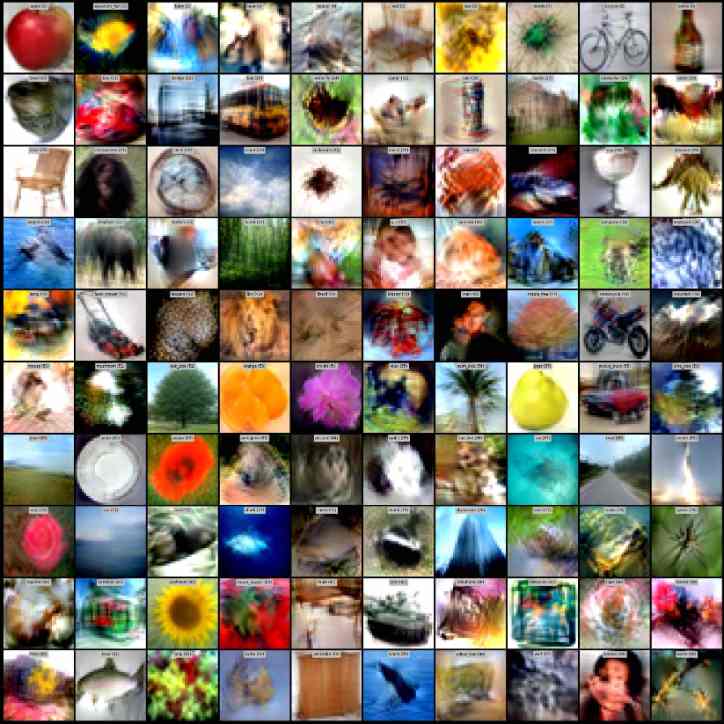}  
  \caption{Conv-NN} 
\end{subfigure}
\begin{subfigure}[b]{0.32\textwidth}
  \includegraphics[width=1.0\linewidth]{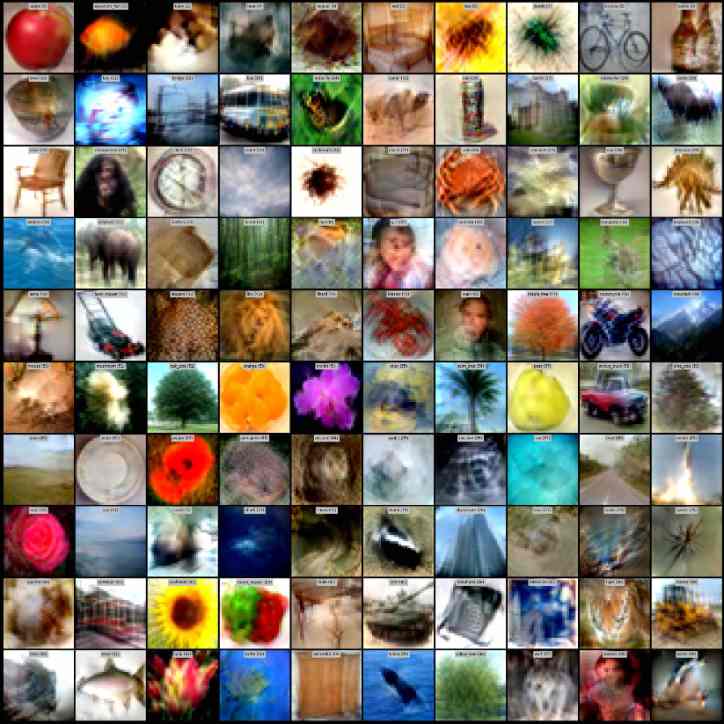}
  \caption{Conv-BN} 
\end{subfigure}
\begin{subfigure}[b]{0.32\textwidth}
  \includegraphics[width=1.0\linewidth]{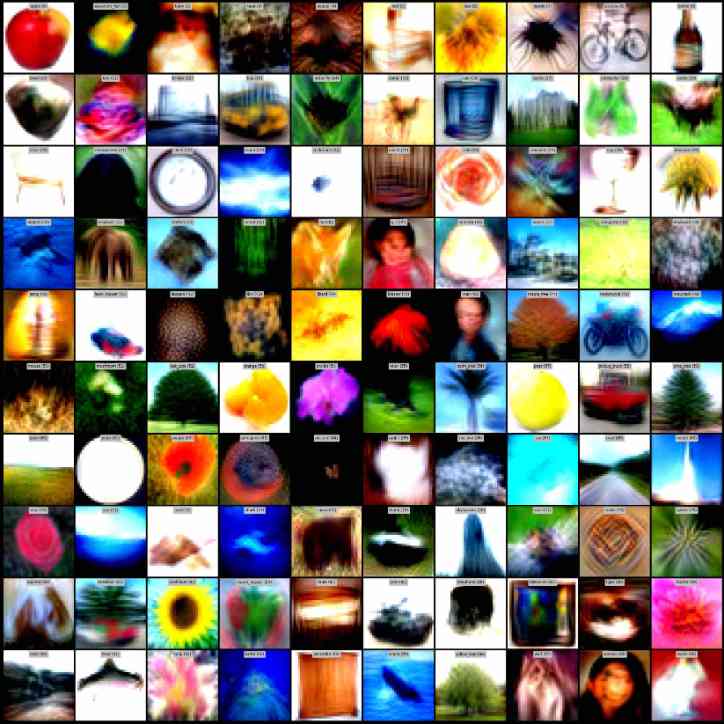}
  \caption{Conv-LN} 
\end{subfigure}
\begin{subfigure}[b]{0.32\textwidth}
  \includegraphics[width=1.0\linewidth]{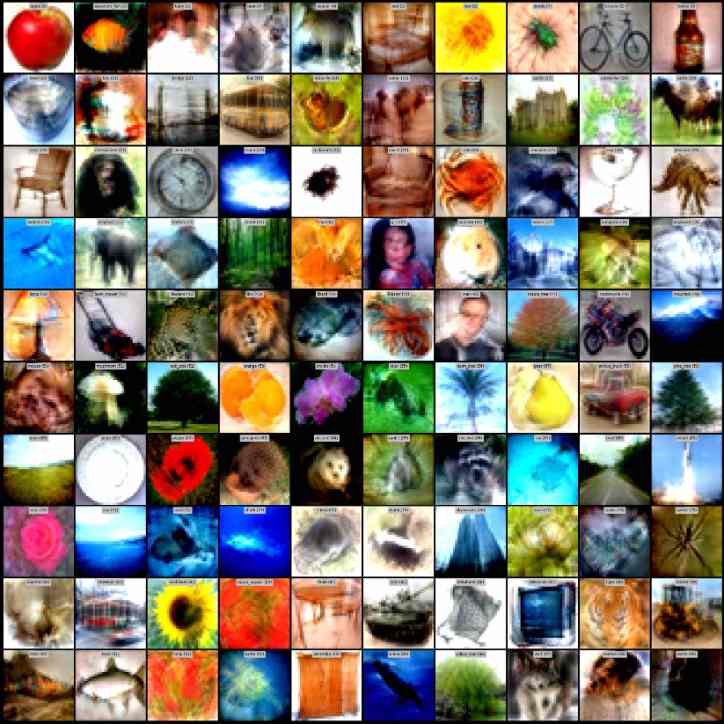}
  \caption{Conv-IN} 
\end{subfigure}
\begin{subfigure}[b]{0.32\textwidth}
  \includegraphics[width=1.0\linewidth]{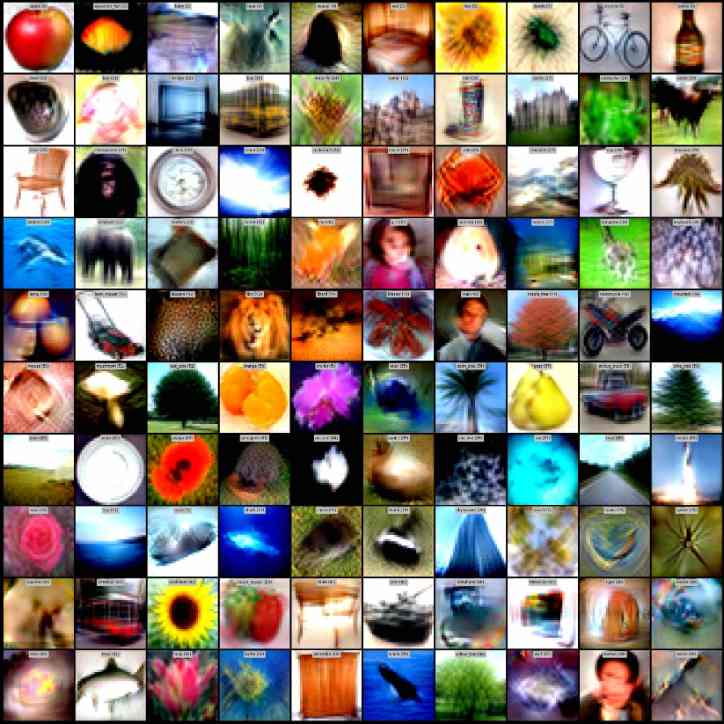}
  \caption{Conv-GN (GroupNum=32)} 
\end{subfigure}
\begin{subfigure}[b]{0.32\textwidth}
  \includegraphics[width=1.0\linewidth]{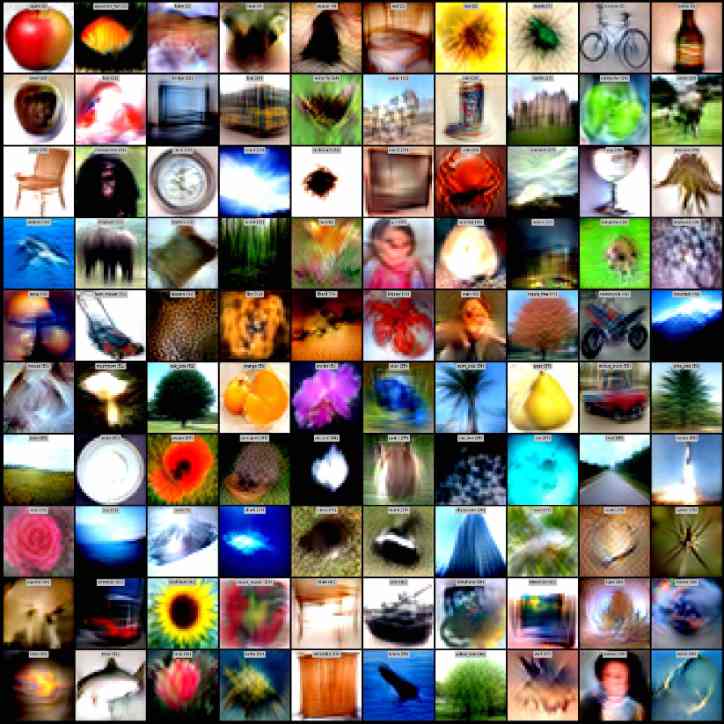}
  \caption{Conv-GN (GroupNum=1)} 
\end{subfigure}
\caption{Ablation Study - Default Model with different Normalization Layer}\label{fig:as_arch_conv}
\end{figure}

\begin{figure}[htbp]
\centering
\begin{subfigure}[b]{0.32\textwidth}
  \includegraphics[width=1.0\linewidth]{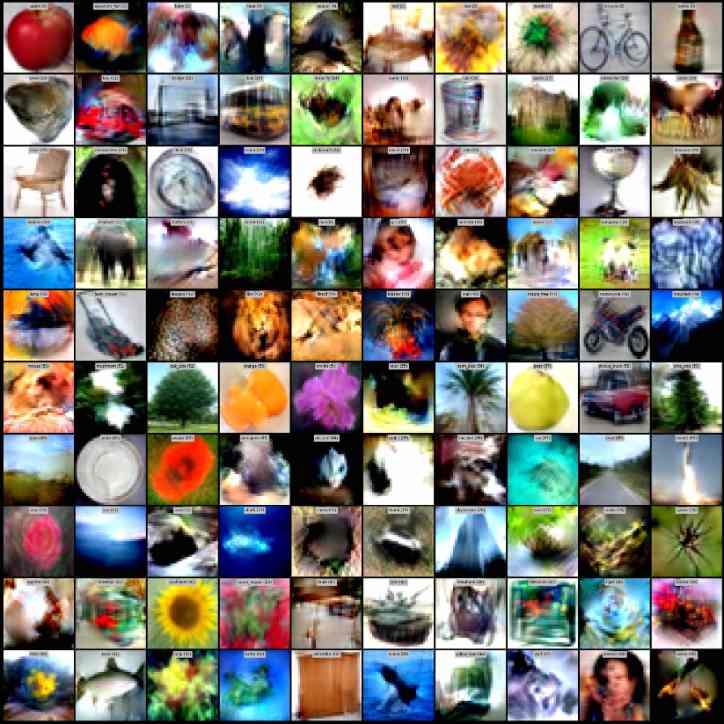}  
  \caption{DCConv-NN} 
\end{subfigure}
\begin{subfigure}[b]{0.32\textwidth}
  \includegraphics[width=1.0\linewidth]{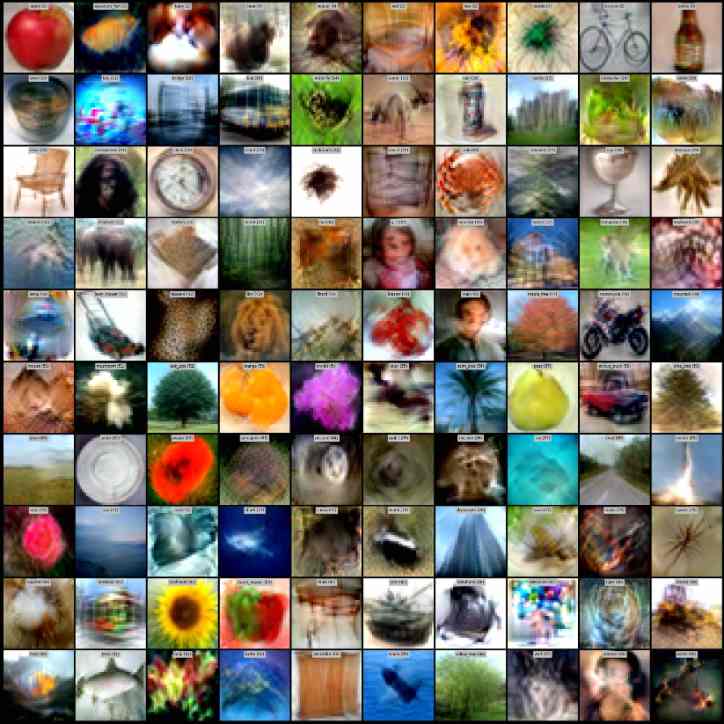}
  \caption{DCConv-BN} 
\end{subfigure}
\begin{subfigure}[b]{0.32\textwidth}
  \includegraphics[width=1.0\linewidth]{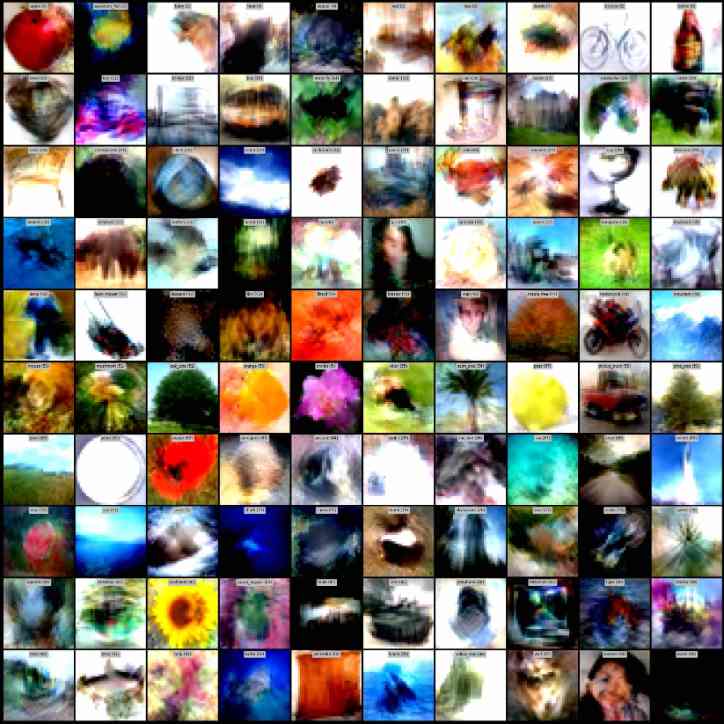}
  \caption{DCConv-LN} 
\end{subfigure}
\begin{subfigure}[b]{0.32\textwidth}
  \includegraphics[width=1.0\linewidth]{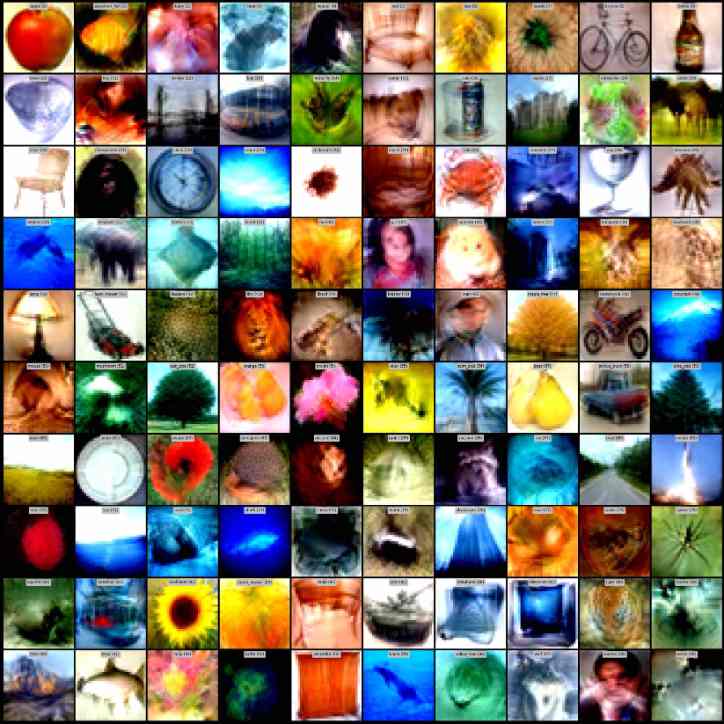}
  \caption{DCConv-IN} 
\end{subfigure}
\begin{subfigure}[b]{0.32\textwidth}
  \includegraphics[width=1.0\linewidth]{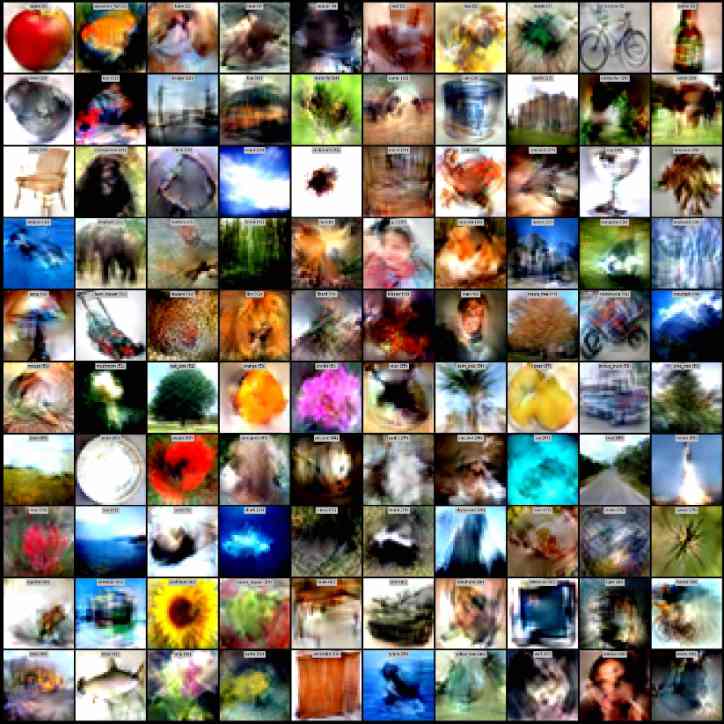}
  \caption{DCConv-GN (GroupNum=32)} 
\end{subfigure}
\begin{subfigure}[b]{0.32\textwidth}
  \includegraphics[width=1.0\linewidth]{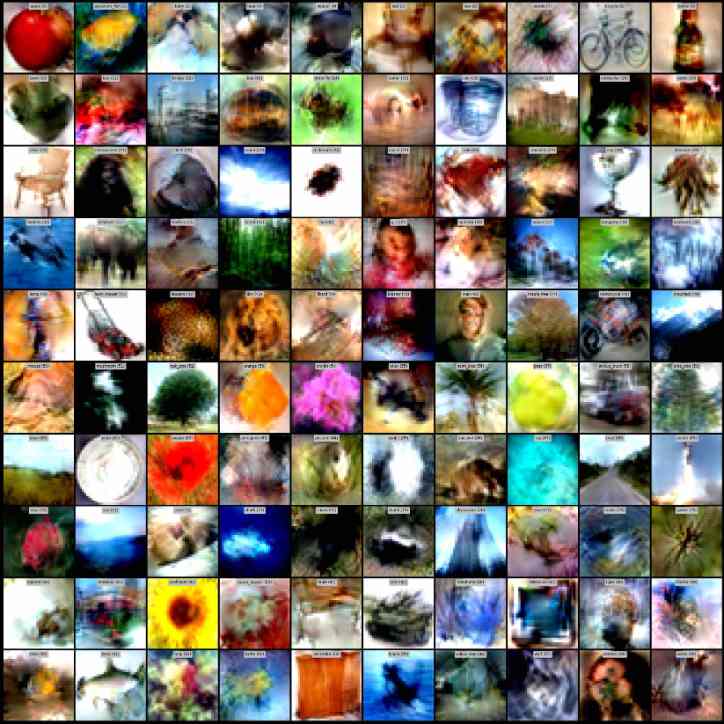}
  \caption{DCConv-GN (GroupNum=1)} 
\end{subfigure}
\caption{Ablation Study - DCConv with different Normalization Layer}\label{fig:as_arch_dcconv}
\end{figure}

\begin{figure}[htbp]
\centering
\begin{subfigure}[b]{0.32\textwidth}
  \includegraphics[width=1.0\linewidth]{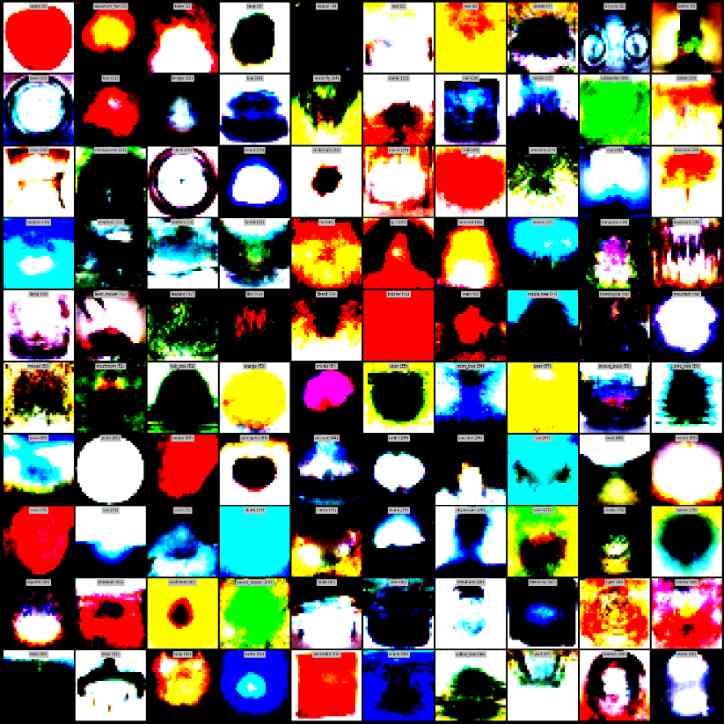}  
  \caption{ResNet-NN} 
\end{subfigure}
\begin{subfigure}[b]{0.32\textwidth}
  \includegraphics[width=1.0\linewidth]{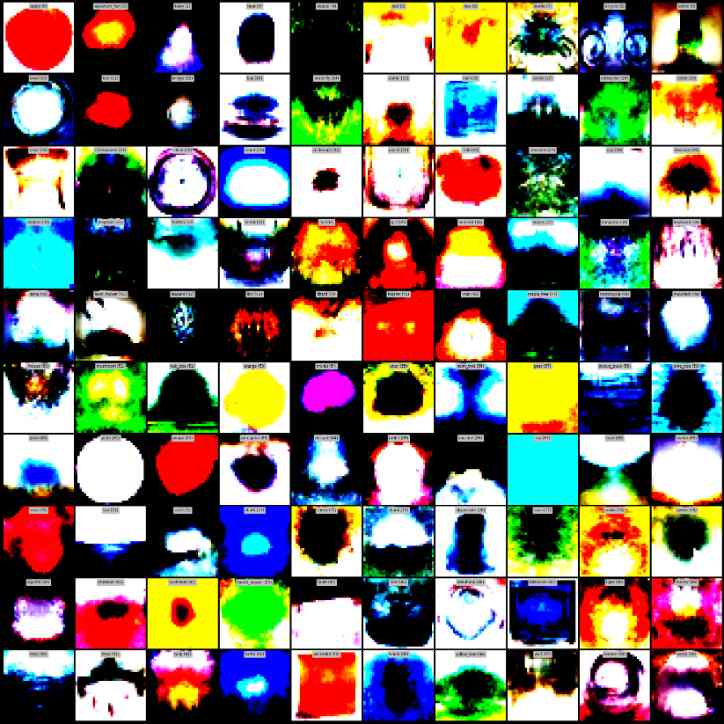}
  \caption{ResNet-BN} 
\end{subfigure}
\begin{subfigure}[b]{0.32\textwidth}
  \includegraphics[width=1.0\linewidth]{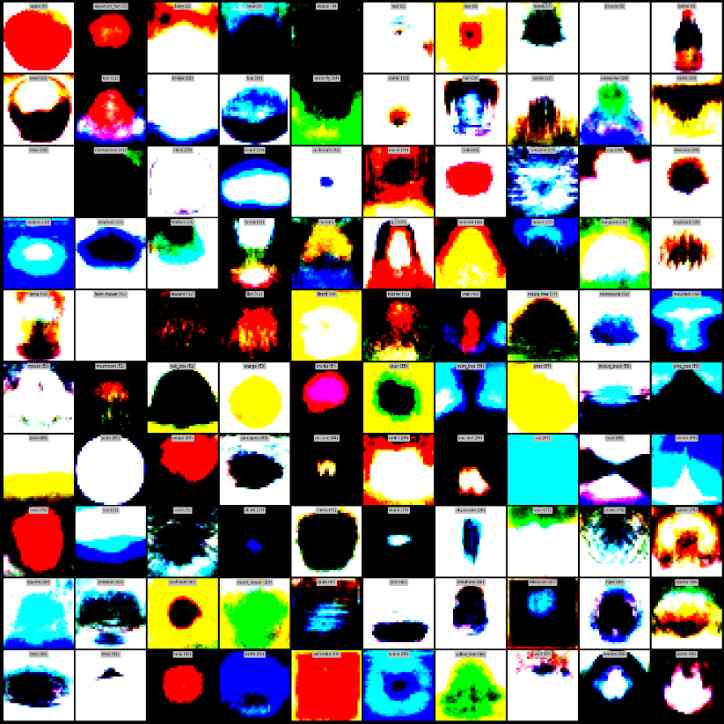}
  \caption{ResNet-LN} 
\end{subfigure}
\begin{subfigure}[b]{0.32\textwidth}
  \includegraphics[width=1.0\linewidth]{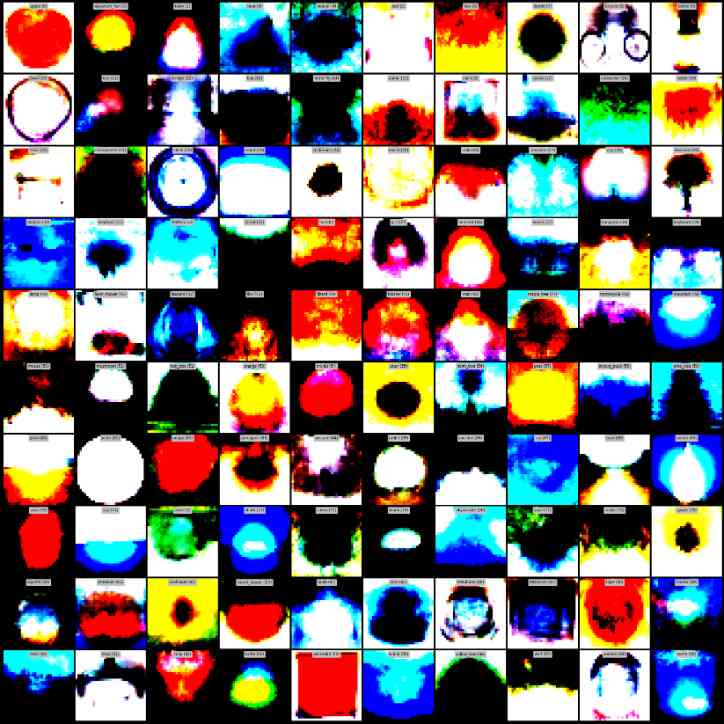}
  \caption{ResNet-IN} 
\end{subfigure}
\begin{subfigure}[b]{0.32\textwidth}
  \includegraphics[width=1.0\linewidth]{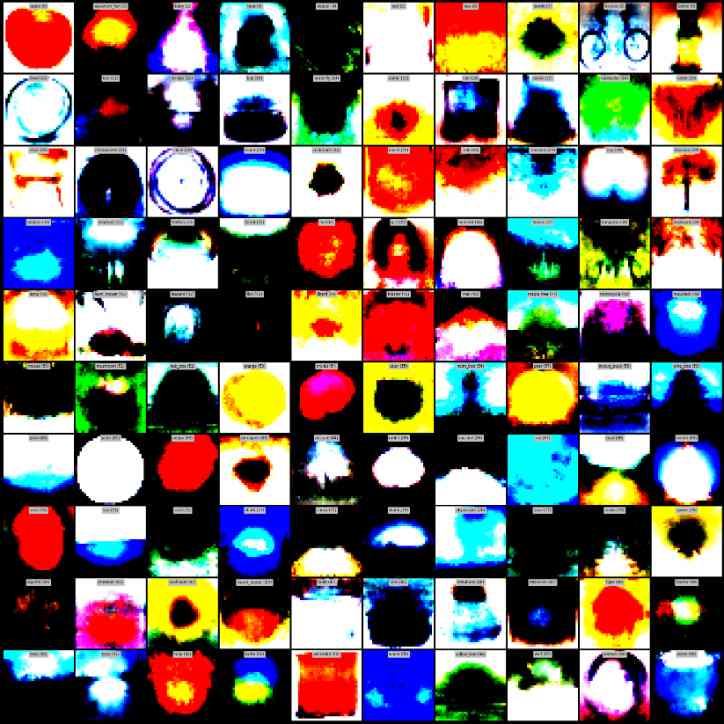}
  \caption{ResNet-GN (GroupNum=32)} 
\end{subfigure}
\begin{subfigure}[b]{0.32\textwidth}
  \includegraphics[width=1.0\linewidth]{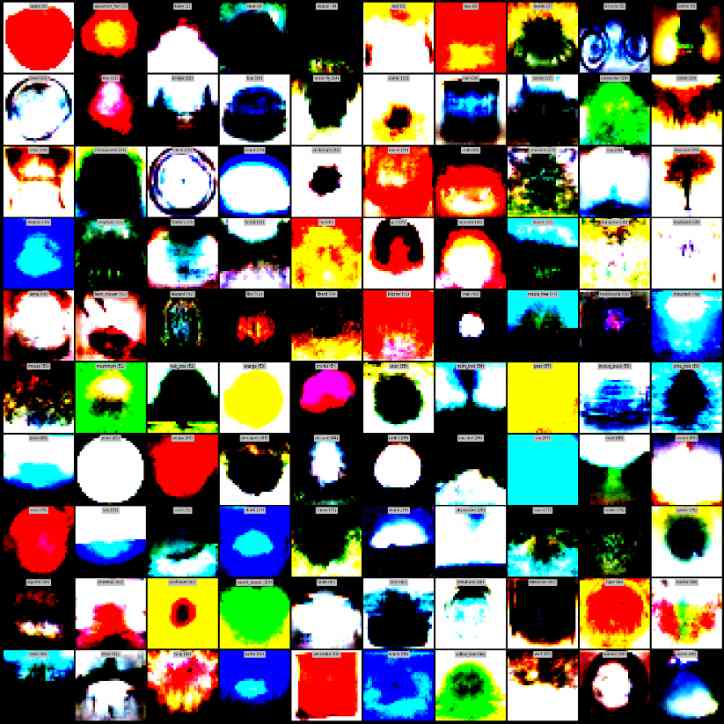}
  \caption{ResNet-GN (GroupNum=1)} 
\end{subfigure}
\caption{Ablation Study - ResNet with different Normalization Layer}\label{fig:as_arch_resnet}
\end{figure}

\begin{figure}[htbp]
\centering
\begin{subfigure}[b]{0.32\textwidth}
  \includegraphics[width=1.0\linewidth]{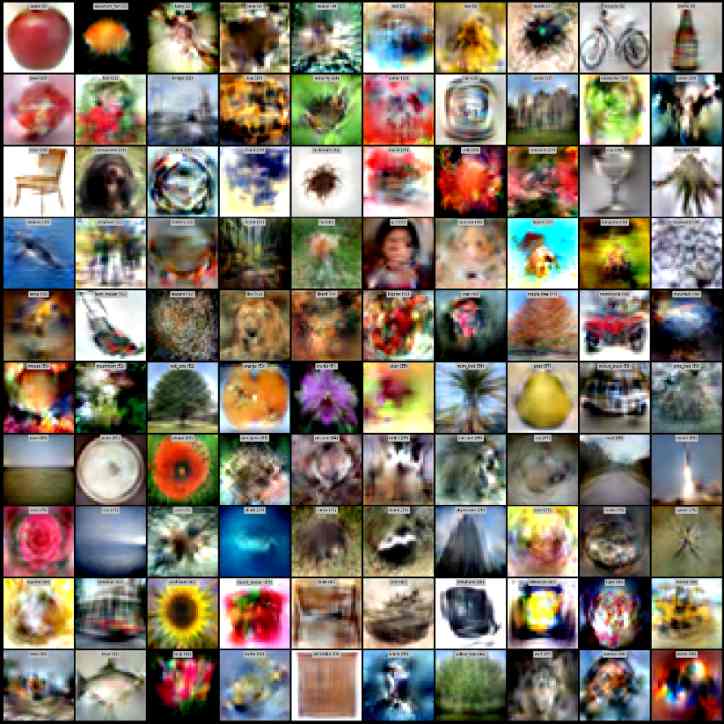}  
  \caption{VGG-NN} 
\end{subfigure}
\begin{subfigure}[b]{0.32\textwidth}
  \includegraphics[width=1.0\linewidth]{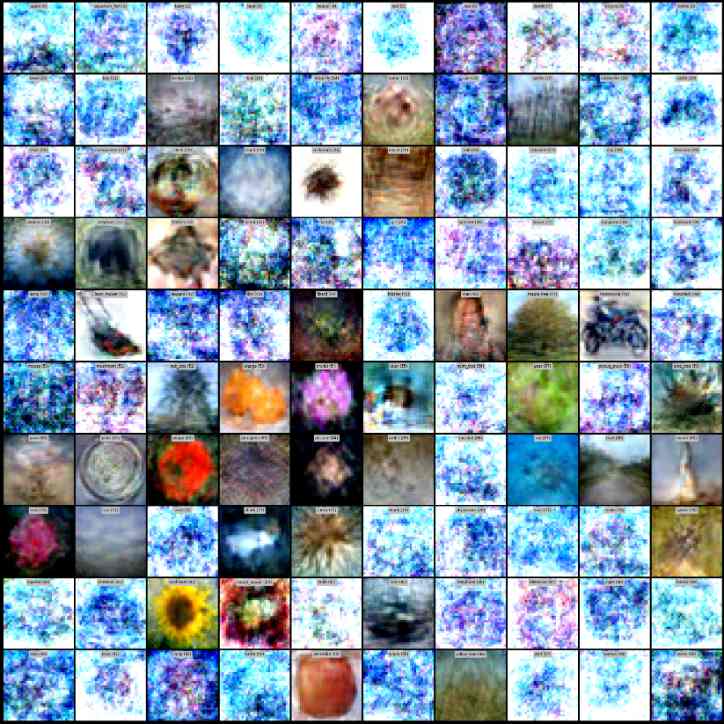}
  \caption{VGG-BN} 
\end{subfigure}
\begin{subfigure}[b]{0.32\textwidth}
  \includegraphics[width=1.0\linewidth]{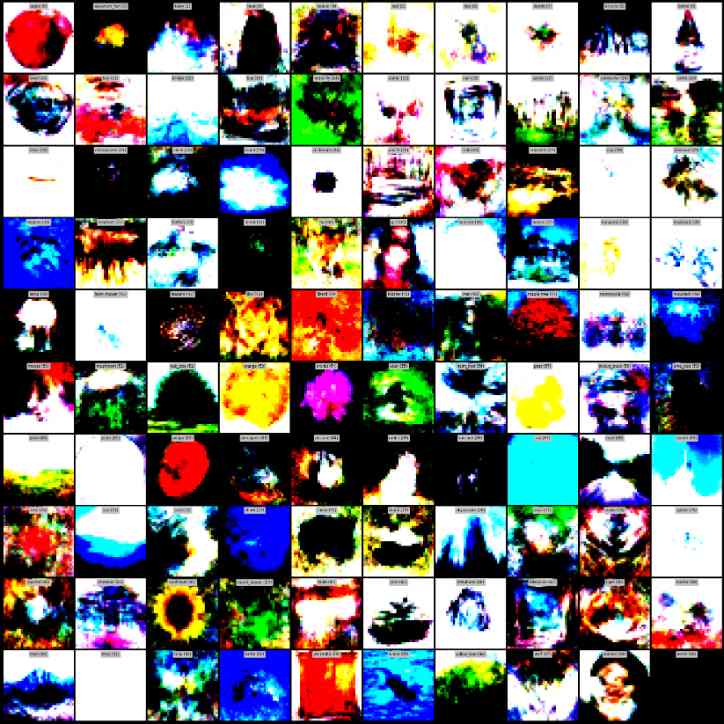}
  \caption{VGG-LN} 
\end{subfigure}
\begin{subfigure}[b]{0.32\textwidth}
  \includegraphics[width=1.0\linewidth]{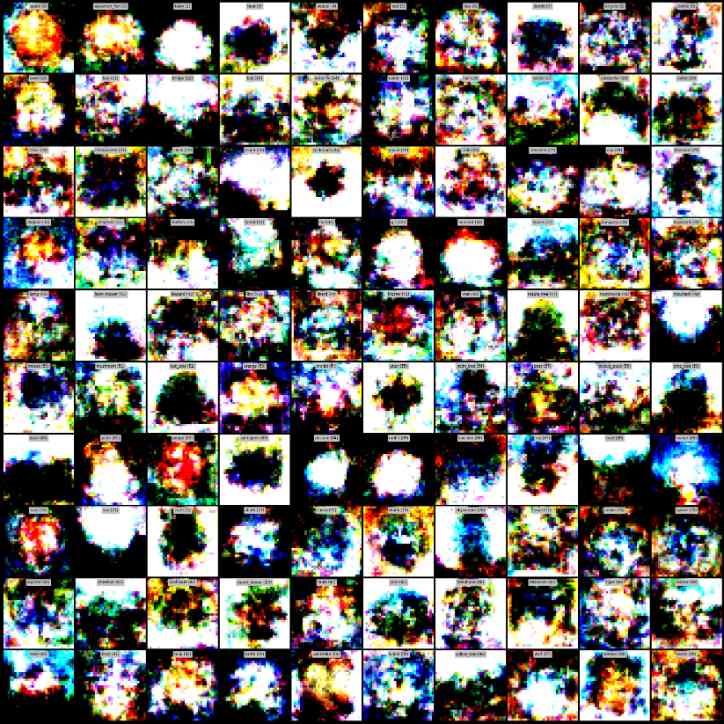}
  \caption{VGG-IN} 
\end{subfigure}
\begin{subfigure}[b]{0.32\textwidth}
  \includegraphics[width=1.0\linewidth]{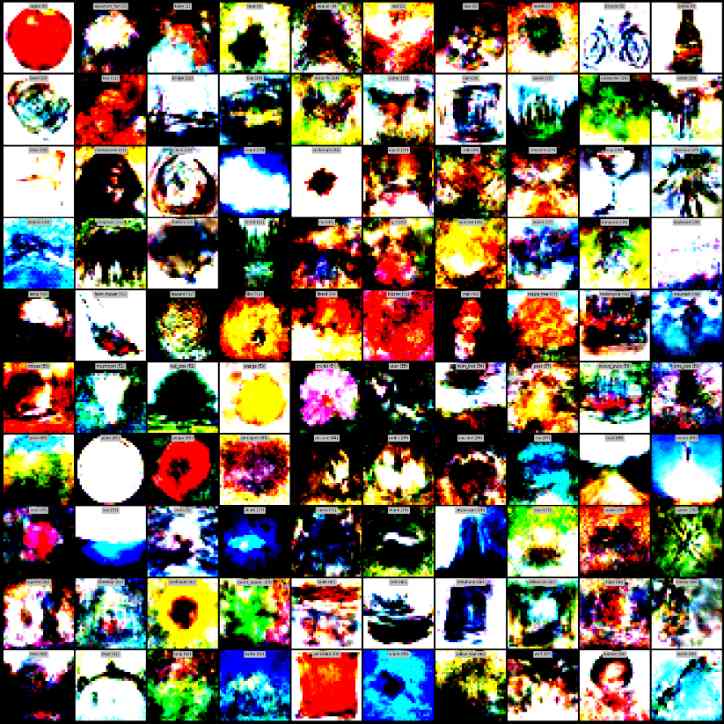}
  \caption{VGG-GN (GroupNum=32)} 
\end{subfigure}
\begin{subfigure}[b]{0.32\textwidth}
  \includegraphics[width=1.0\linewidth]{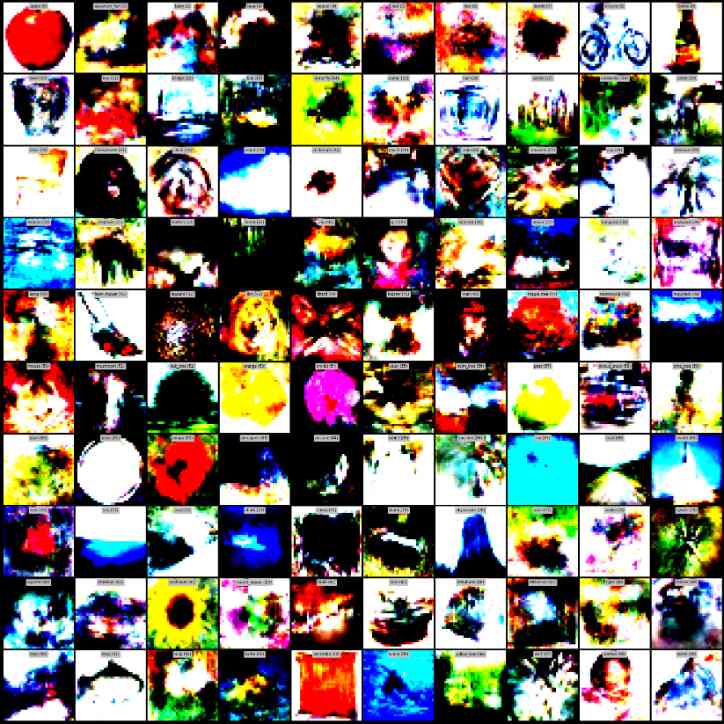}
  \caption{VGG-GN (GroupNum=1)} 
\end{subfigure}
\vspace{-0.05in}
\caption{Ablation Study - VGG with different Normalization Layer.}\label{fig:as_arch_vgg}
\vspace{-0.1in}
\end{figure}

\begin{figure}[htbp]
\centering
\begin{subfigure}[b]{0.32\textwidth}
  \includegraphics[width=1.0\linewidth]{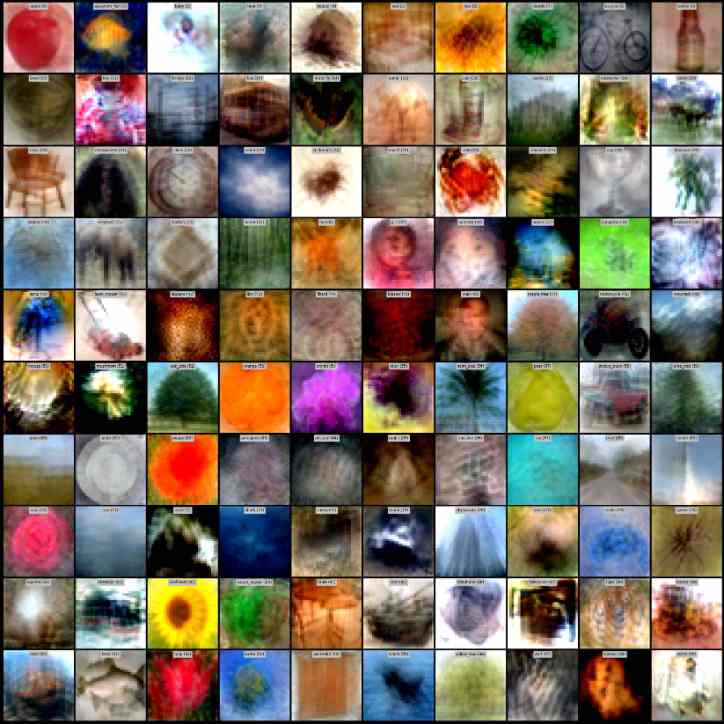}
  \caption{Conv-BN-D2} 
\end{subfigure}
\begin{subfigure}[b]{0.32\textwidth}
  \includegraphics[width=1.0\linewidth]{figure/7-vis_jpeg/image_arch/conv_batch_d3.jpg}
  \caption{Conv-BN-D3} 
\end{subfigure}
\begin{subfigure}[b]{0.32\textwidth}
  \includegraphics[width=1.0\linewidth]{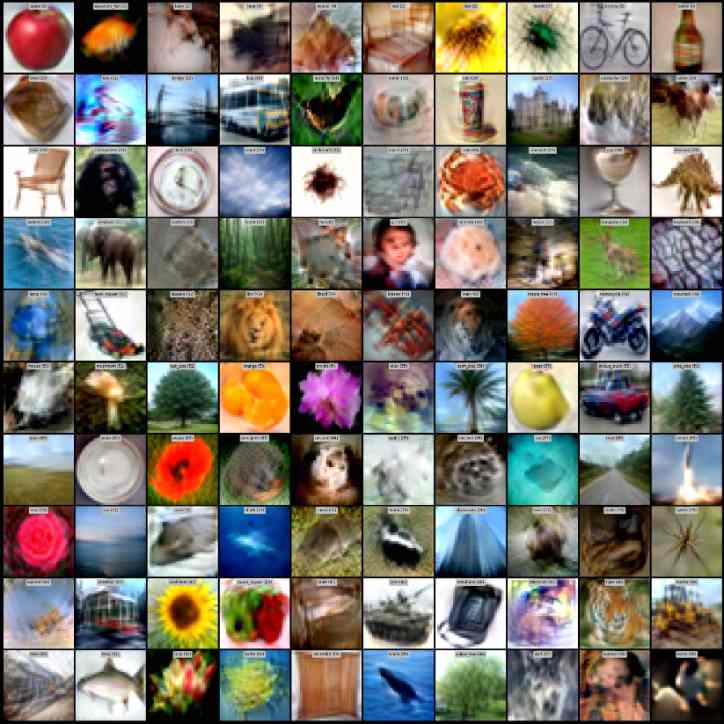}
  \caption{Conv-BN-D4} 
\end{subfigure}
\vspace{-0.05in}
\caption{Ablation Study - Conv with different Depth.}\label{fig:as_arch_conv_depth}
\vspace{-0.1in}
\end{figure}

\textbf{Qualitative Results}: Figure \ref{fig:as_arch_conv}, \ref{fig:as_arch_dcconv}, \ref{fig:as_arch_resnet}, \ref{fig:as_arch_vgg}, and \ref{fig:as_arch_conv_depth} show that 1) the simplest architecture (i.e., Conv) gives the images that look almost like real images; 2) Different normalization layers have different effects on the distilled images, resulting in images with very different brightness and contrasts. No Normalization or Batch Normalization seems to generate the most natural-looking images; 3) The distilled images generated by modern architectures like ResNet and VGG are very different from the natural images. Thus additional attention is needed to transfer the distilled images to a different architecture. On the other hand, additional tricks such as image regularization or projection may help make the distilled images more similar to natural ones. 4) Number of average pooling layers (or size of the final activation map) can affect the distilled image quality. Using a fewer pooling (i.e., Conv-BN-D2) will generate blur images or repeated objects in the images, and we find it is more obvious in high resolution such as Tiny-ImageNet. 5) Distilled images may reflect the similarity of the architecture. For instance, the images generated by Conv and DCConv are almost identical since the only architectural difference is the filter width. Besides, we observe that the images generated by AlexNet are very similar to those generated by Conv, which suggests that the inductive bias of those two architectures is very similar. It may be one reason why Conv's distilled images work extremely well for AlexNet (Table \ref{tab:dd_ca}).

\begin{figure}[htbp]
\centering
\begin{subfigure}[b]{0.32\textwidth}
  \includegraphics[width=1.0\linewidth]{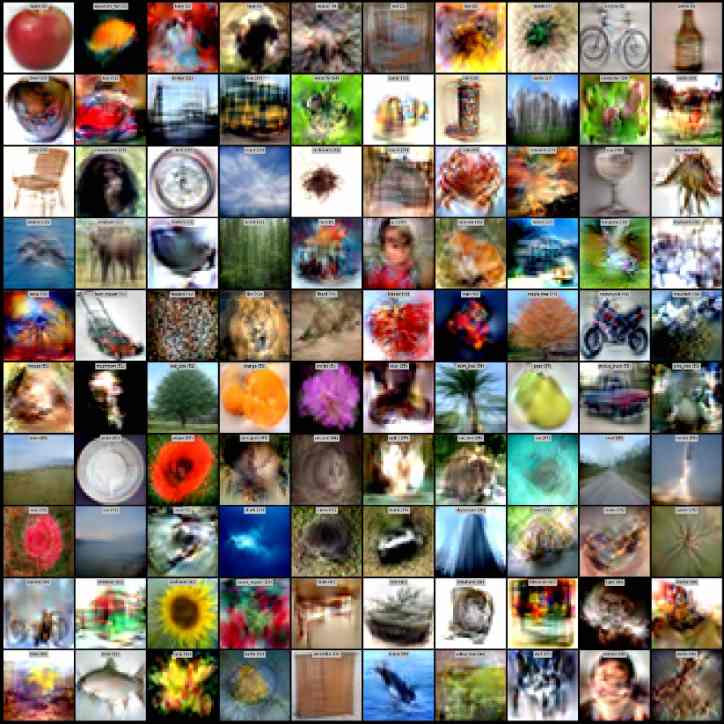}
\end{subfigure}
\vspace{-0.05in}
\caption{Ablation Study - AlexNet.}\label{fig:as_arch_alexnet}
\vspace{-0.1in}
\end{figure}

\textbf{Quantitative Results}: We evaluate the distilled data trained using different architectures (Figure \ref{fig:as_arch_conv} and \ref{fig:as_arch_dcconv}) to study the role of normalization and width in training and evaluation. Table \ref{tab:ca-normalization}, \ref{tab:as-conv-eval}, and \ref{tab:as-dcconv-eval} show that 1) Training with Conv-BN and evaluating using Conv-NN yields the best performance, which is our default choice; 2) Training with Conv-BN yields most generalizable and transferable images as it performs well for other architectures; 3) Evaluating using Conv-GN seems to be the best choice if the training architecture is unknown. 4) Evaluating using Conv-NN is a good way to see whether the inductive bias of architecture has been distilled to the dataset since it is very sensitive to the training architecture. There, it is not a good idea to use this architecture when the training architecture is unknown; 5) Training the distilled data using a wider network Conv achieves slightly better performance than a narrower network DCConv. 

We also perform the same experiments in Table \ref{tab:dd_sota} using Conv-IN, and DCConv-IN and summarize the results in Table \ref{tab:dd_nn} and \ref{tab:dd_krr}. Moreover, we also evaluate the cross-architecture transfer performance of Conv-IN in Table \ref{tab:dd_ca_app}. We observe that DCConv-IN works reasonably well when we distill a small number of images (~100). The performance degrades a lot when distilling 1000 images from CIFAR100 because the KRR component needs a larger feature dimension to perform well. Besides, Conv-IN performs slightly worse than the default Conv-BN. However, Table \ref{tab:dd_ca_app} suggests that the drawback of the instance norm is the transferability. The distilled data trained using instance normalization transfer less well to other architecture, especially those without normalization. 

\begin{table}[htbp]
\centering
\caption{Test accuracies of models trained on the distilled data from scratch. Each row has the same training architecture, while each column has the same evaluation architecture. We report the average performance of four random seeds. C denotes Conv and D denotes DCConv. The standard deviation are shown in Table \ref{tab:as-conv-eval}, and \ref{tab:as-dcconv-eval}. We visualize the distilled images in Figure \ref{fig:as_arch_conv} and \ref{fig:as_arch_dcconv}.}
\label{tab:ca-normalization}
\tiny
\centering
\begin{tabular}{c|cccccccccccc}
\toprule
T\textbackslash{}E & C-NN & C-BN & C-GN & C-GN1 & C-LN & C-IN & D-NN & D-BN & D-GN & D-GN1 & D-LN & D-IN \\
 \midrule
C-NN               & 25.7 & 22.5 & 26.5 & 26.2  & 25.3 & 23.6 & 19.1 & 21.5 & 23.8 & 24.3  & 22.0 & 22.2 \\
C-BN               & 28.8 & 26.0 & 28.1 & 27.6  & 27.4 & 27.0 & 22.7 & 24.6 & 24.3 & 25.4  & 23.4 & 24.1 \\
C-GN               & 26.1 & 24.1 & 28.4 & 27.4  & 26.0 & 25.6 & 20.0 & 23.8 & 25.9 & 26.3  & 24.1 & 24.7 \\
C-GN1              & 25.9 & 23.3 & 27.3 & 26.8  & 25.8 & 24.6 & 19.8 & 23.5 & 25.3 & 25.8  & 23.4 & 24.1 \\
C-LN               & 21.7 & 20.6 & 23.1 & 23.2  & 23.8 & 21.0 & 14.8 & 20.4 & 22.0 & 22.8  & 23.5 & 20.4 \\
C-IN               & 25.1 & 22.8 & 26.1 & 25.7  & 25.2 & 27.0 & 20.6 & 21.5 & 24.6 & 24.2  & 23.1 & 25.7 \\
D-NN               & 25.1 & 22.4 & 27.3 & 27.3  & 26.3 & 22.0 & 17.2 & 21.7 & 25.6 & 26.3  & 24.1 & 23.8 \\
D-BN               & 27.4 & 23.9 & 27.8 & 27.4  & 26.7 & 25.9 & 19.8 & 22.1 & 24.9 & 25.4  & 23.2 & 24.0 \\
D-GN               & 24.8 & 21.1 & 26.4 & 27.0  & 25.3 & 22.1 & 15.7 & 21.1 & 25.5 & 25.6  & 22.3 & 23.1 \\
D-GN1              & 24.8 & 21.5 & 26.4 & 27.2  & 25.7 & 21.5 & 16.3 & 21.4 & 24.4 & 25.7  & 22.9 & 22.0 \\
D-LN               & 24.0 & 21.7 & 24.6 & 25.1  & 25.5 & 20.7 & 13.4 & 21.3 & 21.6 & 23.1  & 25.3 & 19.8 \\
D-IN               & 23.3 & 20.2 & 24.2 & 24.0  & 23.0 & 23.4 & 16.9 & 20.1 & 22.8 & 22.5  & 21.0 & 24.8 \\
\bottomrule
\end{tabular}
\end{table}

\begin{table}[htbp]
\caption{Test accuracies of Conv with different normalizations evaluated on the distilled data trained using Conv and DCConv with different normalizations. We report the mean and standard deviation of four random seeds. The distilled images are shown in Figure \ref{fig:as_arch_conv} and \ref{fig:as_arch_dcconv}.}
\label{tab:as-conv-eval}
\tiny
\centering
\begin{tabular}{c|cccccc}
\toprule
T\textbackslash{}E & Conv-NN & Conv-BN & Conv-GN & Conv-GN1 & Conv-LN & Conv-IN \\
 \midrule
Conv-NN  & $ 25.7\pm0.3 $ & $ 22.5\pm0.4 $ & $ 26.5\pm0.2 $ & $ 26.2\pm0.1 $ & $ 25.3\pm0.2 $ & $ 23.6\pm0.2 $\\
Conv-BN  & $ 28.8\pm0.2 $ & $ 26.0\pm0.3 $ & $ 28.1\pm0.5 $ & $ 27.6\pm0.3 $ & $ 27.4\pm0.1 $ & $ 27.0\pm0.3 $\\
Conv-GN  & $ 26.1\pm0.3 $ & $ 24.1\pm0.1 $ & $ 28.4\pm0.3 $ & $ 27.4\pm0.1 $ & $ 26.0\pm0.1 $ & $ 25.6\pm0.3 $\\
Conv-GN1 & $ 25.9\pm0.1 $ & $ 23.3\pm0.1 $ & $ 27.3\pm0.3 $ & $ 26.8\pm0.3 $ & $ 25.8\pm0.1 $ & $ 24.6\pm0.2 $\\
Conv-LN  & $ 21.7\pm0.2 $ & $ 20.6\pm0.1 $ & $ 23.1\pm0.1 $ & $ 23.2\pm0.3 $ & $ 23.8\pm0.3 $ & $ 21.0\pm0.1 $\\
Conv-IN  & $ 25.1\pm0.3 $ & $ 22.8\pm0.4 $ & $ 26.1\pm0.4 $ & $ 25.7\pm0.2 $ & $ 25.2\pm0.3 $ & $ 27.0\pm0.3 $\\
 \midrule
DCConv-NN  & $ 25.1\pm0.2 $ & $ 22.4\pm0.2 $ & $ 27.3\pm0.3 $ & $ 27.3\pm0.5 $ & $ 26.3\pm0.2 $ & $ 22.0\pm0.2 $\\
DCConv-BN  & $ 27.4+0.2  $ & $ 23.9\pm0.1 $ & $ 27.8\pm0.4 $ & $ 27.4\pm0.3 $ & $ 26.7\pm0.2 $ & $ 25.9\pm0.3 $\\
DCConv-GN  & $ 24.8\pm0.2 $ & $ 21.1\pm0.1 $ & $ 26.4\pm0.3 $ & $ 27.0\pm0.3 $ & $ 25.3\pm0.2 $ & $ 22.1\pm0.3 $\\
DCConv-GN1 & $ 24.8\pm0.4 $ & $ 21.5\pm0.2 $ & $ 26.4\pm0.1 $ & $ 27.2\pm0.1 $ & $ 25.7\pm0.4 $ & $ 21.5\pm0.3 $\\
DCConv-LN  & $ 24.0\pm0.1 $ & $ 21.7\pm0.3 $ & $ 24.6\pm0.4 $ & $ 25.1\pm0.1 $ & $ 25.5\pm0.2 $ & $ 20.7\pm0.2 $\\
DCConv-IN  & $ 23.3\pm0.2 $ & $ 20.2\pm0.3 $ & $ 24.2\pm0.3 $ & $ 24.0\pm0.2 $ & $ 23.0\pm0.4 $ & $ 23.4\pm0.3 $\\
\bottomrule
\end{tabular}
\end{table}

\begin{table}[htbp]
\caption{Test accuracies of DCConv with different normalizations evaluated on the distilled data trained using Conv and DCConv with different normalizations. We report the mean and standard deviation of four random seeds. The distilled images are shown in Figure \ref{fig:as_arch_conv} and \ref{fig:as_arch_dcconv}.}
\label{tab:as-dcconv-eval}
\tiny
\centering
\begin{tabular}{c|cccccc}
\toprule
T\textbackslash{}E & DCConv-NN & DCConv-BN & DCConv-GN & DCConv-GN1 & DCConv-LN & DCConv-IN \\
 \midrule
Conv-NN  & $ 19.1\pm0.1 $ & $ 21.5\pm0.2 $ & $ 23.8\pm0.1 $ & $ 24.3\pm0.3 $ & $ 22.0\pm0.2 $ & $ 22.2\pm0.1 $\\
Conv-BN  & $ 22.7\pm0.1 $ & $ 24.6\pm0.1 $ & $ 24.3\pm0.2 $ & $ 25.4\pm0.2 $ & $ 23.4\pm0.1 $ & $ 24.1\pm0.2 $\\
Conv-GN  & $ 20.0\pm0.7 $ & $ 23.8\pm0.3 $ & $ 25.9\pm0.1 $ & $ 26.3\pm0.1 $ & $ 24.1\pm0.2 $ & $ 24.7\pm0.2 $\\
Conv-GN1 & $ 19.8\pm0.7 $ & $ 23.5\pm0.4 $ & $ 25.3\pm0.1 $ & $ 25.8\pm0.2 $ & $ 23.4\pm0.1 $ & $ 24.1\pm0.1 $\\
Conv-LN  & $ 14.8\pm0.8 $ & $ 20.4\pm0.3 $ & $ 22.0\pm0.3 $ & $ 22.8\pm0.3 $ & $ 23.5\pm0.4 $ & $ 20.4\pm0.2 $\\
Conv-IN  & $ 20.6\pm0.4 $ & $ 21.5\pm0.2 $ & $ 24.6\pm0.1 $ & $ 24.2\pm0.2 $ & $ 23.1\pm0.2 $ & $ 25.7\pm0.2 $\\
\midrule
DCConv-NN  & $ 17.2\pm0.9 $ & $ 21.7\pm0.1 $ & $ 25.6\pm0.2 $ & $ 26.3\pm0.3 $ & $ 24.1\pm0.1 $ & $ 23.8\pm0.2 $\\
DCConv-BN  & $ 19.8\pm0.5 $ & $ 22.1\pm0.3 $ & $ 24.9\pm0.3 $ & $ 25.4\pm0.2 $ & $ 23.2\pm0.2 $ & $ 24.0\pm0.3 $\\
DCConv-GN  & $ 15.7\pm0.7 $ & $ 21.1\pm0.2 $ & $ 25.5\pm0.3 $ & $ 25.6\pm0.3 $ & $ 22.3\pm0.2 $ & $ 23.1\pm0.1 $\\
DCConv-GN1 & $ 16.3\pm0.8 $ & $ 21.4\pm0.2 $ & $ 24.4\pm0.2 $ & $ 25.7\pm0.4 $ & $ 22.9\pm0.3 $ & $ 22.0\pm0.2 $\\
DCConv-LN  & $ 13.4\pm1.1 $ & $ 21.3\pm0.4 $ & $ 21.6\pm0.5 $ & $ 23.1\pm0.3 $ & $ 25.3\pm0.3 $ & $ 19.8\pm0.3 $\\
DCConv-IN  & $ 16.9\pm0.8 $ & $ 20.1\pm0.3 $ & $ 22.8\pm0.1 $ & $ 22.5\pm0.2 $ & $ 21.0\pm0.4 $ & $ 24.8\pm0.3 $\\
\bottomrule
\end{tabular}
\end{table}

\begin{table}[htbp]
  \caption{Cross-architecture transfer performance on CIFAR10 with 10 Img/Cls. Despite being trained for a specific architecture, our distilled data transfer well to various architectures unseen during training. Conv is the default evaluation model used for each method. NN, DN, IN, and BN stand for no normalization, default normalization, Instance Normalization, Batch Normalization, respectively.}
  \label{tab:dd_ca_app}
  \tiny
  \centering
    \begin{tabular}{ccccccccc}
        \toprule
        \multirow{2}{*}[-4pt]{} & \multirow{2}{*}[-4pt]{Train Arch} & \multicolumn{7}{c}{Evaluation Architecture}\\
        \cmidrule(l){3-9}
        && Conv & Conv-NN & ResNet-DN & ResNet-BN & VGG-DN & VGG-BN & AlexNet \\
        \midrule
        DSA \citep{DBLP:conf/icml/DSA} &  Conv-IN & $53.2 \pm 0.8$ & $36.4 \pm 1.5$ & $42.1 \pm 0.7 $ & $ 34.1 \pm 1.4$ & $ 48.3\pm0.7 $ & $ 46.3 \pm 1.3$ & $ 34.0 \pm 2.3$ \\
        DM \citep{DBLP:journals/corr/DM} &  Conv-IN & $49.2 \pm 0.8$ & $35.2 \pm 0.5$ & $36.8 \pm 1.2 $ & $ 35.5 \pm 1.3$ & $ 45.5\pm1.0 $ & $ 41.2 \pm 1.8$ & $ 34.9 \pm 1.1$ \\
        MTT \citep{DBLP:journals/corr/MTT} &  Conv-IN & $64.4 \pm 0.9$ & $41.6 \pm 1.3$ & $49.2 \pm 1.1 $ & $ 42.9 \pm 1.5$ & $ 35.7\pm 3.35$ & $ 46.6 \pm 2.0$ & $ 34.2 \pm 2.6$ \\
        KIP \citep{DBLP:conf/nips/KIP2} &  Conv-NTK & $62.7 \pm 0.3$ & $58.2 \pm 0.4$ & $49.0 \pm 1.2 $ & $ 45.8 \pm 1.4$ & $ 32.0\pm0.4 $ & $ 30.1 \pm 1.5$ & $ 57.2 \pm 0.4$ \\
        \midrule
        \algname & Conv-IN & $ 59.2 \pm 0.3$ & $ 56.2 \pm 0.2$ & $ 51.1 \pm 0.8$ & $ 50.8 \pm 0.2$ & $ \textbf{57.5} \pm \textbf{0.7}$ & $51.8 \pm 0.3$ & $55.3 \pm 0.8 $\\
        \algname & Conv-BN & $ \textbf{65.5} \pm \textbf{0.4}$ & $ \textbf{65.5} \pm \textbf{0.4}$ & $ \textbf{58.1} \pm \textbf{0.6}$ & $ \textbf{57.7} \pm \textbf{0.7}$ & $ 49.1 \pm 0.5 $ & $\textbf{59.4} \pm \textbf{0.7}$ & $\textbf{61.9} \pm \textbf{0.7} $\\
        \bottomrule
    \end{tabular}
\end{table}

\begin{table}[htbp]
  \caption{NN test accuracies of models trained on the distilled data from scratch using Conv, Conv-IN, and DCConv-IN. $^\dag$ denotes performance better than the original reported performance.}
  \label{tab:dd_nn}
  \tiny
  \centering
    \begin{tabular}{ccccccccc}
        \toprule
        \multirow{2}{*}[-4pt]{} & \multirow{2}{*}[-4pt]{Img/Cls} & \multicolumn{4}{c}{Previous SOTA} & \multicolumn{3}{c}{\algname}\\
        \cmidrule(l){3-6} \cmidrule(l){7-9}
        & & DSA \cite{DBLP:conf/icml/DSA} & DM \cite{DBLP:journals/corr/DM}& KIP \cite{DBLP:conf/nips/KIP2} & MTT \cite{DBLP:journals/corr/MTT} & Conv-BN & Conv-IN & DCConv-IN \\
        \midrule
        \multirow{3}{*}{MNIST} & 1 & $88.7 \pm 0.6$ & $89.9 \pm 0.8$\rlap{$^\dag$} & $90.1 \pm 0.1$ & $91.4 \pm 0.9$\rlap{$^\dag$} & $\textbf{93.0} \pm \textbf{0.4}$ & $92.9 \pm 0.5$ & $92.4 \pm 0.5$\\
         & 10 & $97.9 \pm 0.1$\rlap{$^\dag$}  & $97.6 \pm 0.1$\rlap{$^\dag$} & $97.5\pm 0.0$ & $97.3 \pm 0.1$\rlap{$^\dag$} & $\textbf{98.6} \pm \textbf{0.1} $  & $98.9 \pm 0.1$ & $98.4 \pm 0.1$\\
         & 50 & $99.2 \pm 0.1$ & $98.6 \pm 0.1$ & $98.3\pm 0.1$ & $98.5\pm 0.1$\rlap{$^\dag$} & $99.2 \pm 0.0$ & $\textbf{99.4} \pm \textbf{0.1}$ & $98.8 \pm 0.1$\\
        \midrule
        \multirow{3}{*}{F-MNIST} & 1 & $ 70.6 \pm 0.6$ & $ 71.5 \pm 0.5$\rlap{$^\dag$} & $ 73.5 \pm 0.5$ & $75.1 \pm 0.9$\rlap{$^\dag$} & $75.6 \pm 0.3$ & $75.3 \pm 0.8$ & $\textbf{76.4} \pm \textbf{1.2}$\\
         & 10 & $ 84.8\pm 0.3$\rlap{$^\dag$} & $ 83.6 \pm 0.2$\rlap{$^\dag$} & $ 86.8 \pm 0.1$ & $ \textbf{87.2} \pm \textbf{0.3}$\rlap{$^\dag$}  & $86.2 \pm 0.2 $ & $86.0 \pm 0.3$ & $85.7 \pm 0.2$\\
         & 50 & $ 88.8\pm 0.2$\rlap{$^\dag$} & $ 88.2 \pm 0.1$\rlap{$^\dag$} & $ 88.0 \pm 0.1$ & $88.3\pm 0.1$\rlap{$^\dag$} & $ \textbf{89.6}\pm \textbf{0.1}$ & $89.4 \pm 0.1$ & $87.2 \pm 0.2$\\
        \midrule
        \multirow{3}{*}{CIFAR10} & 1 & $36.7 \pm 0.8$\rlap{$^\dag$} & $31.0 \pm 0.6$\rlap{$^\dag$} & $ \textbf{49.9}\pm \textbf{0.2} $ & $ 46.3\pm 0.8 $ & $ 46.8 \pm 0.7 $ & $45.1 \pm 0.5$ & $41.3 \pm 0.5$\\
         & 10 & $53.2 \pm 0.8$\rlap{$^\dag$} & $49.2 \pm 0.8$\rlap{$^\dag$} & $ 62.7\pm 0.3 $ & $ 65.3 \pm 0.7 $ & $ \textbf{65.5}\pm \textbf{0.4}$ & $59.1 \pm 0.3$ & $59.6 \pm 0.3$\\
         & 50 & $66.8 \pm 0.4$\rlap{$^\dag$} & $63.7 \pm 0.5$\rlap{$^\dag$} & $ 68.6 \pm 0.2 $ & $ 71.6\pm 0.2 $ & $\textbf{71.7}\pm \textbf{0.2}$ & $69.6 \pm 0.4$ & $63.6 \pm 0.2$\\
        \midrule
        \multirow{3}{*}{CIFAR100} & 1 & $16.8 \pm 0.2$\rlap{$^\dag$} & $12.2 \pm 0.4$\rlap{$^\dag$} & $ 15.7\pm 0.2 $ & $ 24.3 \pm 0.3 $ & $ \textbf{28.7}\pm \textbf{0.1}$ & $25.9 \pm 0.1$ & $24.8 \pm 0.2$\\
         & 10 & $ 32.3 \pm 0.3$ & $ 29.7 \pm 0.3$ & $ 28.3\pm 0.1 $ & $ 40.1 \pm 0.4 $ & $ \textbf{42.5}\pm \textbf{0.2}$ & $40.9 \pm 0.1$ & $31.2 \pm 0.1$\\
         & 50 & $ 42.8 \pm 0.4$ & $ 43.6 \pm 0.4$ & $-$ & $\textbf{47.7} \pm \textbf{0.2}$ & $ 44.3\pm 0.2$ & $-$ & $-$ \\
        \midrule
        \multirow{2}{*}{T-ImageNet} & 1 & $ 6.6 \pm 0.2$\rlap{$^\dag$} & $3.9 \pm 0.2$ & $ - $ & $ 8.8\pm 0.3 $ & $ \textbf{15.4}\pm \textbf{0.1}$ & $13.5 \pm 0.1$ & $-$\\
         & 10 & $-$ & $ 12.9 \pm 0.4$ &$-$ & $ 23.2\pm 0.2 $ & $\textbf{25.4} \pm \textbf{0.2}$ & $20.4 \pm 0.1$ & $-$\\
        \bottomrule
    \end{tabular}
\end{table}

\begin{table}[htbp]
  \caption{KRR test accuracies of FRePo trained on the distilled data from scratch using Conv, Conv-IN, and DCConv-IN. $^\dag$ denotes performance better than the original reported performance.}
  \label{tab:dd_krr}
  \tiny
  \centering
    \begin{tabular}{ccccccccc}
        \toprule
        \multirow{2}{*}[-4pt]{} & \multirow{2}{*}[-4pt]{Img/Cls} & \multicolumn{4}{c}{Previous SOTA} & \multicolumn{3}{c}{\algname}\\
        \cmidrule(l){3-6} \cmidrule(l){7-9}
        & & DSA \cite{DBLP:conf/icml/DSA} & DM \cite{DBLP:journals/corr/DM}& KIP \cite{DBLP:conf/nips/KIP2} & MTT \cite{DBLP:journals/corr/MTT} & Conv-BN & Conv-IN & DCConv-IN \\
        \midrule
        \multirow{3}{*}{MNIST} & 1 & $88.7 \pm 0.6$ & $89.9 \pm 0.8$\rlap{$^\dag$} & $90.1 \pm 0.1$ & $91.4 \pm 0.9$\rlap{$^\dag$} & $ 92.6 \pm 0.4$ & $\textbf{92.7} \pm \textbf{0.3}$ & $91.1 \pm 0.5$\\
         & 10 & $97.9 \pm 0.1$\rlap{$^\dag$}  & $97.6 \pm 0.1$\rlap{$^\dag$} & $97.5\pm 0.0$ & $97.3 \pm 0.1$\rlap{$^\dag$} & $ 98.6 \pm 0.1 $ & $\textbf{98.8} \pm \textbf{0.1}$ & $98.4 \pm 0.1$\\
         & 50 & $\textbf{99.2} \pm \textbf{0.1}$ & $98.6 \pm 0.1$ & $98.3\pm 0.1$ & $98.5\pm 0.1$\rlap{$^\dag$} & $99.2 \pm 0.1$ & $\textbf{99.3} \pm \textbf{0.1}$ & $98.9 \pm 0.1$\\
        \midrule
        \multirow{3}{*}{F-MNIST} & 1 & $ 70.6 \pm 0.6$ & $ 71.5 \pm 0.5$\rlap{$^\dag$} & $ 73.5 \pm 0.5$ & $75.1 \pm 0.9$\rlap{$^\dag$} & $77.1 \pm 0.2$ & $71.7 \pm 1.2$ & $\textbf{78.5} \pm \textbf{0.2}$\\
         & 10 & $ 84.8\pm 0.3$\rlap{$^\dag$} & $ 83.6 \pm 0.2$\rlap{$^\dag$} & $ 86.8 \pm 0.1$ & $ \textbf{87.2} \pm \textbf{0.3}$\rlap{$^\dag$} & $86.8 \pm 0.1 $ & $86.9 \pm 0.2$ & $86.2 \pm 0.1$\\
         & 50 & $ 88.8\pm 0.2$\rlap{$^\dag$} & $ 88.2 \pm 0.1$\rlap{$^\dag$} & $ 88.0 \pm 0.1$ & $88.3\pm 0.1$\rlap{$^\dag$} & $\textbf{89.9}\pm \textbf{0.1}$ & $\textbf{89.9}\pm \textbf{0.1}$ & $87.4 \pm 0.2$\\
        \midrule
        \multirow{3}{*}{CIFAR10} & 1 & $36.7 \pm 0.8$\rlap{$^\dag$} & $31.0 \pm 0.6$\rlap{$^\dag$} & $ \textbf{49.9}\pm \textbf{0.2} $ & $ 46.3\pm 0.8 $ & $ 47.9 \pm 0.6$ & $46.8 \pm 0.3$ & $43.3 \pm 0.5$\\
         & 10 & $53.2 \pm 0.8$\rlap{$^\dag$} & $49.2 \pm 0.8$\rlap{$^\dag$} & $ 62.7\pm 0.3 $ & $ 65.3 \pm 0.7 $ & $ \textbf{68.0}\pm \textbf{0.2}$ & $61.9 \pm 0.4$ & $61.8 \pm 0.3$\\
         & 50 & $66.8 \pm 0.4$\rlap{$^\dag$} & $63.7 \pm 0.5$\rlap{$^\dag$} & $ 68.6 \pm 0.2 $ & $ 71.6\pm 0.2 $ & $\textbf{74.4}\pm \textbf{0.1}$ & $71.4 \pm 0.3$ & $64.3 \pm 0.1$\\
        \midrule
        \multirow{3}{*}{CIFAR100} & 1 & $16.8 \pm 0.2$\rlap{$^\dag$} & $12.2 \pm 0.4$\rlap{$^\dag$} & $ 15.7\pm 0.2 $ & $ 24.3 \pm 0.3 $ & $ \textbf{32.3}\pm \textbf{0.1}$ & $25.7 \pm 0.2$ & $26.9 \pm 0.1$\\
         & 10 & $ 32.3 \pm 0.3$ & $ 29.7 \pm 0.3$ & $ 28.3\pm 0.1 $ & $ 40.1 \pm 0.4 $ & $ \textbf{44.9}\pm \textbf{0.2}$ & $42.0 \pm 0.3$ & $30.9 \pm 0.1$\\
         & 50 & $ 42.8 \pm 0.4$ & $ 43.6 \pm 0.4$ & $-$ & $\textbf{47.7} \pm \textbf{0.2}$ & $43.0 \pm 0.3$ & $-$ & $-$\\
        \midrule
        \multirow{2}{*}{T-ImageNet} & 1 & $ 6.6 \pm 0.2$\rlap{$^\dag$} & $3.9 \pm 0.2$ & $ - $ & $ 8.8\pm 0.3 $ & $ \textbf{19.1}\pm \textbf{0.3}$ & $15.8 \pm 0.3$ & $-$\\
         & 10 & $-$ & $ 12.9 \pm 0.4$ &$-$ & $ 23.2\pm 0.2 $ & $\textbf{26.5} \pm \textbf{0.1}$ & $20.8 \pm 0.1$ & $-$\\
        \bottomrule
    \end{tabular}
\end{table}
\clearpage
\newpage
\section{Hyperparameter Tuning Guideline} \label{app:tuningguide} 
We find that several modifications to the current method can improve the test accuracy of the model trained on the distilled data. We do not include them in our current algorithm for simplicity or fair comparison, and we guess it may be helpful for practitioners.
\begin{itemize}
    \item \textbf{Dropout:} We find dropout is very effective at alleviating the overfitting when training on a small set of examples. 
    \item \textbf{Learning Rate Schedule for the online model:} Though we use a constant schedule in our experiments, we find that the learning rate schedule, especially the warm-up phase, may be crucial for some architectures. When NAN is in gradient, adding a learning rate schedule to the online model may solve the problem.
    \item \textbf{Data Augmentation during Training:} There are two ways to add data augmentation during training. One is to add to $\supportx$ during the online model update. The other is to add to $\targetx$, which can be thought of as distilling data augmentation to the data. 
    \item \textbf{Tune Max Online Update:} We train the online model up to 100 steps in our experiments. When the distillation size is tiny (e.g., ten examples in total), 100 may be too large. Setting to a lower value (less regularization) turns out to be better. On the contrary, if the distillation size is large, setting it to a higher value (more regularization) can give better results.
    \item \textbf{Exponential Moving Average (EMA):} We find that evaluating on the EMA version of the distilled data or using the EMA version of the model parameters can improve the test accuracy.
    \item \textbf{Soft Cross Entropy Loss:} We train models on the distilled dataset using MSE loss in all our experiments to take advantage of the distilled label. An alternative way is to use the soft cross-entropy loss with a fine-tuned temperature, which usually outperforms the MSE loss.
    \item \textbf{Image Regularization:} Though our method does not have any image regularization, we find a regularization term on the image norm is necessary for some architectures. Otherwise, the image norm will keep increasing or decreasing. We find that using an L2 penalty between the distilled image norm and the real image norm is enough in some cases. An alternative way is to project the distilled image to a norm ball every few iterations.
    \item \textbf{Label Regularization:} Our method has a small regularization to ensure the class-balanced distillation where we force the margin between the target label and any other label is greater than 1/C. We believe a better label regularization incorporating prior knowledge of class similarity can improve the performance. 
    \item \textbf{Maximal Update Parametrization ($\mu P$) \citep{DBLP:journals/corr/abs-2203-03466}:} We observe that using a model parameterization proposed by \citet{DBLP:journals/corr/abs-2203-03466} can give additional test accuracy improvement.
\end{itemize}

Based on our experience, we provide the following hyperparameter tuning guideline for those who want to squeeze the performance of our method or apply our method to a different dataset or use a different model. Besides validation loss and accuracy, we suggest monitoring the norm and gradient norm of the distilled images and labels, which can be good indicators for the final performance.

\begin{itemize}
    \item \textbf{Step1 - Online Model:} Choose an online model architecture and tune its hyperparameter (e.g., optimizer, weight decay) in the standard way on the whole real dataset or a subset of real data. The same hyperparameters can be used for the online model update and final evaluation.
    \item \textbf{Step2 - Distilled Data Optimization and Initialization:} Use the default setting for the model pool (i.e., ten models with max online update $K=100$) and tune the learning rate and batch size for distilled data optimization and the scale of initialization.
    \item \textbf{Step3 - Pool Diversity:} Tune the model pool diversity by adjusting the max online update, adding models with different architectures, or applying data augmentation.
\end{itemize}
\clearpage
\newpage
\section{Additional Visualization}\label{app:addvis}

\subsection{Distilled Image Visualization}
\begin{figure}[htbp]
\centering
\begin{subfigure}[b]{0.47\textwidth}
  \includegraphics[width=1.0\linewidth]{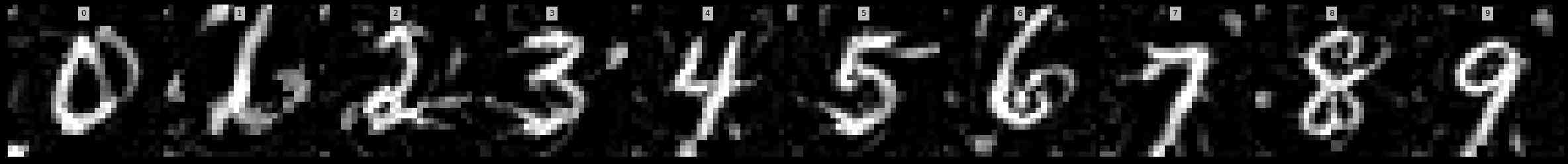}
  \caption{1 Img/Cls, Learn Label=True} 
\end{subfigure}
\hspace{0.2in}
\begin{subfigure}[b]{0.47\textwidth}
  \includegraphics[width=1.0\linewidth]{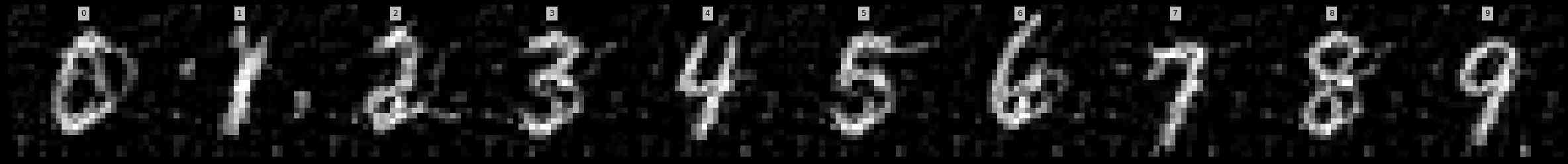}  
  \caption{1 Img/Cls, Learn Label=False} 
\end{subfigure}

\begin{subfigure}[b]{0.47\textwidth}
  \includegraphics[width=1.0\linewidth]{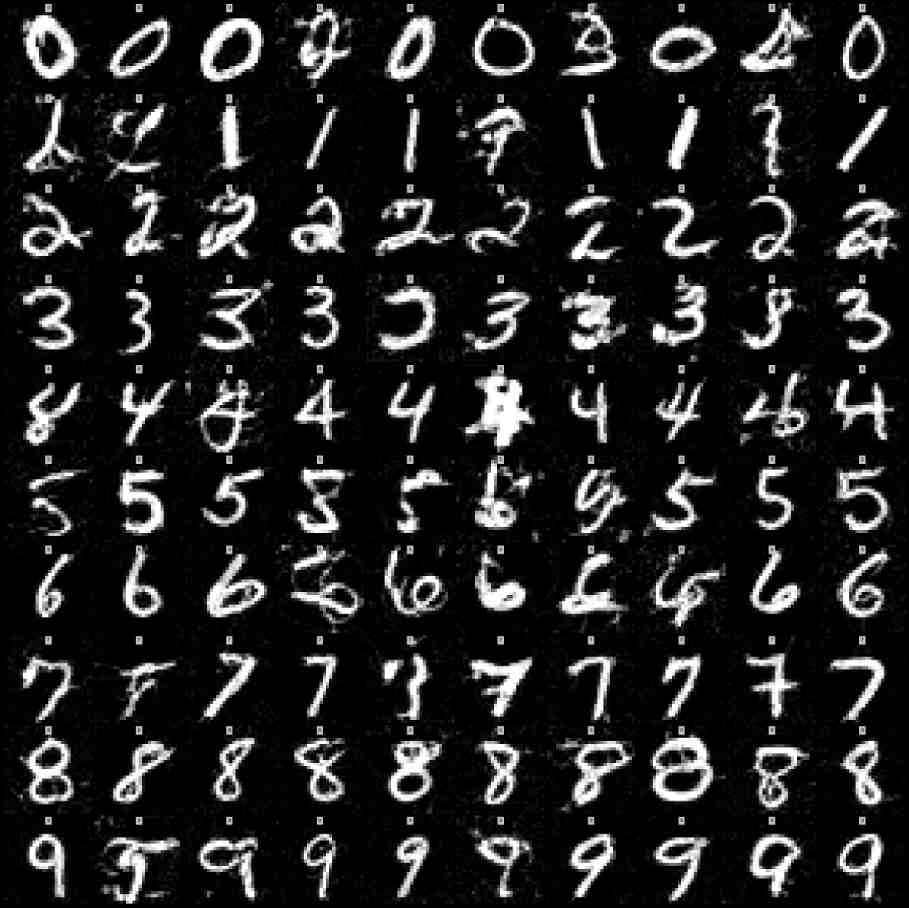}
  \caption{10 Img/Cls, Learn Label=True} 
\end{subfigure}
\hspace{0.2in}
\begin{subfigure}[b]{0.47\textwidth}
  \includegraphics[width=1.0\linewidth]{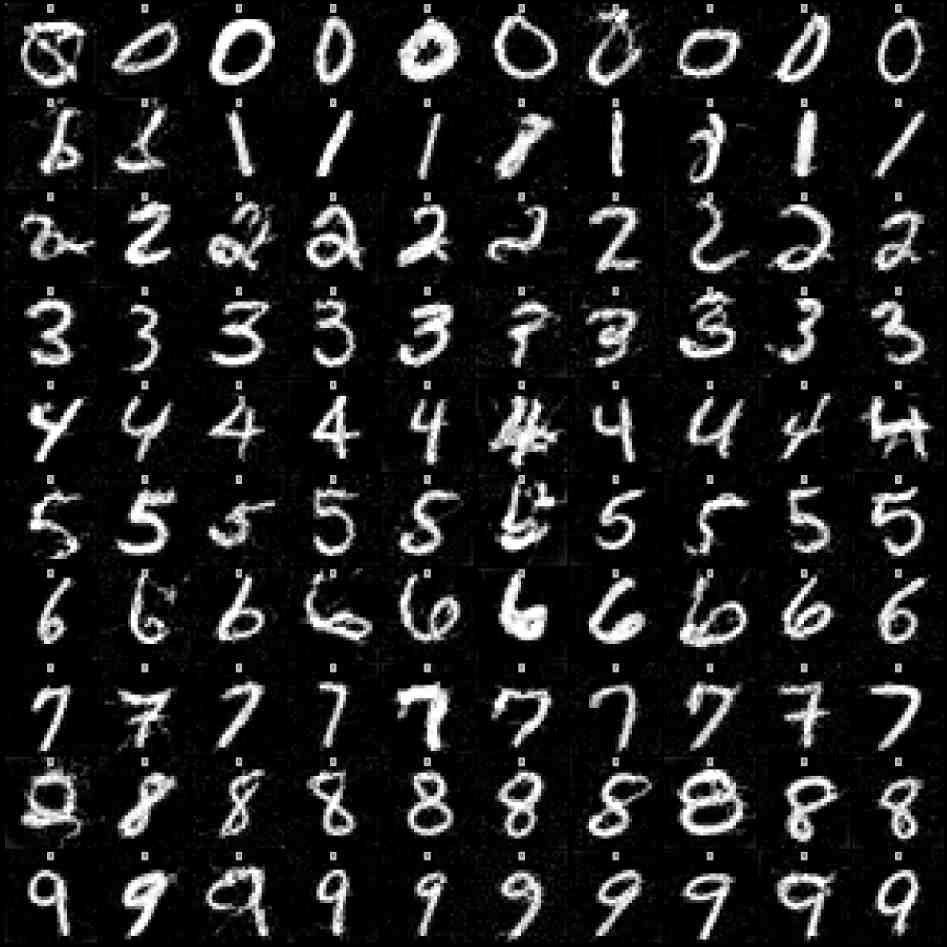}  
  \caption{10 Img/Cls, Learn Label=False} 
\end{subfigure}
\caption{Distilled Image Visualization - MNIST.}\label{fig:vis_mnist}
\end{figure}

\begin{figure}[htbp]
\centering
\begin{subfigure}[b]{0.47\textwidth}
  \includegraphics[width=1.0\linewidth]{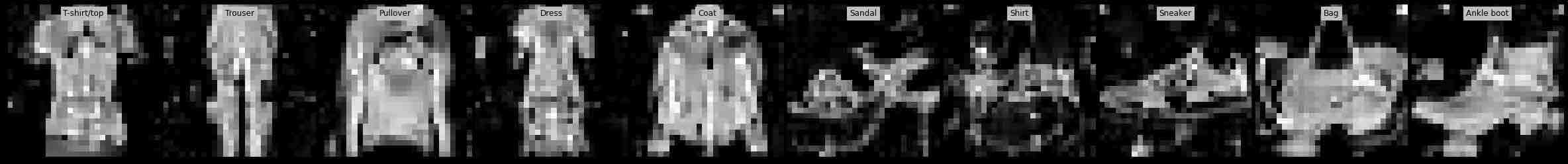}
  \caption{1 Img/Cls, Learn Label=True} 
\end{subfigure}
\hspace{0.2in}
\begin{subfigure}[b]{0.47\textwidth}
  \includegraphics[width=1.0\linewidth]{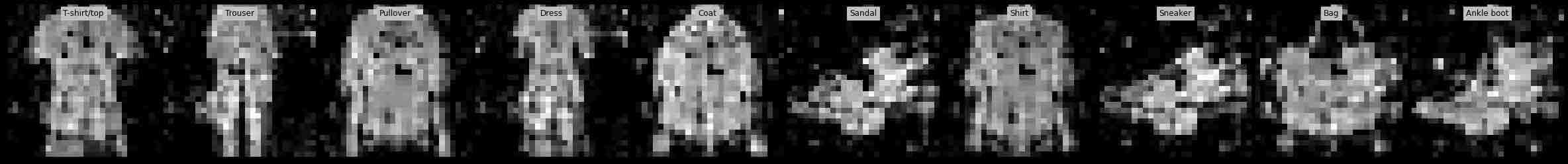}  
  \caption{1 Img/Cls, Learn Label=False} 
\end{subfigure}

\begin{subfigure}[b]{0.47\textwidth}
  \includegraphics[width=1.0\linewidth]{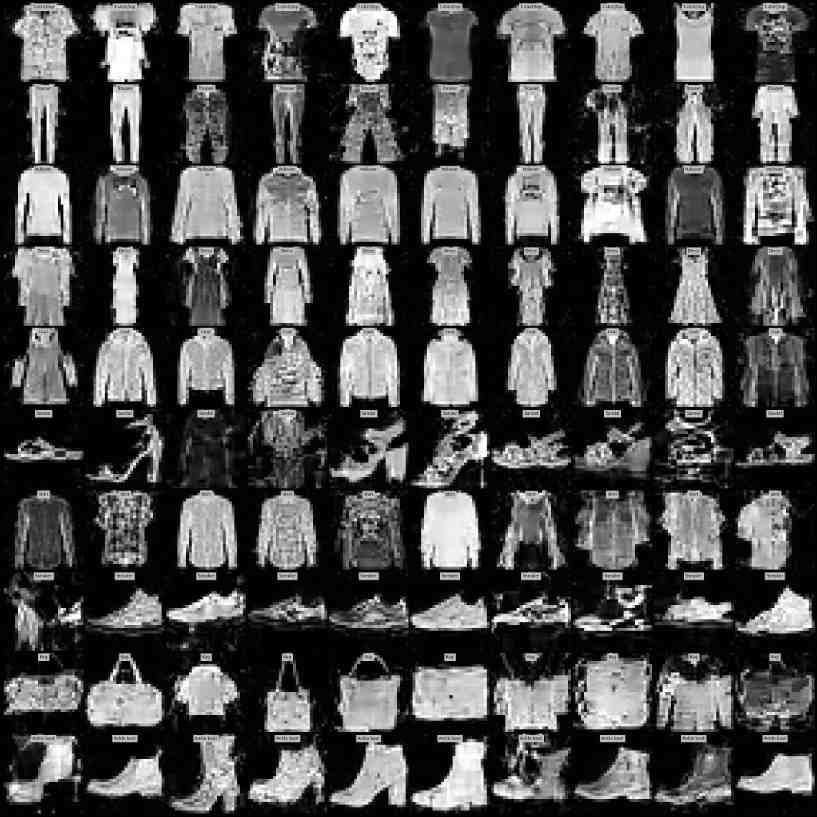}
  \caption{10 Img/Cls, Learn Label=True} 
\end{subfigure}
\hspace{0.2in}
\begin{subfigure}[b]{0.47\textwidth}
  \includegraphics[width=1.0\linewidth]{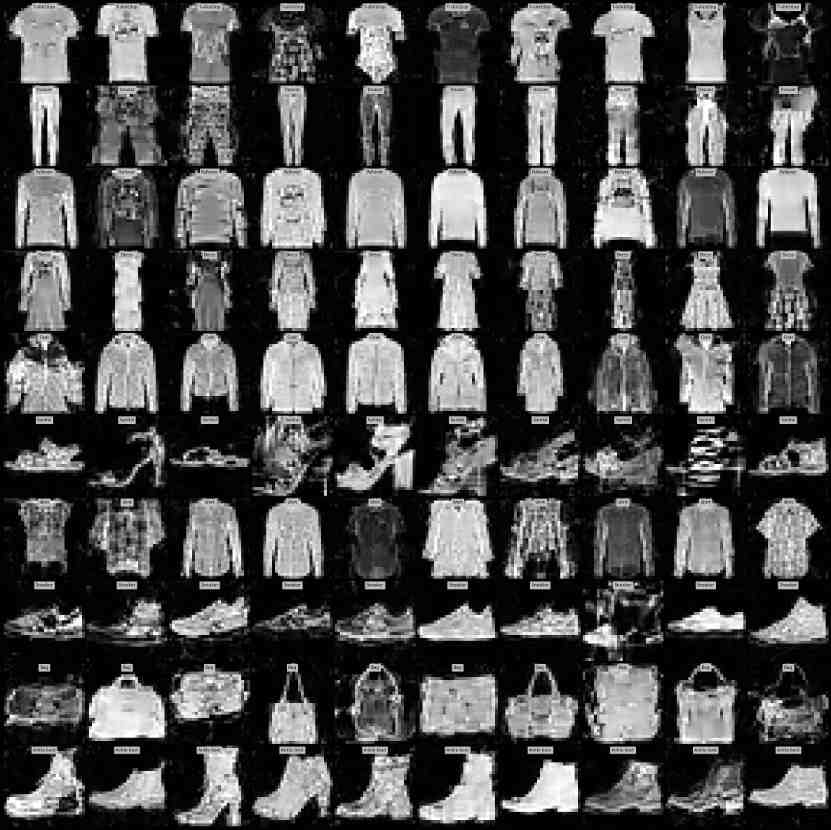}  
  \caption{10 Img/Cls, Learn Label=False} 
\end{subfigure}
\caption{Distilled Image Visualization - Fashion MNIST.}\label{fig:vis_fmnist}
\end{figure}

\begin{figure}[htbp]
\centering
\begin{subfigure}[b]{0.47\textwidth}
  \includegraphics[width=1.0\linewidth]{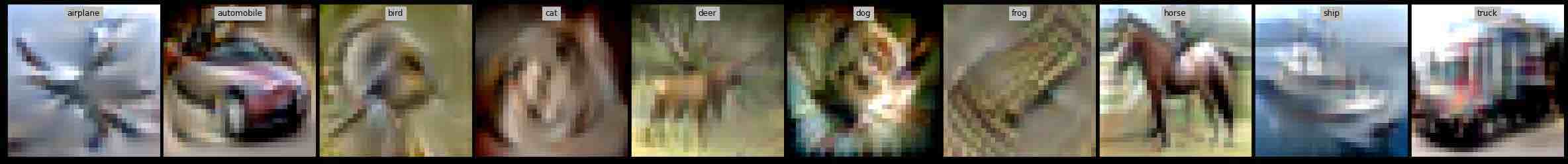}
  \caption{1 Img/Cls, Learn Label=True} 
\end{subfigure}
\hspace{0.2in}
\begin{subfigure}[b]{0.47\textwidth}
  \includegraphics[width=1.0\linewidth]{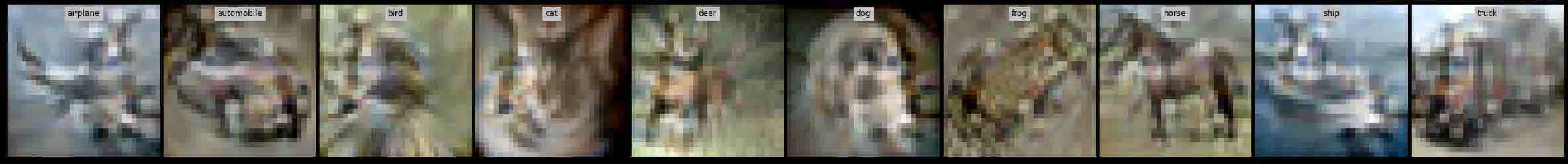}  
  \caption{1 Img/Cls, Learn Label=False} 
\end{subfigure}

\begin{subfigure}[b]{0.47\textwidth}
  \includegraphics[width=1.0\linewidth]{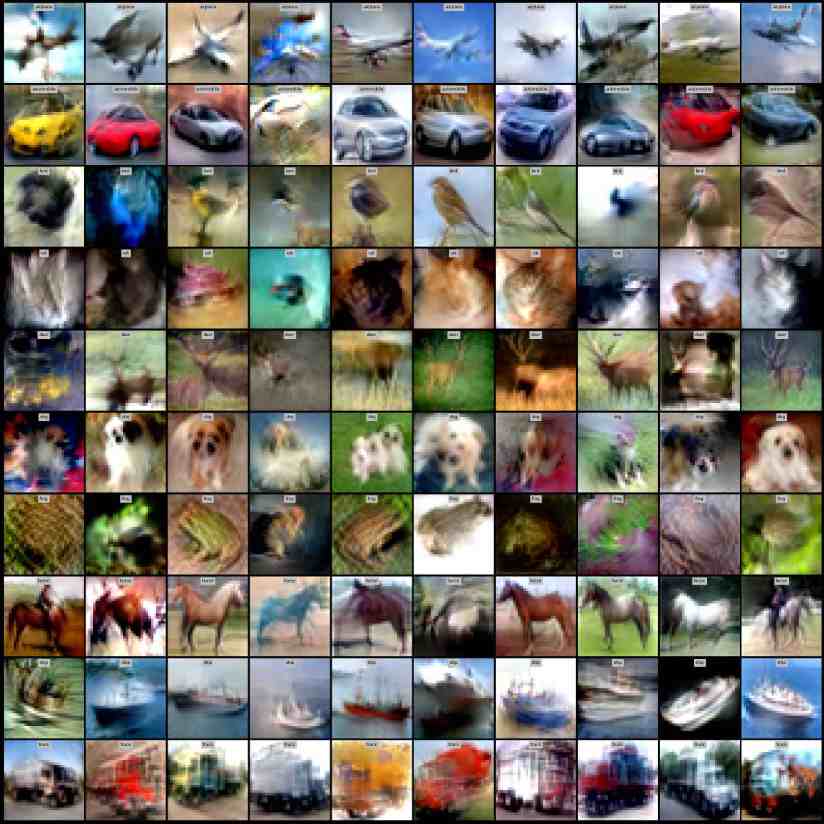}
  \caption{10 Img/Cls, Learn Label=True} 
\end{subfigure}
\hspace{0.2in}
\begin{subfigure}[b]{0.47\textwidth}
  \includegraphics[width=1.0\linewidth]{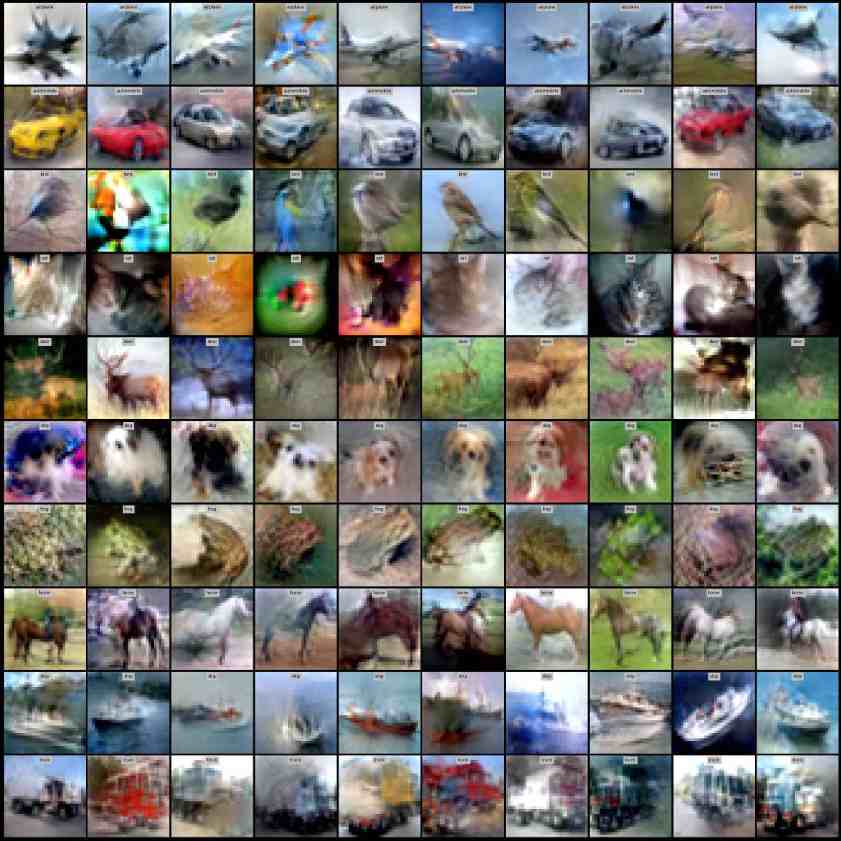}  
  \caption{10 Img/Cls, Learn Label=False} 
\end{subfigure}
\caption{Distilled Image Visualization - CIFAR10}\label{fig:vis_cifar10}
\end{figure}

\begin{figure}[htbp]
\centering
\begin{subfigure}[b]{0.47\textwidth}
  \includegraphics[width=1.0\linewidth]{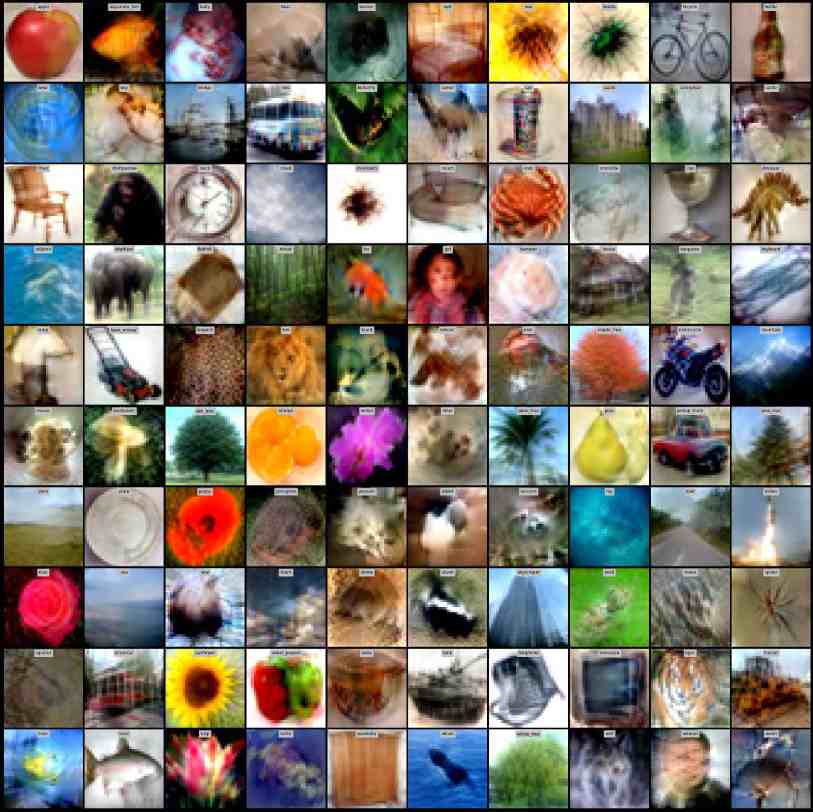}
  \caption{1 Img/Cls, Learn Label=True} 
\end{subfigure}
\hspace{0.2in}
\begin{subfigure}[b]{0.47\textwidth}
  \includegraphics[width=1.0\linewidth]{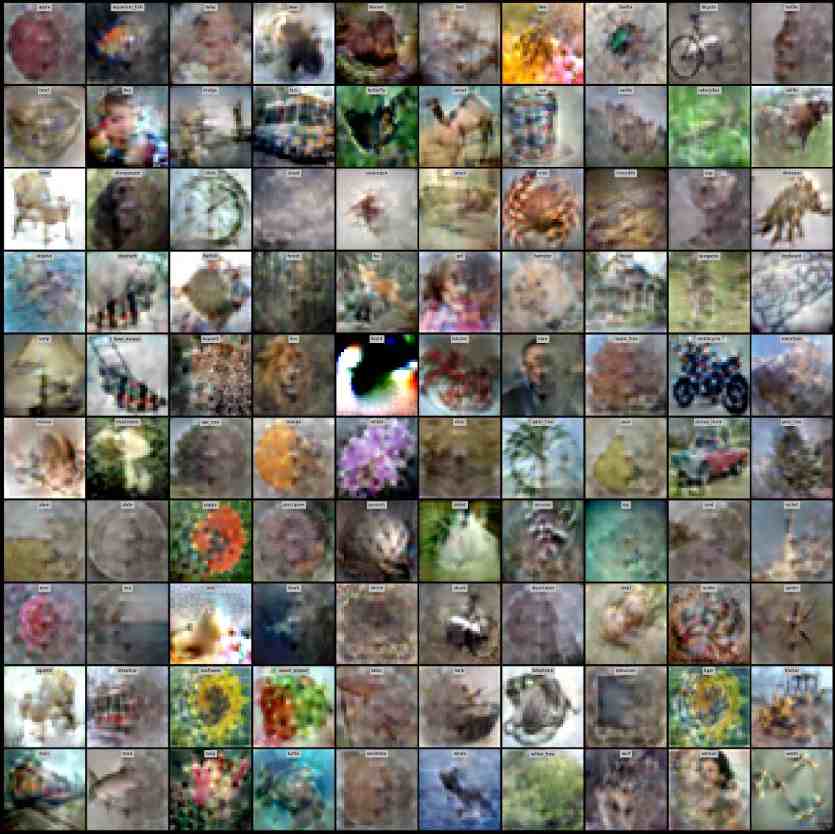}  
  \caption{1 Img/Cls, Learn Label=False} 
\end{subfigure}
\caption{Distilled Image Visualization - CIFAR100}\label{fig:vis_cifar100}
\end{figure}

\begin{figure}[htbp]
\centering
\begin{subfigure}[b]{0.47\textwidth}
  \includegraphics[width=1.0\linewidth]{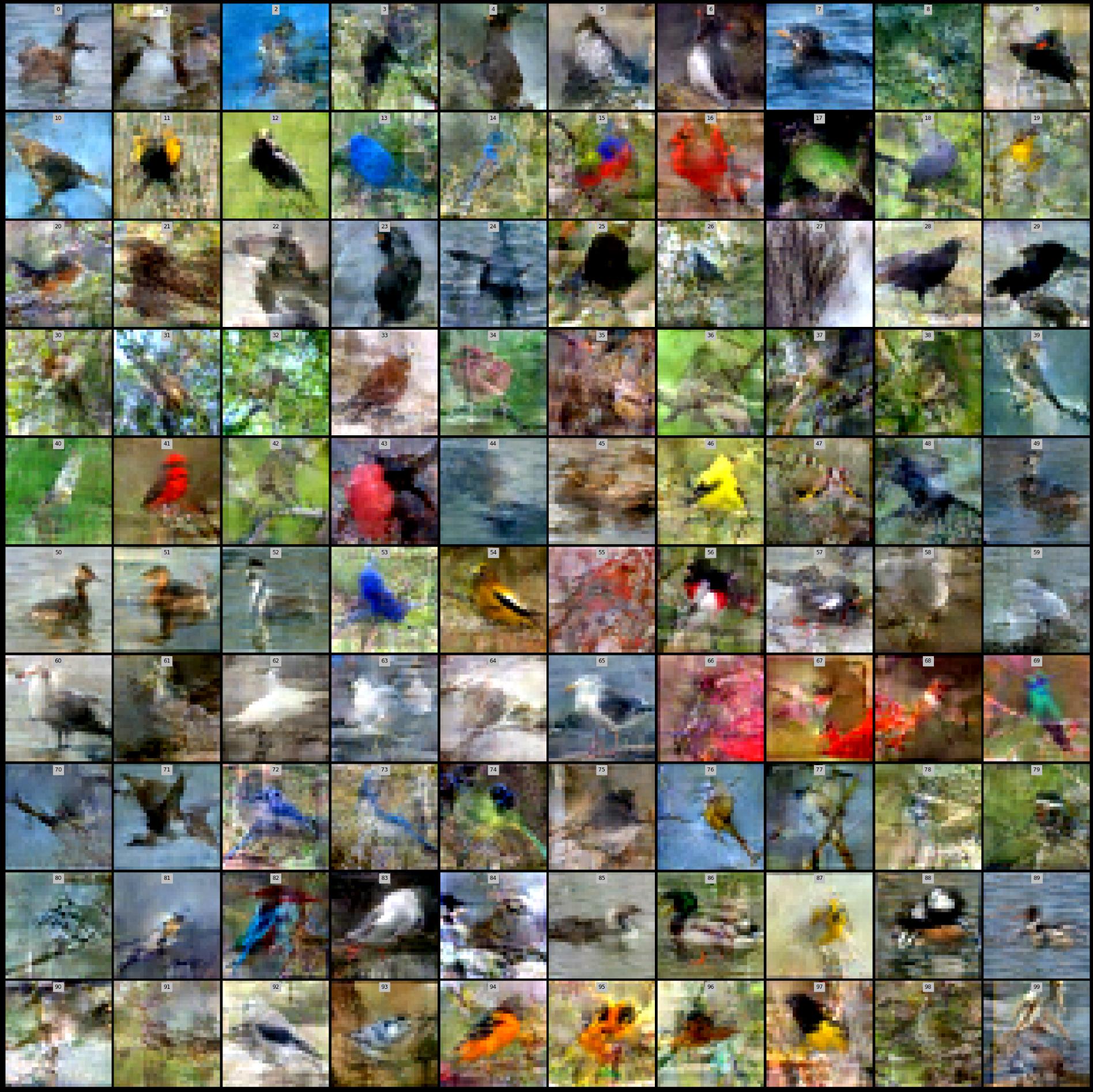}
  \caption{Learn Label=True, Class ID = [0-99]} 
\end{subfigure}
\hspace{0.2in}
\begin{subfigure}[b]{0.47\textwidth}
  \includegraphics[width=1.0\linewidth]{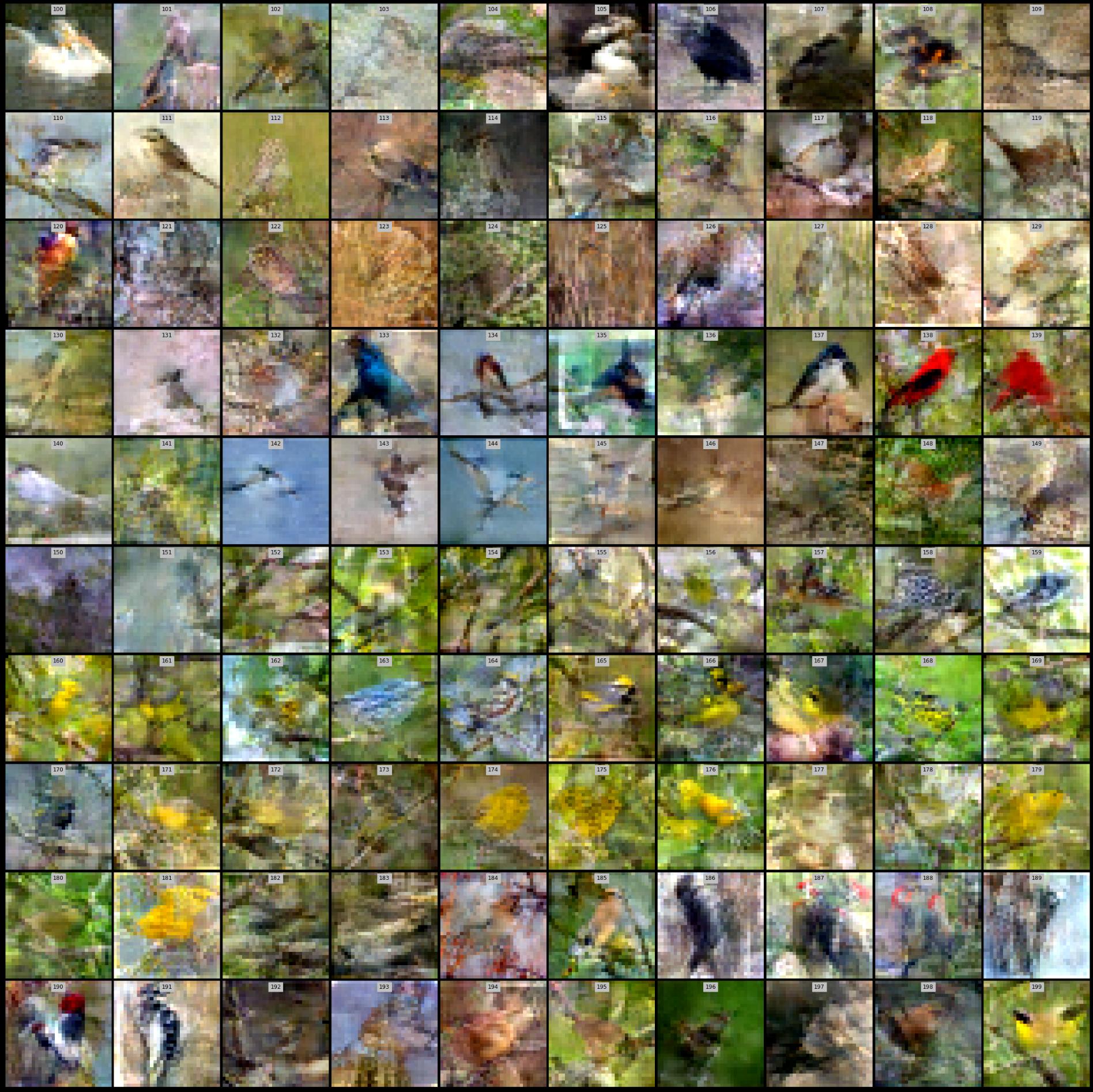}  
  \caption{Learn Label=True, Class ID = [100-199]} 
\end{subfigure}

\begin{subfigure}[b]{0.47\textwidth}
  \includegraphics[width=1.0\linewidth]{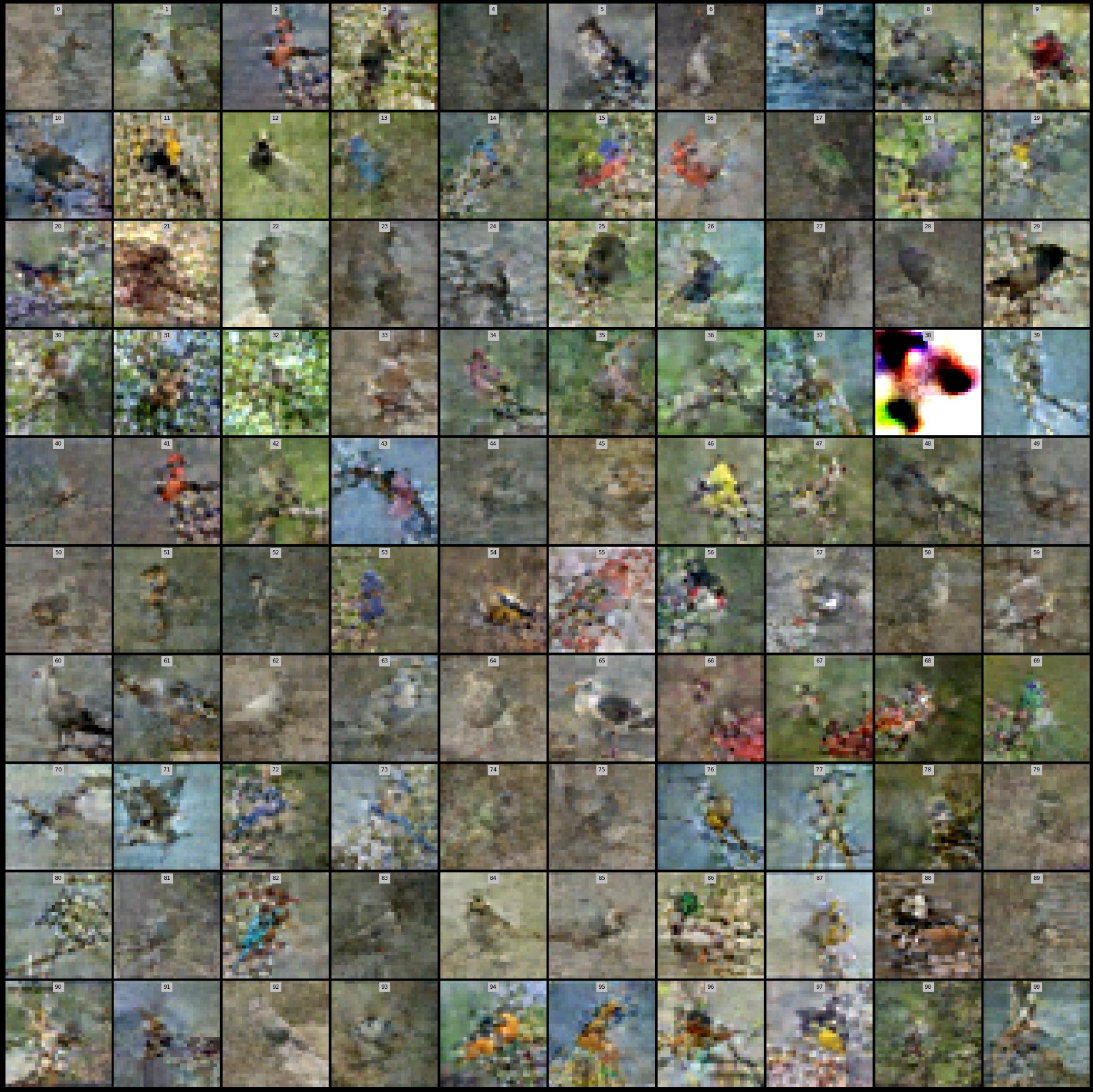}
  \caption{Learn Label=False, Class ID = [0-99]} 
\end{subfigure}
\hspace{0.2in}
\begin{subfigure}[b]{0.47\textwidth}
  \includegraphics[width=1.0\linewidth]{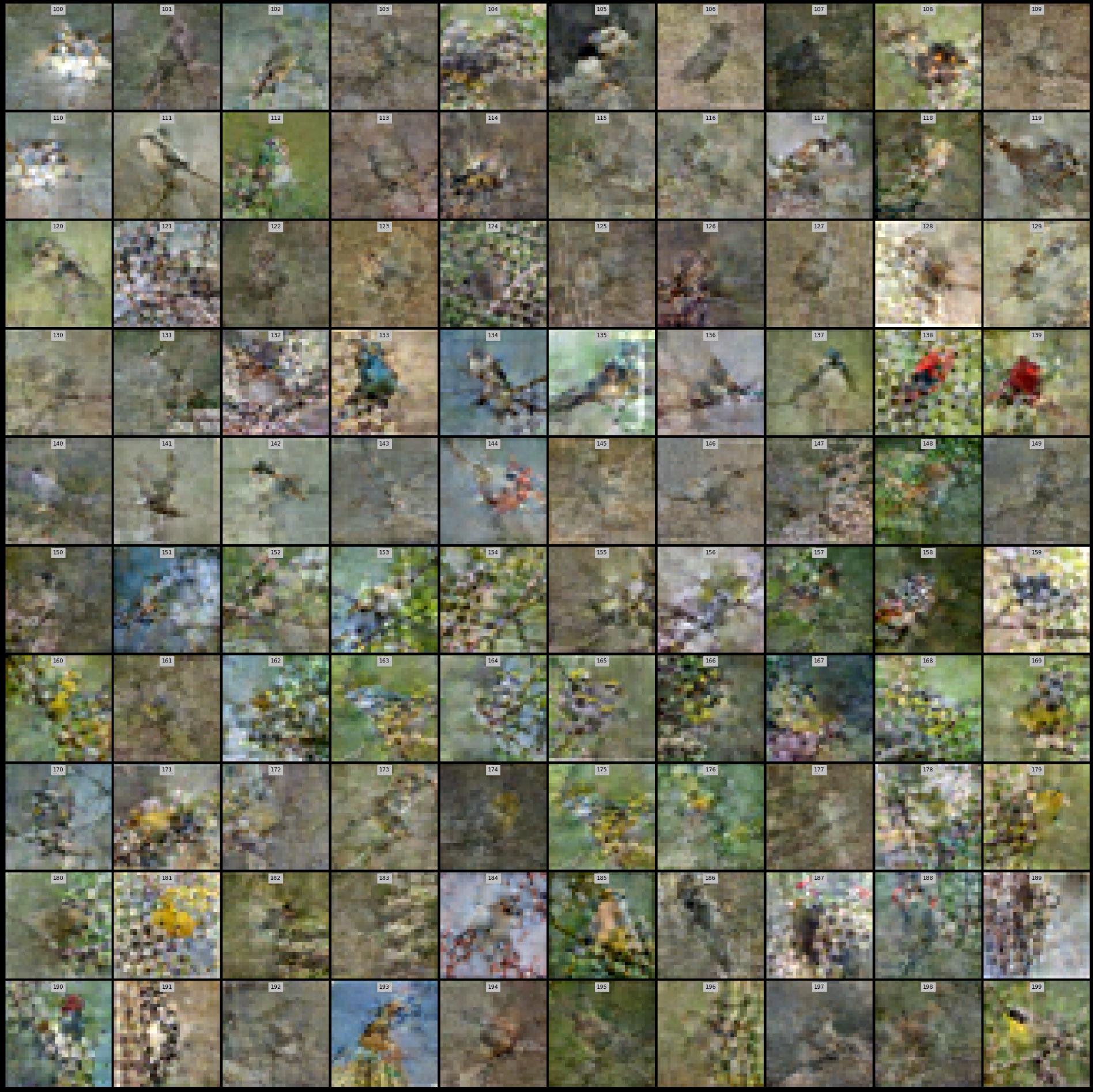}  
  \caption{Learn Label=False, Class ID = [100-199]} 
\end{subfigure}
\caption{Distilled Image Visualization - CUB-200 (1 Img/Cls)}\label{fig:vis_cub200}
\end{figure}

\begin{figure}[htbp]
\centering
\begin{subfigure}[b]{0.47\textwidth}
  \includegraphics[width=1.0\linewidth]{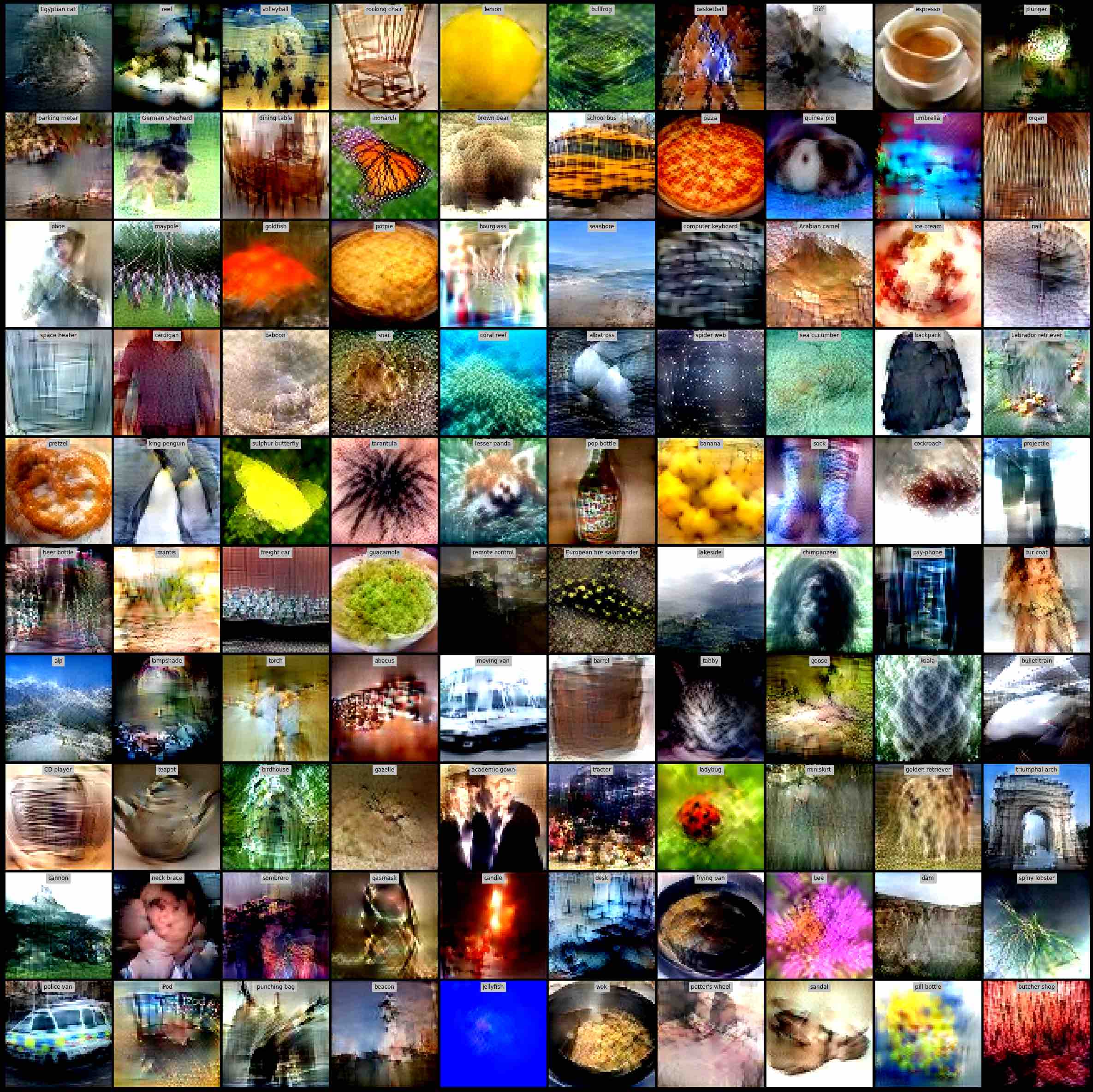}
  \caption{Learn Label=True, Class ID = [0-99]} 
\end{subfigure}
\hspace{0.2in}
\begin{subfigure}[b]{0.47\textwidth}
  \includegraphics[width=1.0\linewidth]{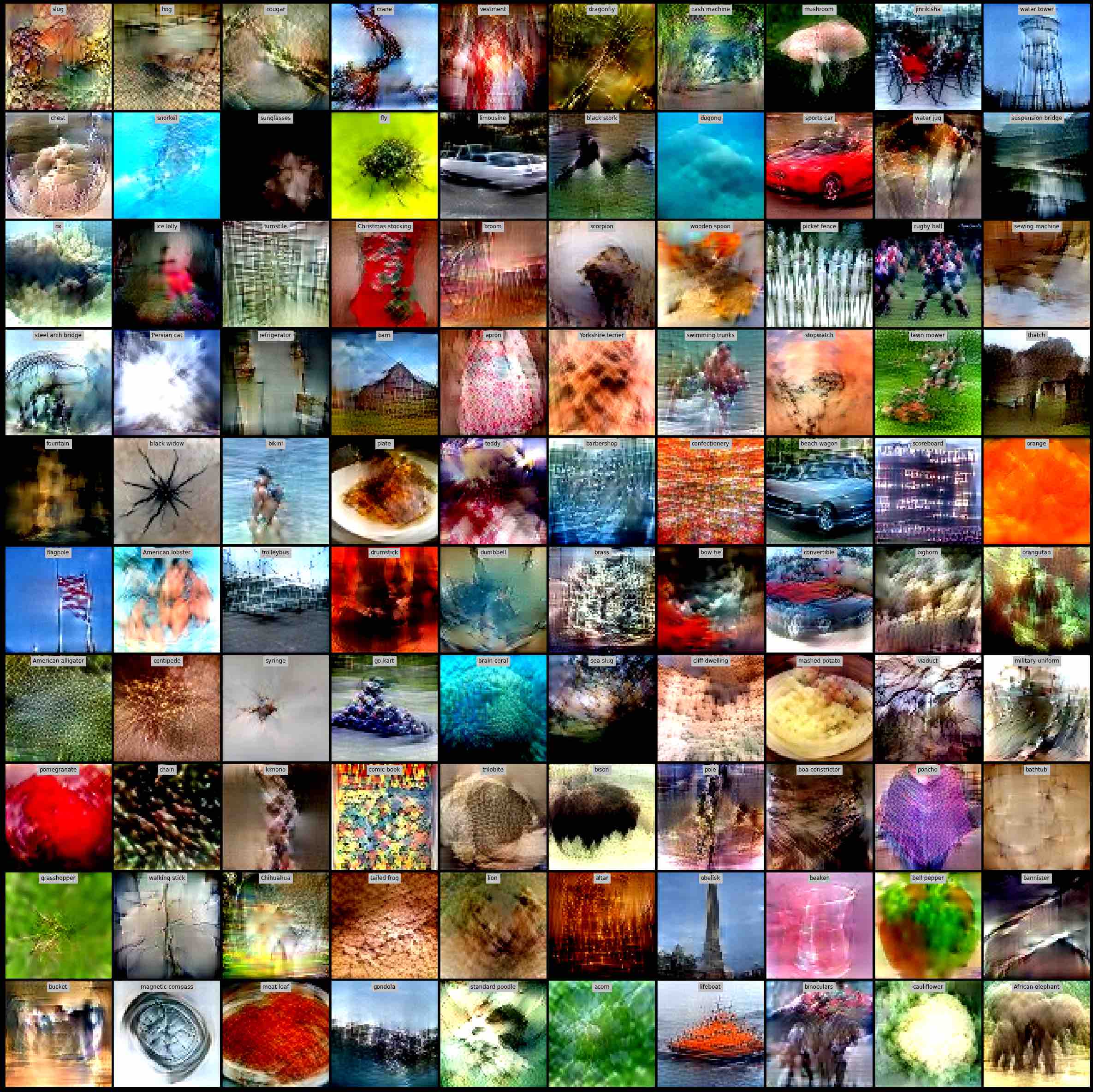}  
  \caption{Learn Label=True, Class ID = [100-199]} 
\end{subfigure}

\begin{subfigure}[b]{0.47\textwidth}
  \includegraphics[width=1.0\linewidth]{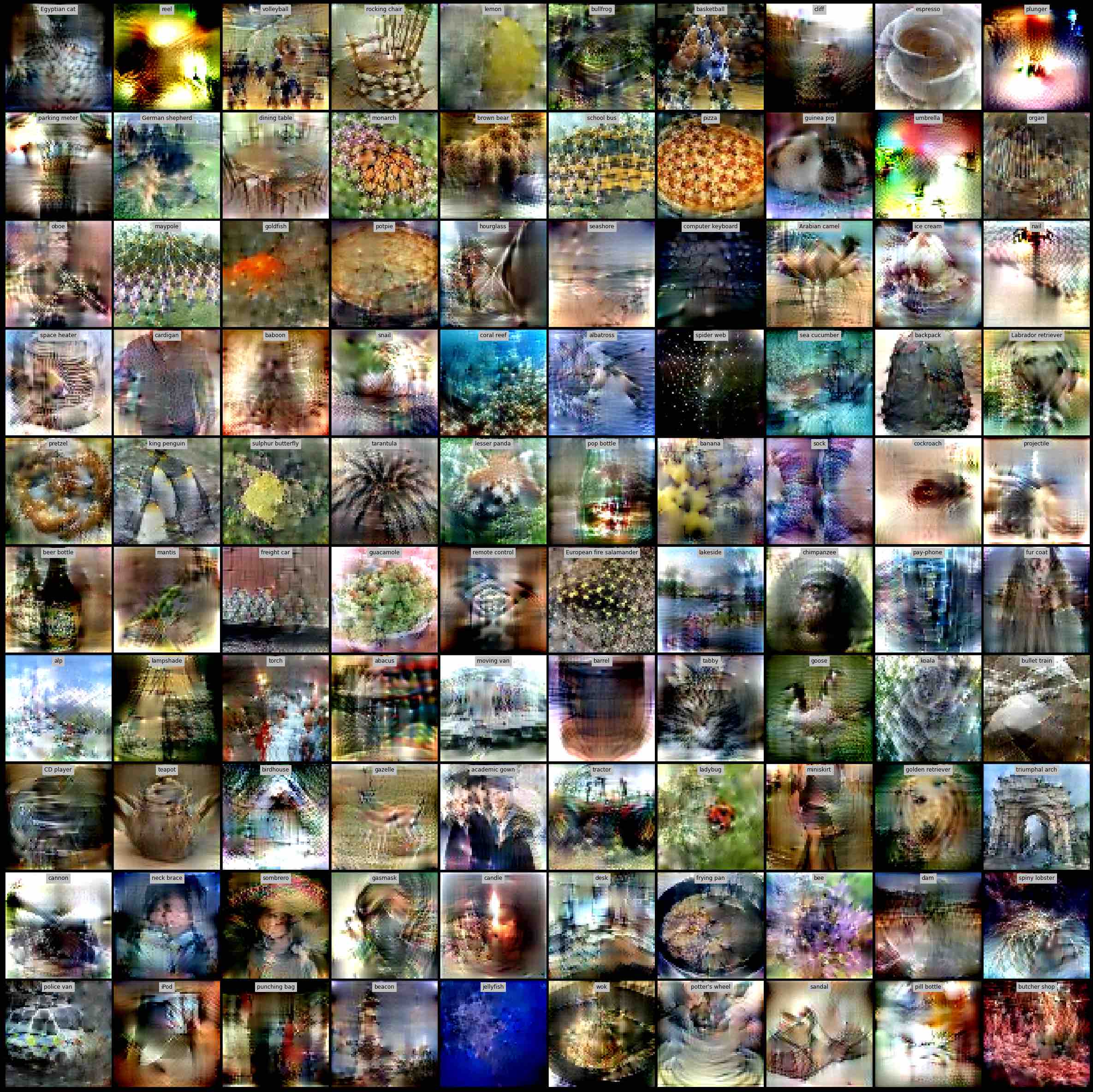}
  \caption{Learn Label=False, Class ID = [0-99]} 
\end{subfigure}
\hspace{0.2in}
\begin{subfigure}[b]{0.47\textwidth}
  \includegraphics[width=1.0\linewidth]{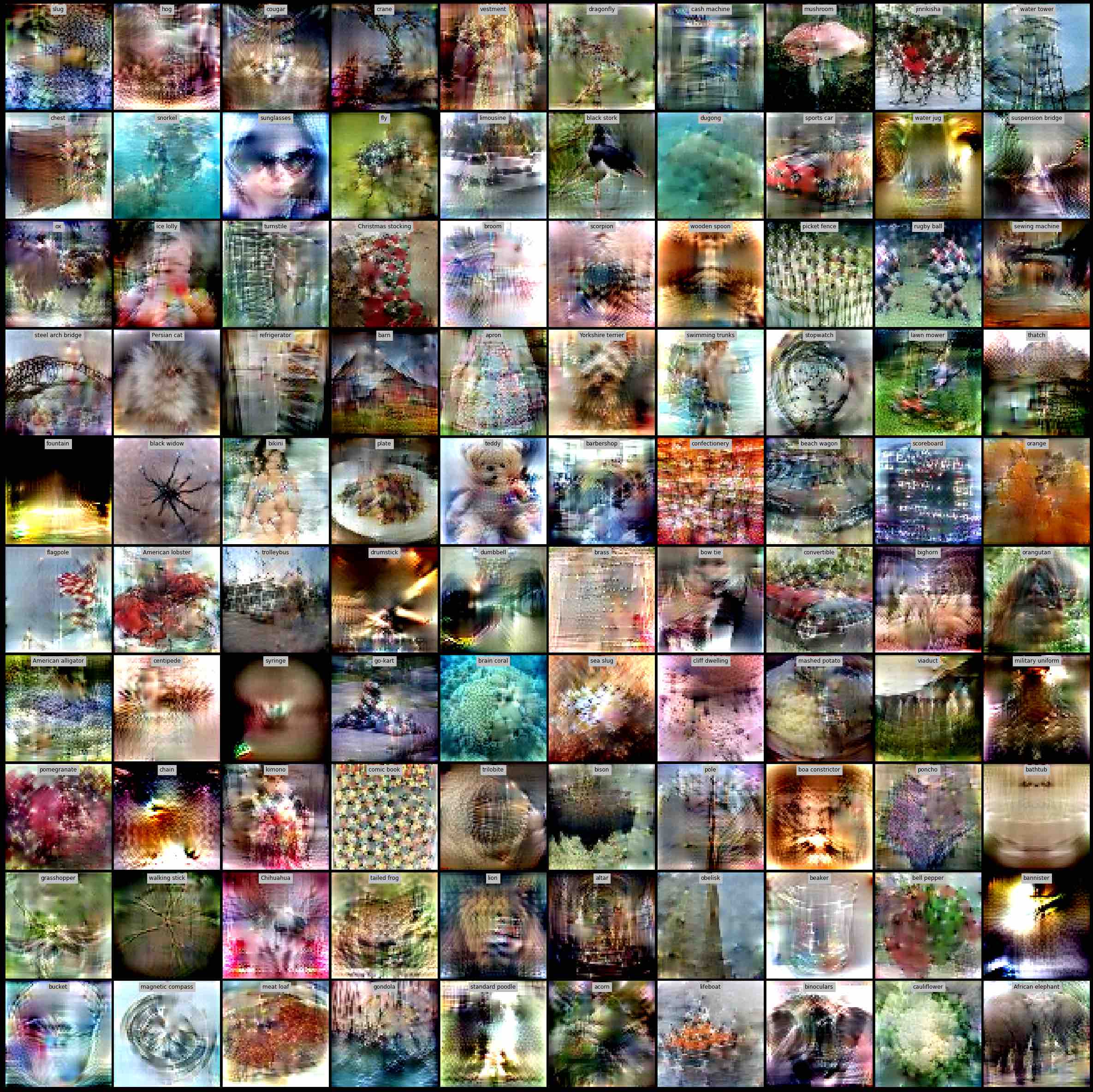}  
  \caption{Learn Label=False, Class ID = [100-199]} 
\end{subfigure}
\caption{Distilled Image Visualization - Tiny ImageNet (1 Img/Cls)}\label{fig:vis_timagenet}
\end{figure}

\begin{figure}[htbp]
\centering
\begin{subfigure}[b]{0.47\textwidth}
  \includegraphics[width=1.0\linewidth]{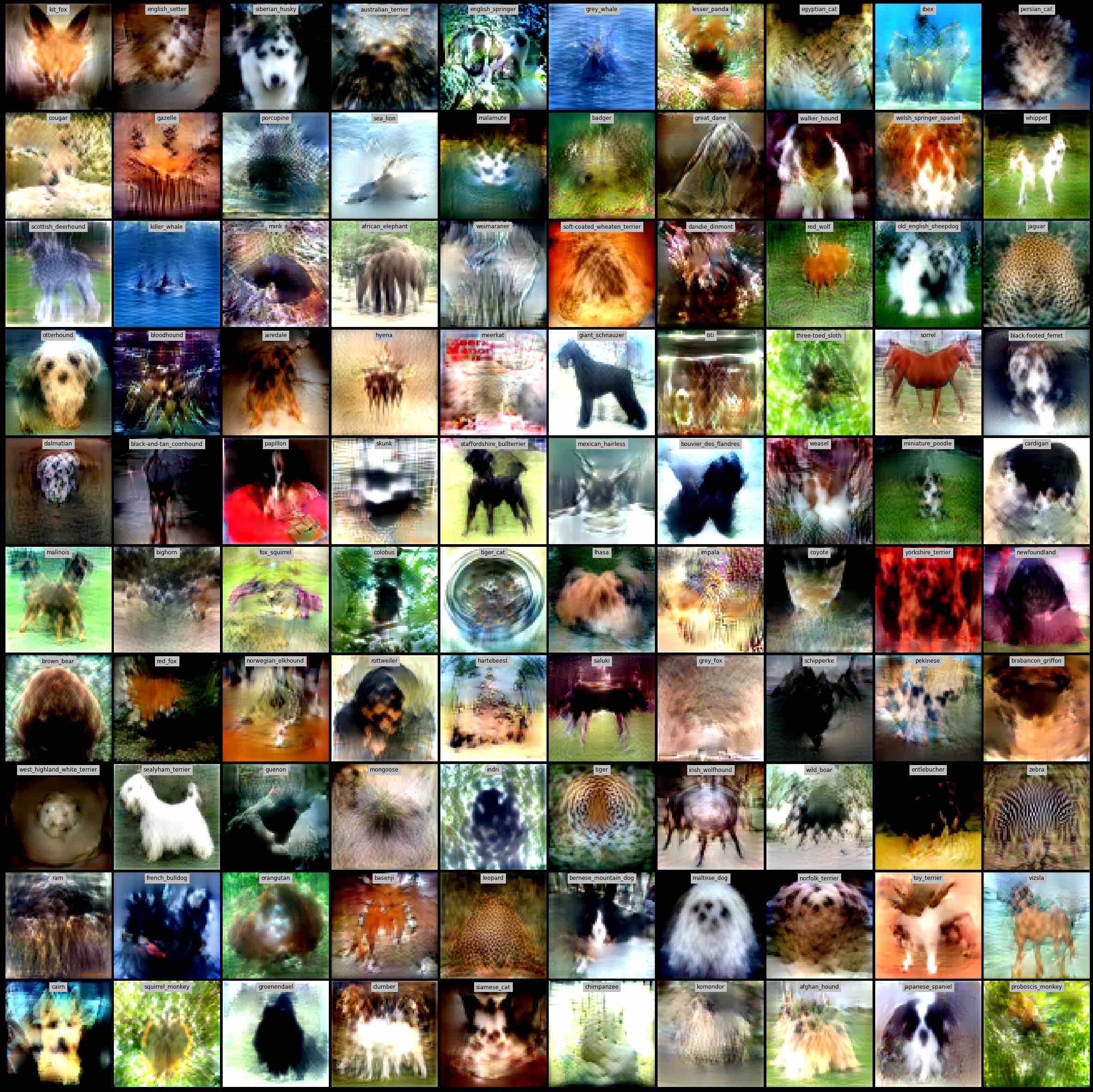}
  \caption{Class ID = [0-99]} 
\end{subfigure}
\hspace{0.2in}
\begin{subfigure}[b]{0.47\textwidth}
  \includegraphics[width=1.0\linewidth]{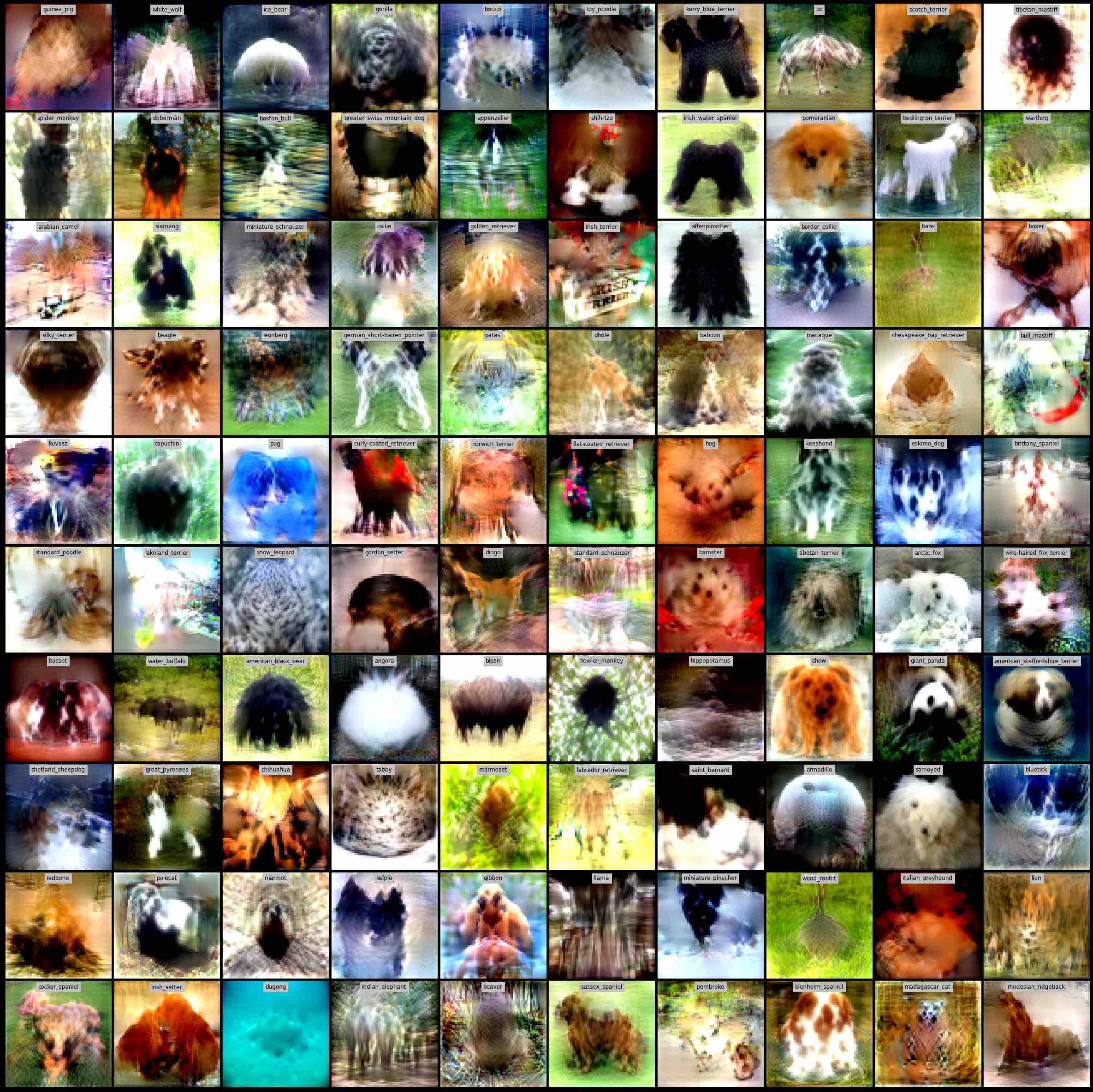}  
  \caption{Class ID = [100-199]} 
\end{subfigure}

\begin{subfigure}[b]{0.47\textwidth}
  \includegraphics[width=1.0\linewidth]{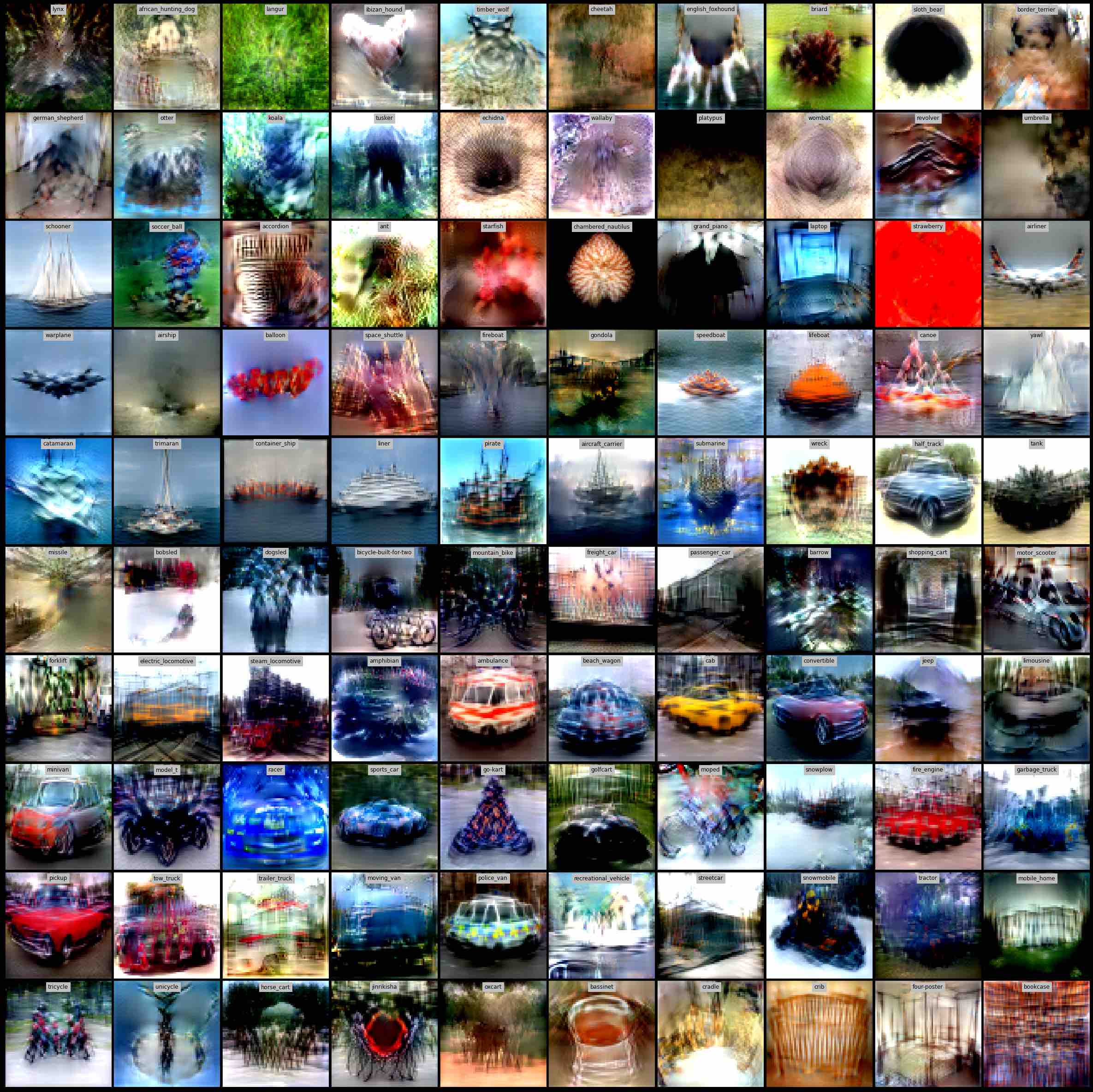}
  \caption{Class ID = [200-299]} 
\end{subfigure}
\hspace{0.2in}
\begin{subfigure}[b]{0.47\textwidth}
  \includegraphics[width=1.0\linewidth]{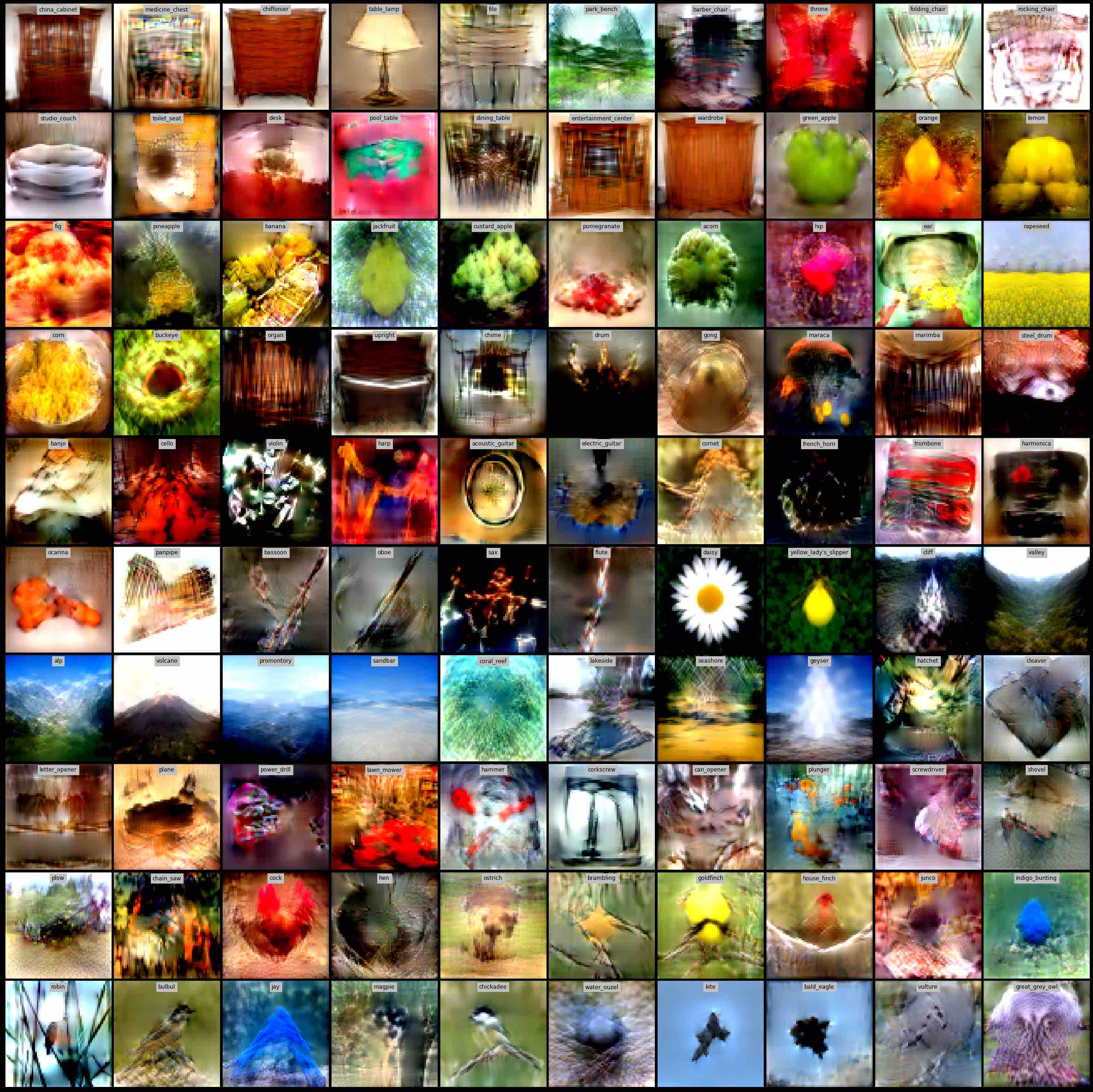}  
  \caption{Class ID = [300-399]} 
\end{subfigure}

\begin{subfigure}[b]{0.47\textwidth}
  \includegraphics[width=1.0\linewidth]{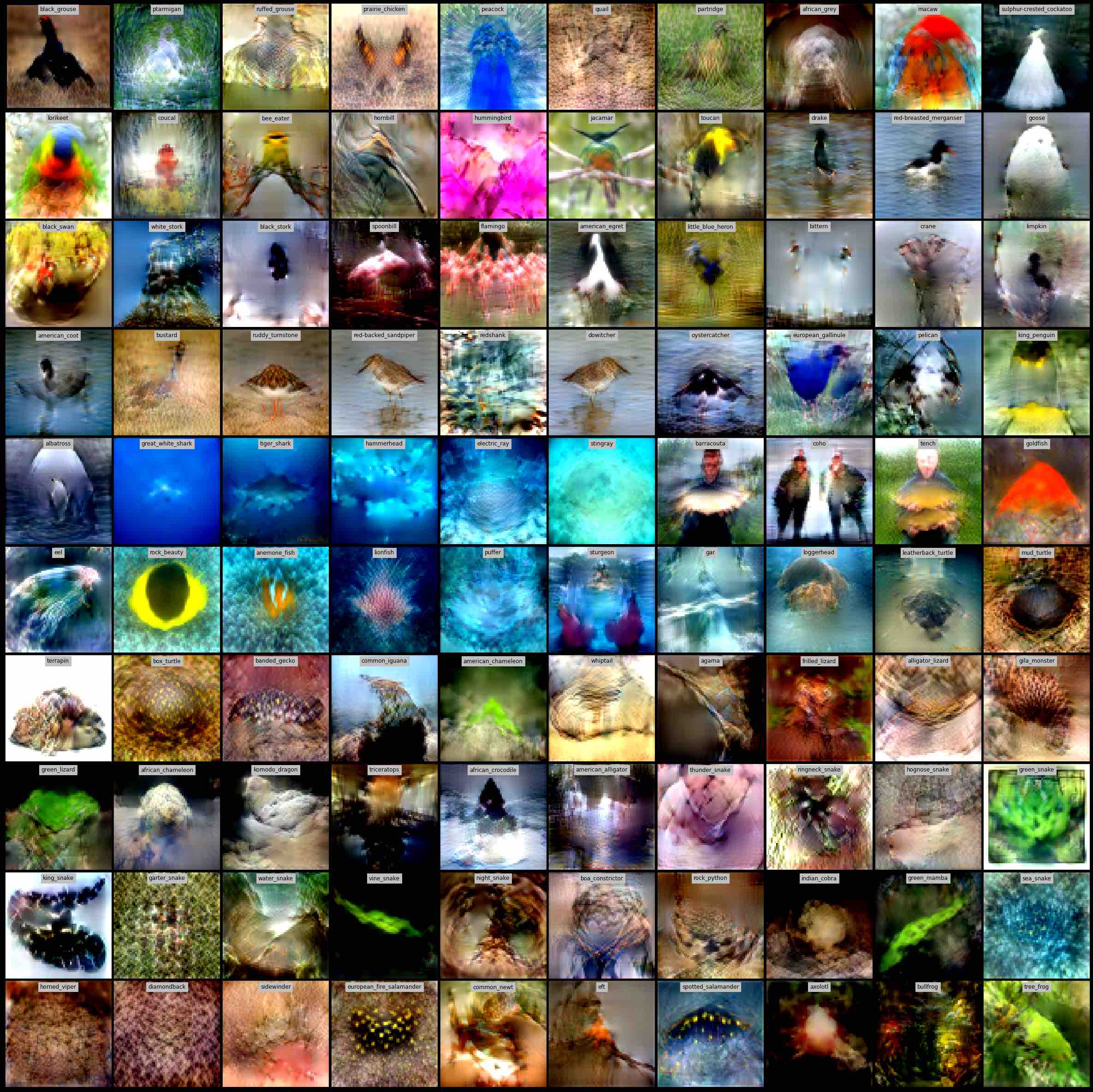}
  \caption{Class ID = [400-499]} 
\end{subfigure}
\hspace{0.2in}
\begin{subfigure}[b]{0.47\textwidth}
  \includegraphics[width=1.0\linewidth]{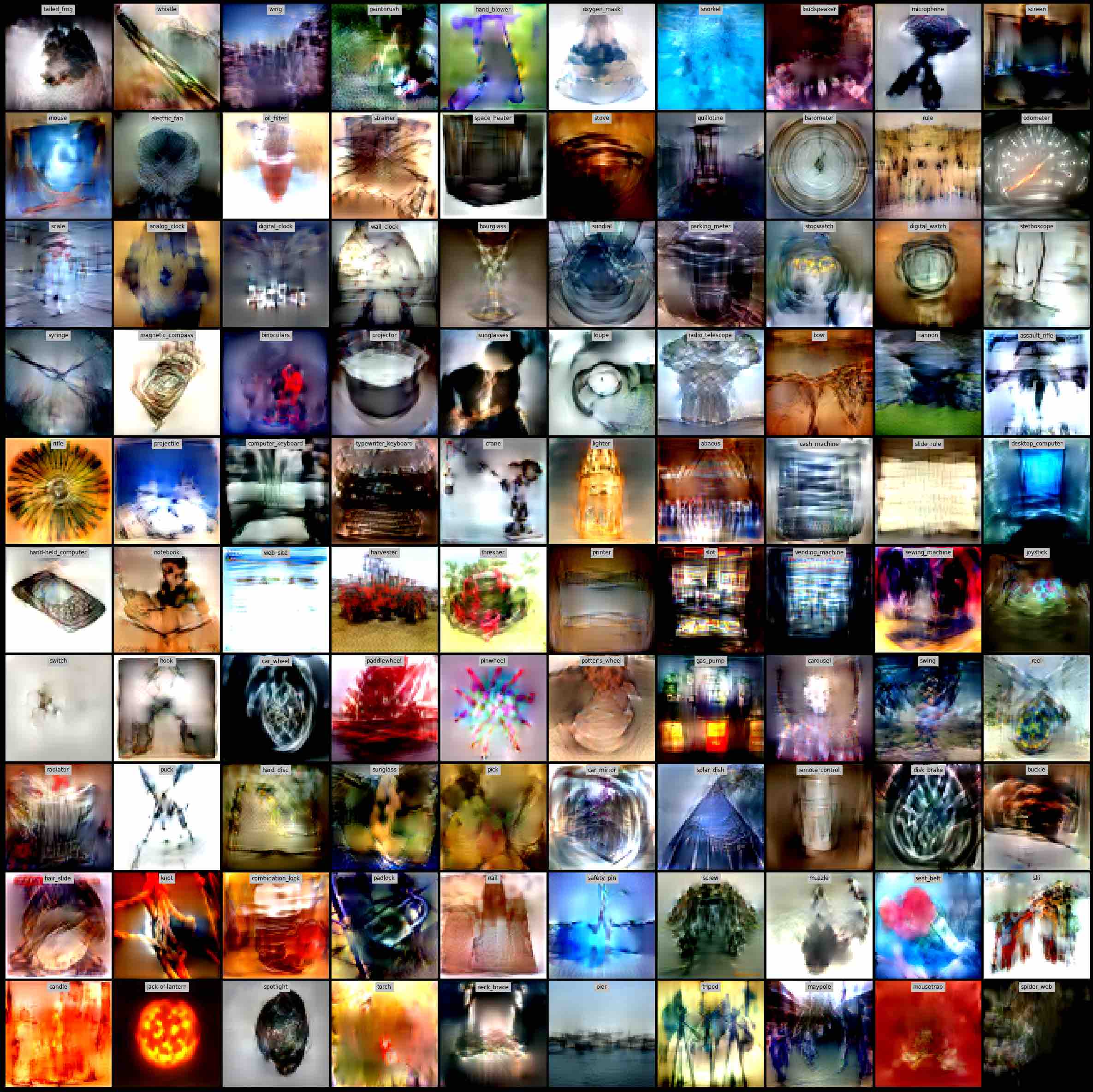}  
  \caption{Class ID = [500-599]} 
\end{subfigure}
\caption{Distilled Image Visualization - ImageNet (1 Img/Cls), Learn Label=True}\label{fig:vis_imagenet1}
\end{figure}

\begin{figure}[htbp]
\centering
\begin{subfigure}[b]{0.47\textwidth}
  \includegraphics[width=1.0\linewidth]{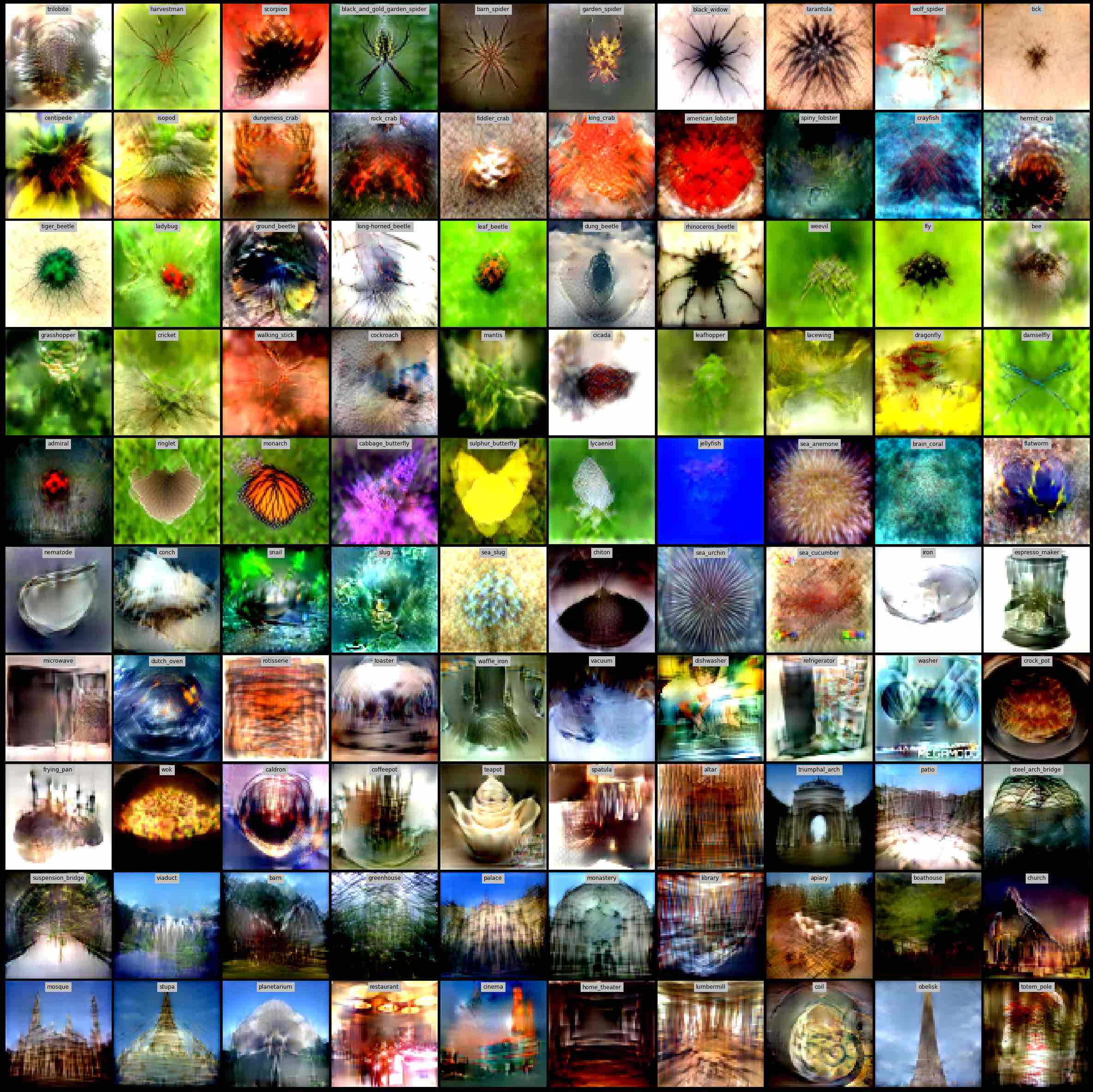}
  \caption{Class ID = [600-699]} 
\end{subfigure}
\hspace{0.2in}
\begin{subfigure}[b]{0.47\textwidth}
  \includegraphics[width=1.0\linewidth]{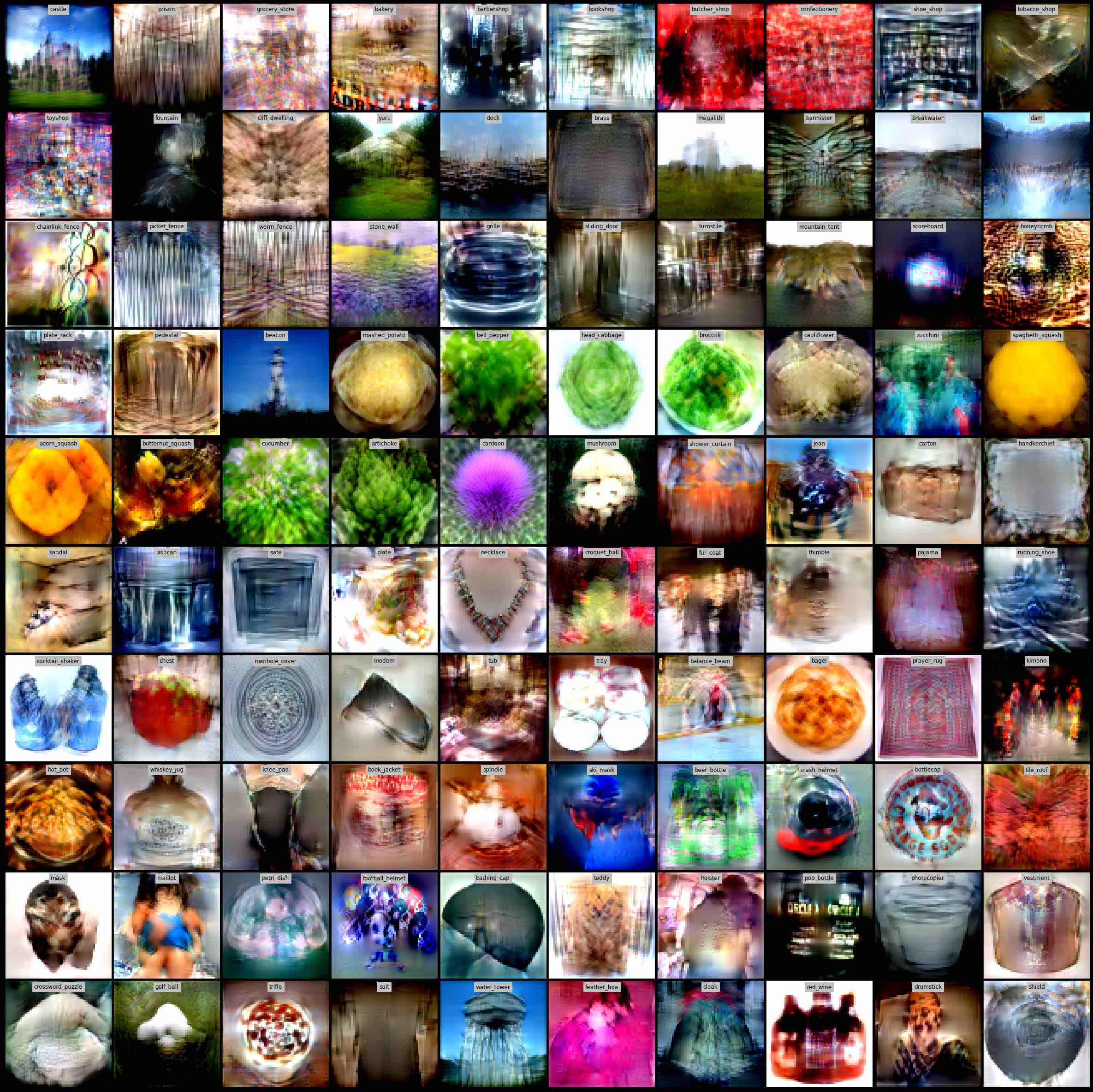}  
  \caption{Class ID = [700-799]} 
\end{subfigure}
\begin{subfigure}[b]{0.47\textwidth}
  \includegraphics[width=1.0\linewidth]{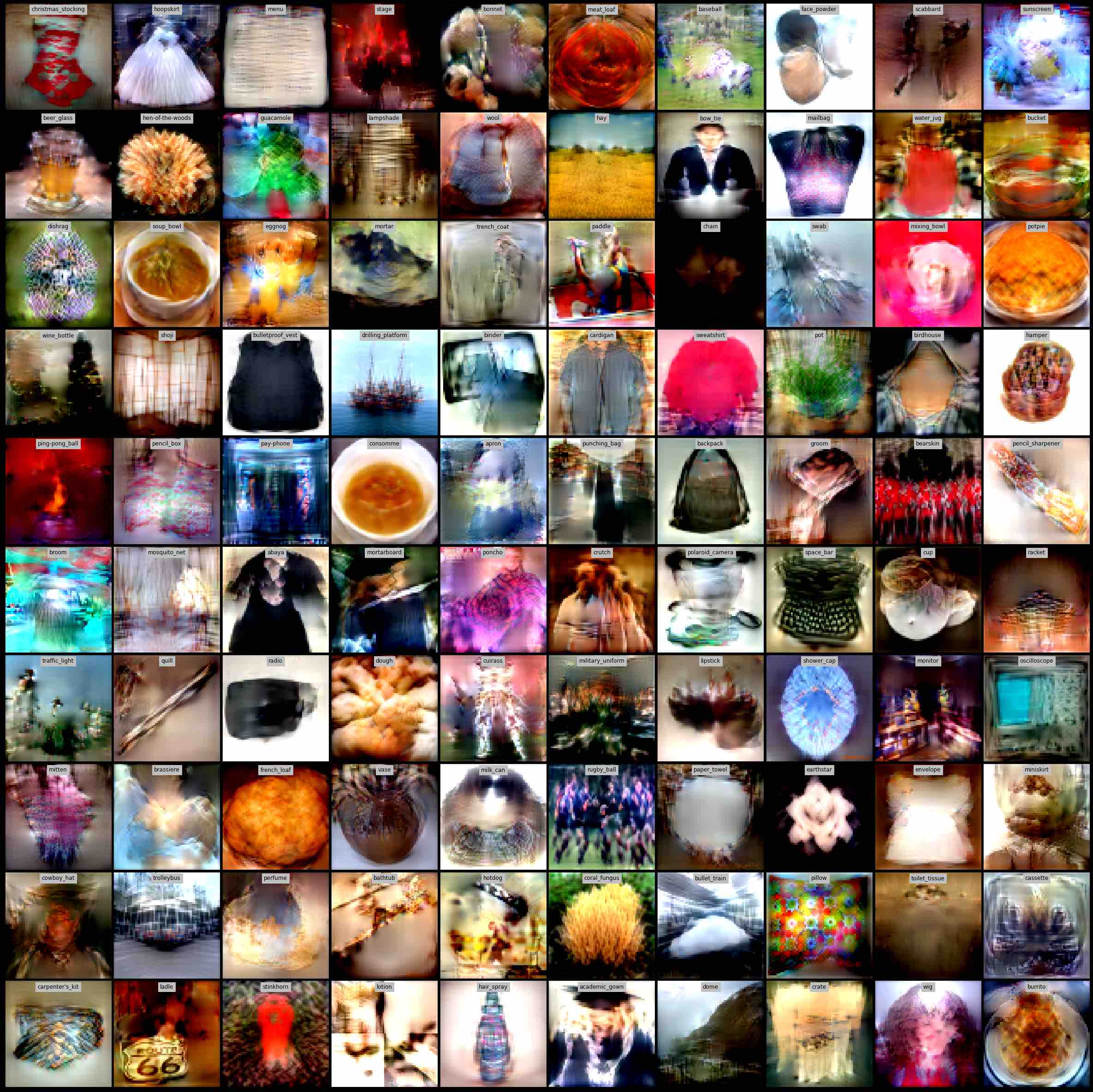}
  \caption{Class ID = [800-899]} 
\end{subfigure}
\hspace{0.2in}
\begin{subfigure}[b]{0.47\textwidth}
  \includegraphics[width=1.0\linewidth]{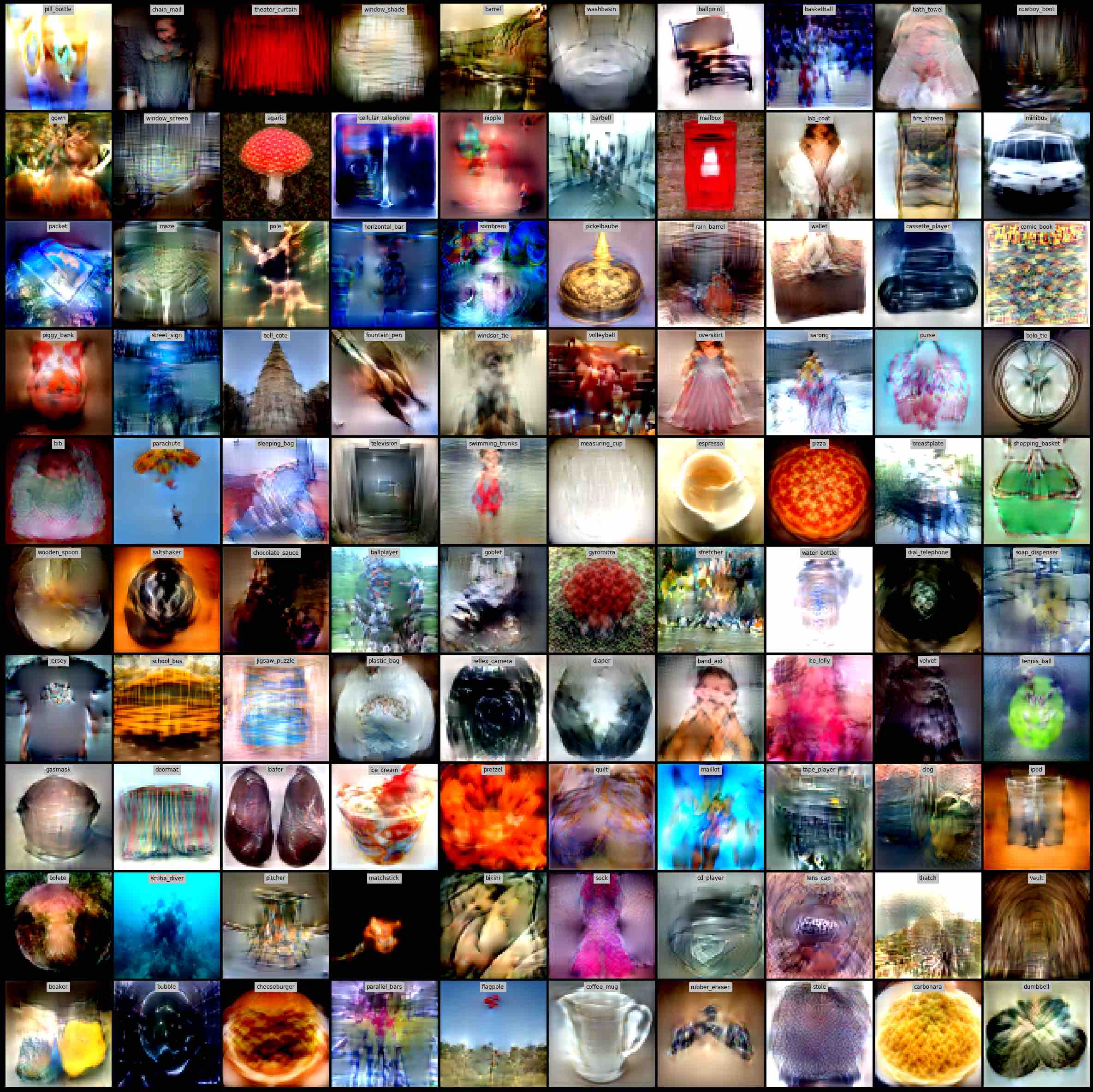}  
  \caption{Class ID = [900-999]} 
\end{subfigure}
\caption{Distilled Image Visualization - ImageNet (1 Img/Cls), Learn Label=True}\label{fig:vis_imagenet2}
\end{figure}

\begin{figure}[htbp]
\centering
\begin{subfigure}[b]{0.47\textwidth}
  \includegraphics[width=1.0\linewidth]{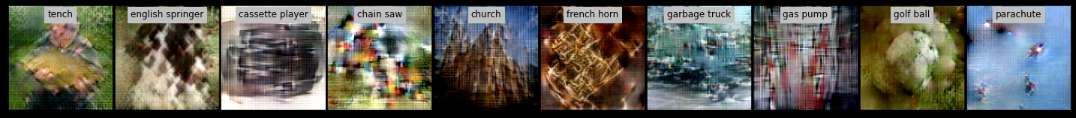}
  \caption{1 Img/Cls, Learn Label=True} 
\end{subfigure}
\hspace{0.2in}
\begin{subfigure}[b]{0.47\textwidth}
  \includegraphics[width=1.0\linewidth]{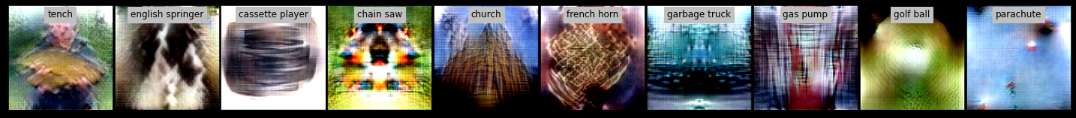}  
  \caption{1 Img/Cls, Learn Label=False} 
\end{subfigure}

\begin{subfigure}[b]{0.47\textwidth}
  \includegraphics[width=1.0\linewidth]{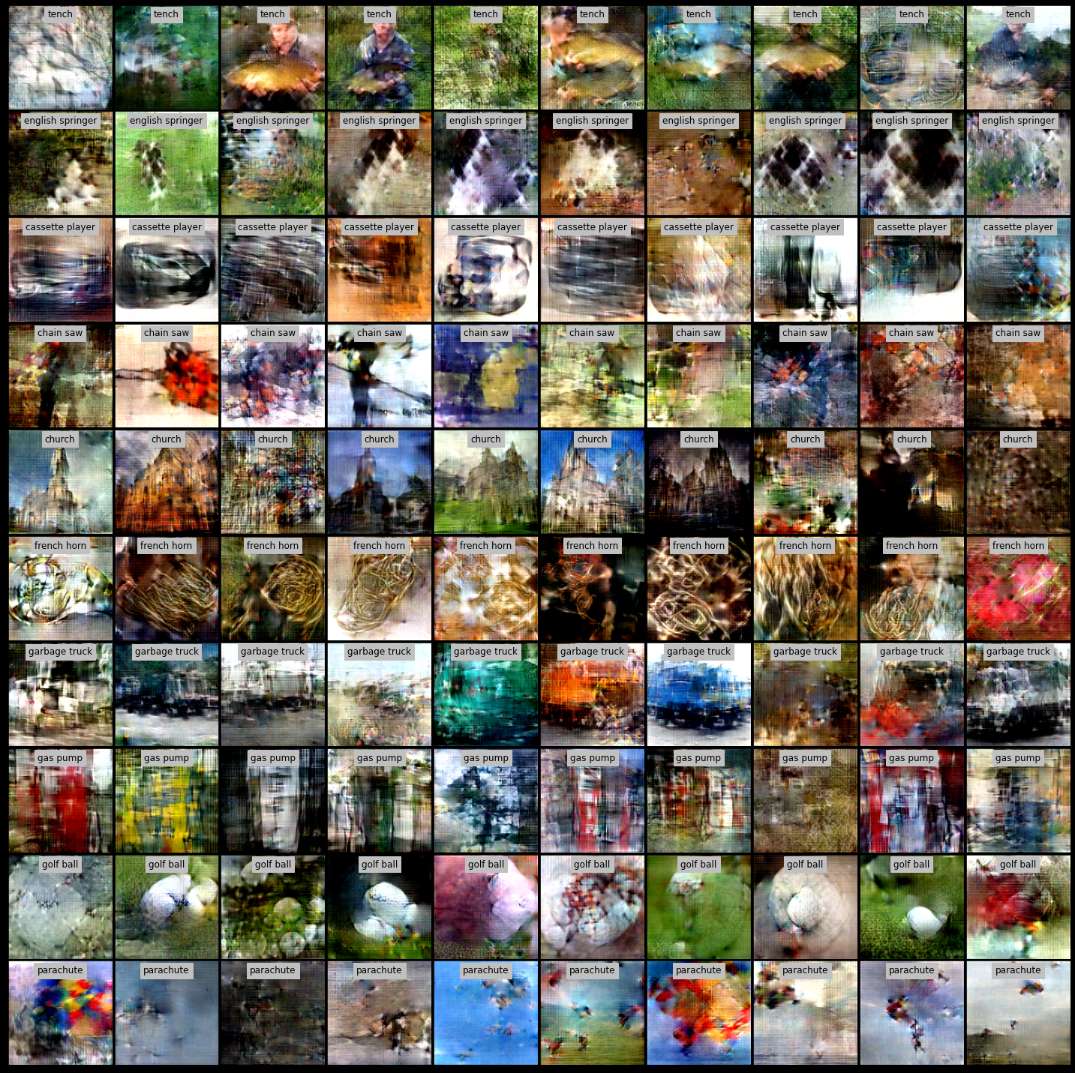}
  \caption{10 Img/Cls, Learn Label=True} 
\end{subfigure}
\hspace{0.2in}
\begin{subfigure}[b]{0.47\textwidth}
  \includegraphics[width=1.0\linewidth]{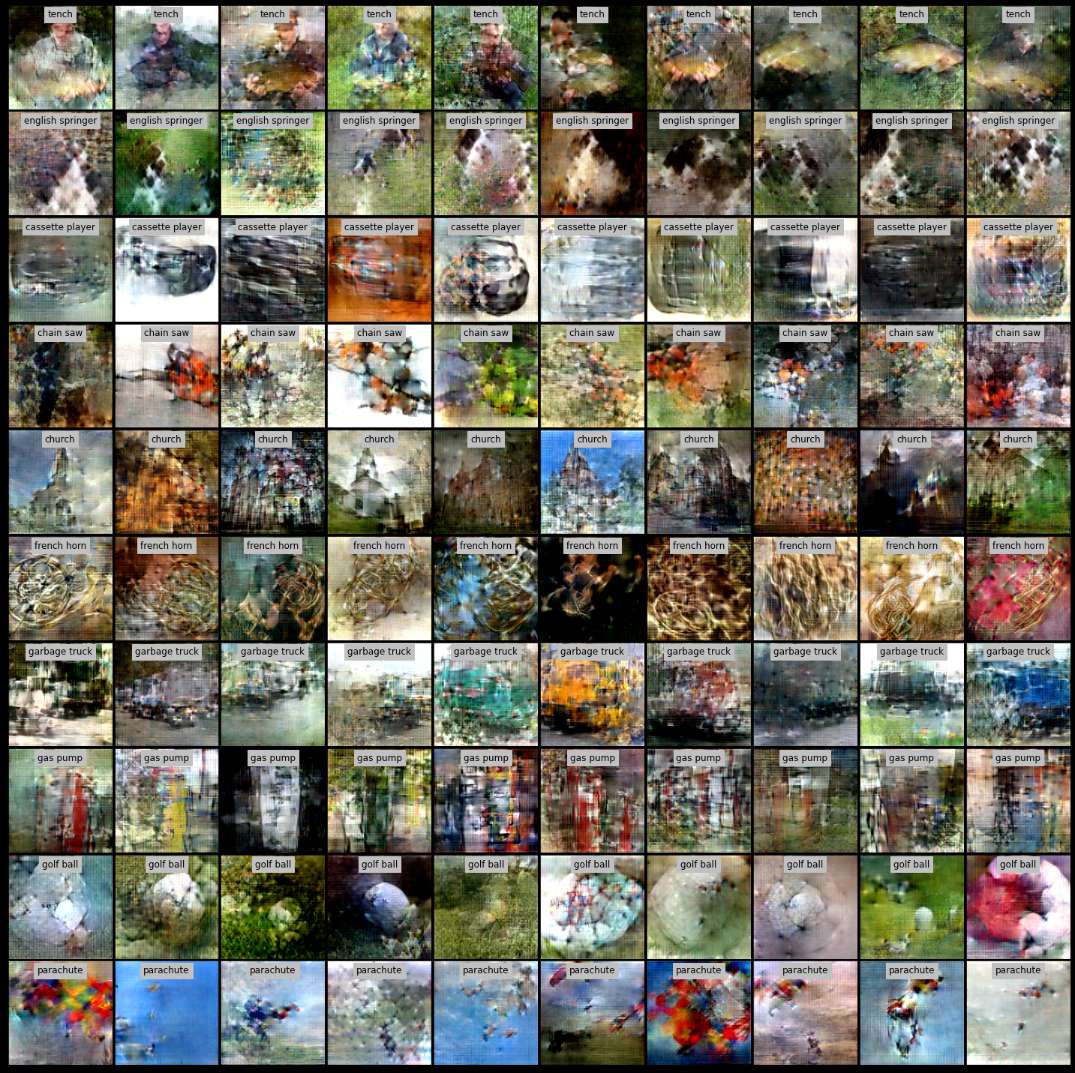}  
  \caption{10 Img/Cls, Learn Label=False} 
\end{subfigure}
\caption{Distilled Image Visualization - ImageNette.}\label{fig:vis_imagenette}
\end{figure}

\begin{figure}[htbp]
\centering
\begin{subfigure}[b]{0.47\textwidth}
  \includegraphics[width=1.0\linewidth]{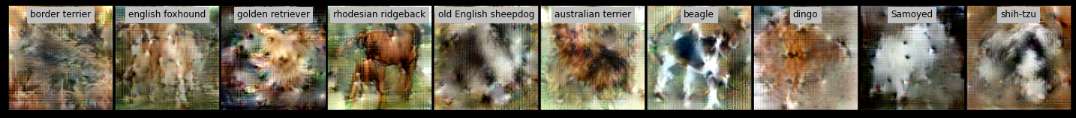}
  \caption{1 Img/Cls, Learn Label=True} 
\end{subfigure}
\hspace{0.2in}
\begin{subfigure}[b]{0.47\textwidth}
  \includegraphics[width=1.0\linewidth]{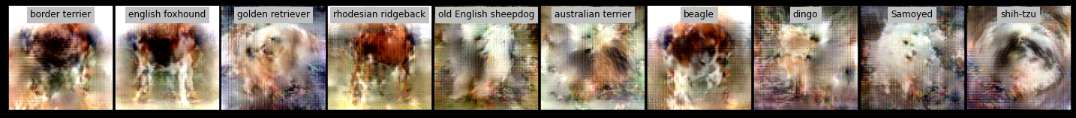}  
  \caption{1 Img/Cls, Learn Label=False} 
\end{subfigure}

\begin{subfigure}[b]{0.47\textwidth}
  \includegraphics[width=1.0\linewidth]{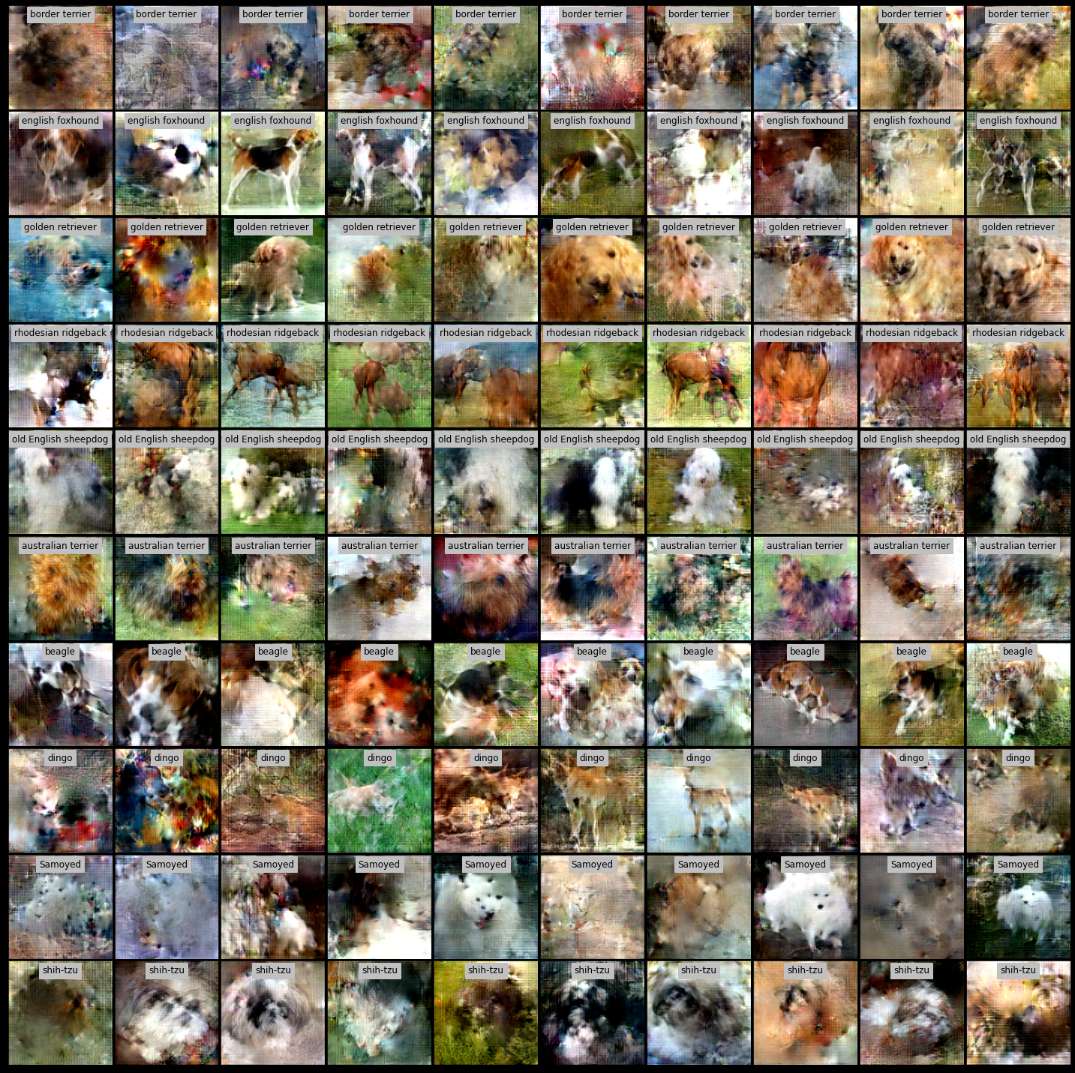}
  \caption{10 Img/Cls, Learn Label=True} 
\end{subfigure}
\hspace{0.2in}
\begin{subfigure}[b]{0.47\textwidth}
  \includegraphics[width=1.0\linewidth]{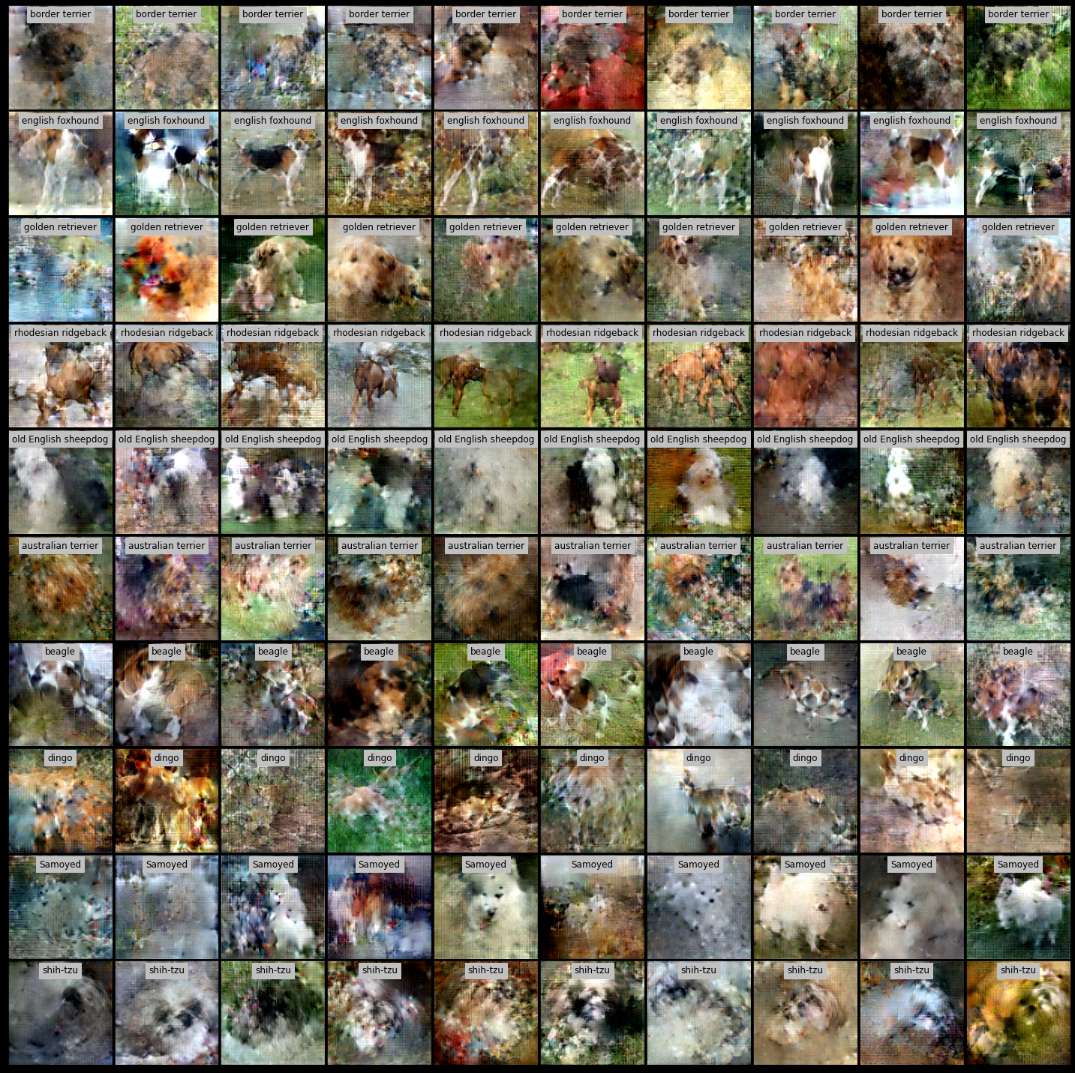}  
  \caption{10 Img/Cls, Learn Label=False} 
\end{subfigure}
\caption{Distilled Image Visualization - ImageWoof.}\label{fig:vis_imagewoof}
\end{figure}

We provide some additional distilled images visualization for MNIST (Figure  \ref{fig:vis_mnist}), FashionMNIST (Figure \ref{fig:vis_fmnist}), CIFAR10 (Figure \ref{fig:vis_cifar10}), CIFAR100 (Figure \ref{fig:vis_cifar100}), CUB-200 (Figure \ref{fig:vis_cub200}), Tiny ImageNet (Figure \ref{fig:vis_timagenet}), ImageNet (Figure \ref{fig:vis_imagenet1}, \ref{fig:vis_imagenet2}), ImageNette (Figure \ref{fig:vis_imagenette}), and ImageWoof (Figure \ref{fig:vis_imagewoof}). 

\end{document}